\documentclass{article}

\usepackage[preprint, nonatbib]{neurips_2026}        

\usepackage[utf8]{inputenc} 
\usepackage[T1]{fontenc}    
\usepackage{hyperref}       
\usepackage{url}            
\usepackage{booktabs}       
\usepackage{amsfonts}       
\usepackage{amsmath}        
\usepackage{amssymb}        
\usepackage{bm}             
\usepackage{graphicx}       
\usepackage{algorithm}      
\usepackage{algpseudocode}  
\usepackage{nicefrac}       
\usepackage{microtype}      
\usepackage{xcolor}         
\usepackage{etoc}           
\usepackage{placeins}       
\usepackage[numbers,compress]{natbib}  
\usepackage[colorinlistoftodos]{todonotes}
\newcommand{\paratio}{fig_pa/ratio_sweep_figures}
\newcommand{\pabudget}{fig_pa/budget_scaling_figures}
\newcommand{\pacost}{fig_pa/cost_sweep_figs}
\newcommand{\pacross}{fig_pa/cross_source_extra_figs}
\newcommand{\parw}{fig_pa/plots_rw_NC}

\title{MAST: A Multi-fidelity Augmented Surrogate model via Spatial Trust-weighting}

\author{%
  Ahmed Mohamed Eisa Nasr \\
  Department of Aeronautics and Astronautics\\
  University of Southampton\\
  SO17 1BJ, Southampton, United Kingdom \\
  \texttt{A.M.E.Nasr@soton.ac.uk}
  \And
  Ali Elham \\
  Department of Aeronautics and Astronautics\\
  University of Southampton\\
  SO17 1BJ, Southampton, United Kingdom \\
  \texttt{A.Elham@soton.ac.uk}
  \And
  Haris Moazam Sheikh \\
  Department of Aeronautics and Astronautics\\
  University of Southampton\\
  SO17 1BJ, Southampton, United Kingdom \\
  \texttt{H.M.Sheikh@soton.ac.uk}
}

\begin{document}

\maketitle

\etocdepthtag.toc{main}  

\begin{abstract}
In engineering design and scientific computing, computational cost and predictive accuracy are intrinsically coupled. High-fidelity simulations provide accurate predictions but at substantial computational costs, while lower-fidelity approximations offer efficiency at the expense of accuracy. Multi-fidelity surrogate modelling addresses this trade-off by combining abundant low-fidelity data with sparse high-fidelity observations. However, existing methods rely on global correlation assumptions that can often fail in practice to capture how fidelity relationships vary across the input space, leading to poor performance, particularly under tight budget constraints. We introduce MAST, a method that blends corrected low-fidelity observations with high-fidelity predictions, trusting high-fidelity near observed samples and relying on corrected low-fidelity elsewhere. MAST achieves this through explicit discrepancy modelling and distance-based weighting with closed-form variance propagation, producing a single heteroscedastic Gaussian process. Across multi-fidelity synthetic benchmarks, MAST shows a marked improvement over the current state-of-the-art techniques. Crucially, MAST maintains robust performance across varying total budget and fidelity gaps, conditions under which competing methods exhibit significant degradation or unstable behaviour. More broadly, MAST provides a spatially adaptive framework for multi-fidelity Gaussian-process modelling, in which the contribution of low-fidelity information is governed by its proximity to high-fidelity calibration data, opening a new direction for more reliable surrogate construction under sparse and budget-constrained settings.
\end{abstract}

\section{Introduction}

The trade-off between predictive accuracy and computational cost is a fundamental constraint in modern simulation-based design and optimisation. High-fidelity models, while necessary for precise uncertainty quantification and bias correction, are often too  expensive to sample densely for accurate surrogate construction. Conversely, low-fidelity approximations are abundant and cheap but can suffer from systematic errors that preclude their standalone use. Multi-fidelity Gaussian process (MFGP) models \citep{Do2024MULTI-FIDELITYREVIEW, Fernandez-Godino2024ReviewModels, Morris2004TheExperiments, Forrester2007Multi-fidelityModelling} bridge the two by exploiting their statistical correlation, recovering an accurate response surface at a fraction of the cost of an HF-only fit \citep{Perdikaris2015Multi-fidelityFields}.

The key assumption behind most multi-fidelity methods is that lower-fidelity models, although biased, remain sufficiently correlated with the high-fidelity response for sparse high-fidelity data to correct the residual discrepancy. Existing approaches encode this assumption through \emph{global correlation structures}. Autoregressive models \citep{Kennedy2000PredictingAvailable, LeGratiet2014RecursiveFidelity} assume linear inter-fidelity relationships that hold across the input domain; multi-output formulations \citep{Alvarez2012KernelsReview} impose shared covariance structures that imply global cross-fidelity correlations; and fidelity-augmented kernel methods \citep{Poloczek2017Multi-informationOptimization, Wu2020PracticalTuning} learn global kernel parameters that must extrapolate from sparse high-fidelity data. Fusion-based methods \citep{Lam2015MultifidelitySources, Thomison2017ASources} avoid hierarchical assumptions by combining independent per-fidelity surrogates at the distributional level, but fuse the models at prediction time rather than constructing a single augmented surrogate. However, when the inter-fidelity relationships vary \textit{spatially} across the input space \citep{Yu2026FIRE}, global correlation assumptions can lead to model misspecification and degraded predictive performance. More expressive approaches, such as nonlinear autoregressive GPs \citep{Perdikaris2017NonlinearModelling} and deep multi-fidelity GPs \citep{Cutajar2019DeepModeling}, can capture complex inter-fidelity mappings but require careful hyperparameter tuning and more training data \citep{Brevault2020OverviewSystems}, both of which are often limiting under tight budget constraints.

We introduce \textbf{MAST} (\emph{Multi-fidelity Augmented Surrogate model with Spatial Trust-weighting}), which replaces classical correlation learning with a \emph{geometric inductive bias}. The core insight is simple: corrected low-fidelity observations should be trusted far from high-fidelity samples where no calibration information exists, but should progressively yield to high-fidelity predictions as high-fidelity data become available \textit{nearby}. MAST operationalises this principle through explicit discrepancy modelling followed by distance-based, cost-aware weighted fusion of corrected low-fidelity values with high-fidelity posterior predictions. The uncertainty induced by both discrepancy estimation and spatial blending is analytically propagated into fixed heteroscedastic noise terms, yielding a single Gaussian process surrogate that naturally down-weights unreliable data through geometric proximity rather than learned correlation structures. Our contributions can be summarised as:
\begin{enumerate}
    \item A geometrically driven augmentation scheme that adapts inter-fidelity fusion locally without parametric correlation assumptions.
    \item Explicit incorporation of fidelity cost into trust weighting, establishing an economic hierarchy that governs information fusion.
    \item Closed-form variance propagation enabling efficient single-GP training while avoiding optimisation instabilities of learned heteroscedastic models.
    \item State-of-the-art predictive accuracy and uncertainty calibration with robust performance across budget allocations, total budget scales, and inter-fidelity correlation levels.
\end{enumerate}
The remainder of this paper is organised as follows. Section~\ref{sec:relatedwork} surveys related work. Section~\ref{sec:background} reviews Gaussian process fundamentals. Section~\ref{sec:methodology} presents the MAST framework. Section~\ref{sec:syn} presents the synthetic experimental setup. Section~\ref{sec:results} reports experimental results. Section~\ref{sec:discussion} discusses the findings and reports a three-perturbation sensitivity analysis. Section~\ref{sec:conclusion} concludes.

\section{Related Work}
\label{sec:relatedwork}

Existing multi-fidelity surrogate modelling approaches \citep{Fernandez-Godino2024ReviewModels, Perdikaris2015Multi-fidelityFields, Wang2023RecentOptimization} can be broadly categorised by how inter-fidelity relationships are modelled.

\paragraph{Linear Autoregressive Models.}
The foundational framework for multi-fidelity GP modelling was introduced by \citet{Kennedy2000PredictingAvailable}, who proposed an autoregressive (AR1) co-kriging model relating adjacent fidelity levels via
\begin{equation}
f^{(s)}(\mathbf{x}) = \rho_s f^{(s-1)}(\mathbf{x}) + \delta^{(s)}(\mathbf{x}),
\label{eq:ar1}
\end{equation}
where $\rho_s$ is a scalar correlation parameter and $\delta^{(s)}(\mathbf{x}) \sim \mathcal{GP}(0, k_{\delta}^{(s)})$ is an independent discrepancy process. Subsequent work extended this formulation through hierarchical and recursive variants, multi-fidelity optimisation, and gradient-enhanced modelling \citep{Han2012HierarchicalModeling, LeGratiet2014RecursiveFidelity, Forrester2007Multi-fidelityModelling, Chung2002UsingProblems, Leary2004ASimulation, Han2013ImprovingFunction}. Despite broad engineering adoption \citep{Matar2025Cost-effectiveLES, Ding2018AStructures, Liu2022Multi-fidelityOptimization, Wiegand2025RobustOptimization, Toal2011EfficientCokriging, Laurenceau2008BuildingCokriging, Keane2012CokrigingOptimization}, AR models often require nested sampling designs \citep{Wang2023RecentOptimization} and assume a spatially constant inter-fidelity relationship through the global parameter $\rho_s$, which can fail under spatially varying discrepancies \citep{Brevault2020OverviewSystems}.

\paragraph{Nonlinear and Deep Multi-Fidelity Models.}
Nonlinear autoregressive Gaussian processes (NARGP) relax the linear scaling in Eq.~\eqref{eq:ar1} by modelling $f^{(s)}(\mathbf{x}) = g^{(s)}\big(\mathbf{x}, f^{(s-1)}(\mathbf{x})\big)$, where $g^{(s)}$ is a GP over the augmented input space \citep{Perdikaris2017NonlinearModelling}. However, NARGP trains each fidelity level greedily and independently, preventing higher-fidelity information from refining lower-fidelity representations. Deep multi-fidelity GPs address this by jointly training coupled GP layers \citep{Cutajar2019DeepModeling}, but introduce non-Gaussian posteriors, approximate inference, higher computational cost, and more hyperparameters. As a result, these models can underperform simpler baselines in data-scarce regimes where hyperparameters cannot be reliably estimated \citep{Brevault2020OverviewSystems, Do2024MULTI-FIDELITYREVIEW}.

\paragraph{Specialised Kernel Models.}
Bayesian optimisation methods often incorporate fidelity through specialised kernels or augmented input spaces, including multi-information source kernels \citep{Poloczek2017Multi-informationOptimization}, fidelity-augmented GPs \citep{Kandasamy2017Multi-fidelityApproximations,Wu2020PracticalTuning}, and additive latent-function models \citep{Song2019AProcesses}. These methods are effective for optimisation, but typically rely on globally learned kernel structures, limiting their ability to adapt fidelity relationships or uncertainty estimates in sparse data regions \citep{Kandasamy2017Multi-fidelityApproximations}.

\paragraph{Fusion-Based and Reliability-Aware Methods.}
Fusion-based approaches relax strict fidelity hierarchy by combining independent surrogates, for example through precision-weighted averaging \citep{Lam2015MultifidelitySources} or correlation-aware model reification \citep{Thomison2017ASources}. Reliability-aware formulations treat fidelity differences as source-dependent uncertainty within unified probabilistic models \citep{Ghoreishi2018APolicies,Fan2024Multi-fidelityFidelity}. Although these approaches support non-nested information sources, they typically fuse distributions without explicitly using geometric proximity between fidelity samples to calibrate local discrepancies.

\section{Preliminaries}
\label{sec:background}

\paragraph{Gaussian Process Regression.}
A Gaussian process (GP) defines a distribution over functions such that any finite set of function values is jointly Gaussian \citep{Rasmussen2006GaussianLearning}. Placing a zero-mean GP prior $f(\mathbf{x}) \sim \mathcal{GP}\big(0,\, k(\mathbf{x}, \mathbf{x}')\big)$ with positive-semidefinite kernel $k$, and given noisy observations $\mathcal{D} = \{(\mathbf{x}_i, y_i)\}_{i=1}^N$ with $y_i = f(\mathbf{x}_i) + \varepsilon_i$, $\varepsilon_i \sim \mathcal{N}(0, \sigma^2)$, the posterior at any test input $\mathbf{x}_\star$ is Gaussian with
\[
    \mu(\mathbf{x}_\star) = \mathbf{k}_\star^\top (\mathbf{K} + \sigma^2 \mathbf{I})^{-1} \mathbf{y}, \qquad
    \sigma^2(\mathbf{x}_\star) = k(\mathbf{x}_\star, \mathbf{x}_\star) - \mathbf{k}_\star^\top (\mathbf{K} + \sigma^2 \mathbf{I})^{-1} \mathbf{k}_\star,
\]
where $\mathbf{K}$ is the training kernel matrix and $\mathbf{k}_\star$ collects the cross-covariances between $\mathbf{x}_\star$ and the training inputs.

\paragraph{Multi-fidelity Problem Setting.}
\label{sec:problem_setting}
We consider a budget-constrained multi-fidelity setting with $M$ fidelity levels indexed by $m \in \{1, \ldots, M\}$, with the highest fidelity at $m = M$. High-fidelity data are assumed to be sparse owing to their high evaluation cost, whereas lower-fidelity observations are more abundant. Evaluation costs satisfy $\mathcal{C}_1 < \mathcal{C}_2 < \cdots < \mathcal{C}_M$, establishing an economic hierarchy in which higher indices correspond to more expensive, and typically more accurate, fidelity levels. Subject to a total computational budget $\mathcal{B}$, each fidelity contributes a dataset
\begin{equation}
    \mathcal{D}_m = \big\{(\mathbf{x}_i^{(m)},\, y_i^{(m)})\big\}_{i=1}^{N_m}, \qquad m = 1, \ldots, M,
    \label{eq:mf_data}
\end{equation}
with $y_i^{(m)} = f_m(\mathbf{x}_i^{(m)}) + \varepsilon_i^{(m)}$ and Gaussian observation noise $\varepsilon_i^{(m)} \sim \mathcal{N}(0, \sigma_m^2)$. The goal is to approximate the highest-fidelity function $f_M(\mathbf{x})$ as accurately as possible by leveraging all $M$ data sources jointly.
\section{MAST Framework}
\label{sec:methodology}

The MAST framework is built on a simple intuition: the reliability of corrected low-fidelity information should depend on its proximity to high-fidelity calibration data. Near high-fidelity samples, locally calibrated high-fidelity predictions should dominate, while low-fidelity contributions are down-weighted. Far from high-fidelity samples, high-fidelity surrogates are necessarily extrapolative and more uncertain, whereas corrected low-fidelity observations can reveal local response structure with propagated uncertainty. MAST implements this through three stages: independent GP training, discrepancy-based correction with spatially weighted blending, and heteroscedastic GP fusion. We motivate the framework for the general $M$-fidelity case; Algorithm~\ref{alg:mast_compact} summarises the procedure, and the full pseudocode and pipeline schematic are in Appendix~\ref{app:algorithm_app}.

\begin{algorithm}[!t]
\caption{Multi-fidelity Augmented Surrogate with Spatial Trust-weighting (MAST)}
\label{alg:mast_compact}
\begin{algorithmic}[1]
\State \textbf{Input:} Multi-fidelity data $\{\mathcal{D}_m\}_{m=1}^{M}$, cost ratios $\{\mathcal{C}_M / \mathcal{C}_m\}$
\State Train independent vanilla $\mathcal{GP}_m$ on each $\mathcal{D}_m$
\For{$m = 1,2,\dots,M-1$}
    \State Compute residuals $\delta_j = y_j^{(M)} - \mu_m(\mathbf{x}_j^{(M)})$
    \State Fit discrepancy $\mathcal{GP}_{\delta_m}$ to augmented inputs $\mathbf{z}_j^{(m)} = \bigl[\,\mathbf{x}_j^{(M)},\; \mu_m(\mathbf{x}_j^{(M)}),\; \sigma_m^2(\mathbf{x}_j^{(M)})\,\bigr]$
    \For{$i = 1,\dots, |\mathcal{D}_m|$}
        \State Compute local trust region $\mathcal{N}_i$ and weight $W_m^{(i)}$ for $\mathbf{x}_i^{(m)}$
        \State Calculate corrected observations $\tilde{y}_i$ via Eq.~\eqref{eq:augmented_mean} and propagate variance $\tilde{\sigma}_i^2$ via Eq.~\eqref{eq:augmented_var}
    \EndFor
\EndFor
\State Train final $\mathcal{GP}$ on $\mathcal{D}_{\mathrm{aug}} = \mathcal{D}_M \cup \bigcup_{m=1}^{M-1} \tilde{\mathcal{D}}_m$ with fixed per-point variances and input warping
\end{algorithmic}
\end{algorithm}

\paragraph{Independent Baseline Modelling.}
We first train independent vanilla GPs with inferred noise as baseline surrogates for each fidelity level to capture their intrinsic characteristics and noise levels. A GP prior is placed on each function $f_m \sim \mathcal{GP}(0, k_m)$ for $m \in \{1, \ldots, M\}$, using an RBF kernel. The hyperparameters $\{\boldsymbol{\ell}_m, \sigma_{f,m}^2\}$ and the observation noise variances $\sigma_m^2$ are estimated via Type-II Maximum Likelihood Estimation (MLE). The learned noise levels $\hat{\sigma}_m$ serve as estimates of the intrinsic uncertainty at each fidelity.

\paragraph{Discrepancy Modelling and Low-Fidelity Conditioning.}
\label{sec:stage2}
The central contribution of MAST is the augmentation scheme that first corrects each lower-fidelity observation using a learned discrepancy function, then blends the corrected value with the high-fidelity GP prediction using spatially-adaptive weights. For each fidelity level $m < M$, the procedure transforms $\mathcal{D}_m$ into an augmented dataset $\tilde{\mathcal{D}}_m$. We model the discrepancy between fidelity level $m$ and the highest fidelity $M$ by fitting a Gaussian process to the residuals at the high-fidelity observation locations:
\begin{equation}
    \delta_m(\mathbf{x}_j^{(M)}) = y_j^{(M)} - \mu_m(\mathbf{x}_j^{(M)}), \quad j = 1, \ldots, N_M,
    \label{eq:disc}
\end{equation}
where $\mu_m(\cdot)$ is the posterior mean of the fidelity-$m$ GP. Rather than fitting $\mathcal{GP}_{\delta_m}$ as a function of location alone, we condition it on the local low-fidelity posterior summaries by forming the augmented input
\begin{equation}
    \mathbf{z}_j^{(m)} = \bigl[\,\mathbf{x}_j^{(M)},\; \mu_m(\mathbf{x}_j^{(M)}),\; \sigma_m^2(\mathbf{x}_j^{(M)})\,\bigr] \in \mathbb{R}^{D+2},
    \label{eq:lf_augmented_input}
\end{equation}
and fitting the discrepancy GP on $\{(\mathbf{z}_j^{(m)},\delta_j)\}_{j=1}^{N_M}$ with an ARD-RBF kernel over the full $(D+2)$-dimensional space. The lengthscales on the two distributional features are learned jointly with the spatial lengthscales, so the model decides how informative the local low-fidelity posterior is for predicting discrepancy. Unlike nonlinear autoregressive GPs \citep{Perdikaris2017NonlinearModelling}, which augment the discrepancy input only with the LF posterior mean, we condition on both mean and posterior variance, exposing local LF calibration to the kernel. At any LF input $\mathbf{x}_i^{(m)}$ we evaluate $\mathcal{GP}_m$ at that point to construct $\mathbf{z}_i^{(m)}$ and query the discrepancy GP on the resulting augmented point. Appendix~\ref{app:lf_conditioning_justification} provides the theoretical justification.

\paragraph{Spatially-Adaptive Weighting.}
For each low-fidelity point $\mathbf{x}_i^{(m)}$, we compute a weight that determines the relative contribution of the corrected low-fidelity value versus the high-fidelity GP prediction. This weight depends on the spatial proximity to high-fidelity observations: points near high-fidelity samples should favour high-fidelity predictions, while distant points should favour corrected low-fidelity values. We first compute the Euclidean distance from $\mathbf{x}_i^{(m)}$ to each high-fidelity point in the normalised input space:
\begin{equation}
    d_{ij} = \|\tilde{\mathbf{x}}_i^{(m)} - \tilde{\mathbf{x}}_j^{(M)}\|_2, \quad j = 1, \ldots, N_M.
\end{equation}
Euclidean distance is the natural default for continuous inputs, but nothing in the construction depends on it. Any metric appropriate to the input space can be substituted without changing the rest of the pipeline: Euclidean on continuous inputs, Hamming on categorical inputs (as in Appendix~\ref{app:rw_hoip} for HOIP), or Mahalanobis on anisotropic spaces. Sensitivity to the metric choice is reported in Appendix~\ref{app:ablation_distance}.
A naive approach would aggregate contributions from all high-fidelity points, but this leads to problematic behaviour: a low-fidelity point with a nearby high-fidelity neighbour would still be influenced by distant high-fidelity observations elsewhere in the domain. To ensure that only \emph{locally relevant} high-fidelity information contributes, we introduce an {adaptive trust region} based on an {$N$-sphere} centred at $\mathbf{x}_i^{(m)}$. The radius of this trust region should scale with the distance to the nearest high-fidelity point, i.e., $r_i = g(d_{\min})$. In principle, any monotonically increasing function $g$ could be used; we adopt:
\begin{equation}
    r_i = \sqrt{d_{\min}}, \quad \text{where} \quad d_{\min} = \min_j d_{ij},
    \label{eq:nsphere_radius}
\end{equation}
which we found empirically to provide robust performance. Only high-fidelity points within this $N$-sphere contribute to the weight computation: $\mathcal{N}_i = \{j : d_{ij} \leq r_i\}$.

For each high-fidelity point $j \in \mathcal{N}_i$, we compute a distance-based weight:
\begin{equation}
    w_j = 1 - d_{ij}^{\,\alpha_m}, \quad \text{where} \quad \alpha_m = \frac{\log_{10}(\mathcal{C}_M/\mathcal{C}_m)}{2}.
    \label{eq:individual_weight}
\end{equation}
The exponent $\alpha_m$ controls how quickly the weight decays with distance. Critically, $\alpha_m$ encodes the fidelity hierarchy, determining how much spatial influence each fidelity level exerts. In principle, $\alpha_m$ can be derived from any quantity that establishes an ordering among fidelities: evaluation cost ratios, user-specified confidence levels, known approximation orders, or domain expertise. In this work, we define $\alpha_m$ as a function of the fidelity cost ratio $\mathcal{C}_M/\mathcal{C}_m$, explicitly incorporating evaluation cost into the weighting scheme.
Alternative weight functions are also possible (e.g., exponential decay) provided $w_j = g(d_{ij}; \alpha_m)$ where $g(\cdot)$ is monotonically decreasing in distance; in our experiments the form in Eq.~\eqref{eq:individual_weight} performed best. Appendix~\ref{app:ablation} sweeps the trust-region radius, weight function, and distance metric to document the sensitivity of MAST to these choices, including alternatives to Euclidean distance that mitigate high-dimensional pathology.

The individual weights are aggregated into a single low-fidelity weight $W^{(i)}_m$ for each point $\mathbf{x}^{(m)}_i$:
\begin{equation}
    W^{(i)}_m = 1 - \frac{1}{|\mathcal{N}_i|} \sum_{j \in \mathcal{N}_i} w_j,
    \label{eq:aggregate_weight}
\end{equation}
By construction, $W^{(i)}_m$ decreases with proximity to high-fidelity observations. This locality mechanism ensures that only nearby high-fidelity observations influence the weighting.

\paragraph{Corrected Observations and Variance Propagation.}
Each low-fidelity observation is corrected by adding the predicted discrepancy, then blended with the high-fidelity GP prediction, with the low-fidelity weight $W^{(i)}_m$ determining the contribution of each component:
\begin{equation}
    \tilde{y}^{(m)}_i = W^{(i)}_m \cdot \left( y^{(m)}_i + \mu_{\delta_m}(\mathbf{z}^{(m)}_i) \right) + (1 - W^{(i)}_m) \cdot \mu_M(x^{(m)}_i).
    \label{eq:augmented_mean}
\end{equation}

Distributing the weights in Eq.~\eqref{eq:augmented_mean} decomposes the predictive error on $\tilde y_i^{(m)}$ into three independent uncertainty sources,
\begin{equation}
    \tilde{y}^{(m)}_i = \underbrace{W^{(i)}_m\, y^{(m)}_i}_{T_1:\ \text{weighted raw LF}} + \underbrace{W^{(i)}_m\, \mu_{\delta_m}(\mathbf{z}^{(m)}_i)}_{T_2:\ \text{weighted discrepancy}} + \underbrace{(1 - W^{(i)}_m)\, \mu_M(\mathbf{x}^{(m)}_i)}_{T_3:\ \text{weighted HF posterior}},
\end{equation}
each with its own variance: $\operatorname{Var}(T_1) = (W^{(i)}_m)^2 \hat\sigma_m^2$ from the raw-observation noise, $\operatorname{Var}(T_2) = (W^{(i)}_m)^2 \sigma_{\delta_m}^2(\mathbf{x}^{(m)}_i)$ from the discrepancy GP's posterior, and $\operatorname{Var}(T_3) = (1 - W^{(i)}_m)^2 \sigma_M^2(\mathbf{x}^{(m)}_i)$ from the HF GP's posterior. Conditional on the training data, the three contributions are mutually independent (two exactly and one under a standard multi-fidelity simplification, see Appendix~\ref{app:variance_derivation} for the full first-principles derivation), so the cross-covariances vanish and
\begin{equation}
        \tilde{\sigma}^2_i = \left(W^{(i)}_m\right)^2 \left(\hat{\sigma}^2_m + \sigma^2_{\delta_m}(\mathbf{x}^{(m)}_i)\right) + \left(1 - W^{(i)}_m\right)^2 \sigma^2_M(\mathbf{x}^{(m)}_i).
    \label{eq:augmented_var}
\end{equation}
This formulation ensures that corrected points near high-fidelity data (where $W^{(i)}_m \to 0$) inherit the lower uncertainty of high-fidelity predictions, while points far from high-fidelity observations (where $W^{(i)}_m \to 1$) retain the combined uncertainty of the original measurement and discrepancy estimation.

\paragraph{Gaussian Process Fusion with Fixed Observation Variances.}
\label{sec:stage3}
The final surrogate is trained on the combined dataset consisting of original high-fidelity observations and augmented lower-fidelity observations from all levels:
\begin{equation}
    \mathcal{D}_{\mathrm{aug}} = \bigl\{(\mathbf{x}_j^{(M)}, y_j^{(M)}, \hat{\sigma}_M^2)\bigr\}_{j=1}^{N_M} \bigcup_{m=1}^{M-1} \bigl\{(\mathbf{x}_i^{(m)}, \tilde{y}_i^{(m)}, \tilde{\sigma}_i^2)\bigr\}_{i=1}^{N_m}.
\end{equation}
The final GP is fitted with each observation's noise variance fixed at the value computed via Eq.~\eqref{eq:augmented_var}, so the surrogate weights every point by its pre-determined reliability and naturally down-weights uncertain augmented data while trusting high-fidelity observations. The kernel operates on warped coordinates $\tilde{\mathbf{x}} = \Phi(\mathbf{x};\boldsymbol{a},\boldsymbol{b})$, where $\Phi$ is a per-dimension Kumaraswamy cumulative distribution function on $[0,1]^D$ with shape parameters $(\boldsymbol{a},\boldsymbol{b})$ learned jointly with the kernel hyperparameters \citep{Snoek2014InputWarping}. This handles non-stationarity while keeping the kernel form stationary in the warped space.

\section{Benchmarks}
\label{sec:syn}

\paragraph{Baseline Algorithms.} In our empirical evaluation, we benchmark MAST against representative methods from each category: the autoregressive co-kriging model \citep{Kennedy2000PredictingAvailable}, recursive co-kriging \citep{LeGratiet2014RecursiveFidelity}, multi-information source precision-weighted fusion \citep{Lam2015MultifidelitySources}, model reification-based fusion \citep{Thomison2017ASources}, nonlinear autoregressive GPs (NARGP) \citep{Perdikaris2017NonlinearModelling}, deep multi-fidelity GPs (MF-DGP) \citep{Cutajar2019DeepModeling}, and BoTorch's specialised-kernel multi-fidelity GP \citep{Poloczek2017Multi-informationOptimization, Wu2020PracticalTuning}. We also include a multi-fidelity Bayesian neural network (MFBNN) \citep{Yi2024MFBML} that fuses a deterministic low-fidelity DNN with a Bayesian residual network.

\paragraph{Test Functions.} We benchmark MAST on a suite of ten standard test functions spanning 2D--20D (full specifications in Appendix~\ref{app:test_functions}), complemented by real-world engineering benchmarks of up to 23D described in Appendix~\ref{app:rw_datasets}. All synthetic functions exhibit spatially-varying inter-fidelity discrepancy, except Ackley (noise-only) and Park2 (linear scaling). Inter-fidelity correlation is controlled through a degradation-based formulation,
\begin{equation}
    f_{LF}(\mathbf{x}) = f_{HF}(\mathbf{x}) + d \cdot \delta(\mathbf{x}),
    \label{eq:cost_controlled}
\end{equation}
where $\delta(\mathbf{x})$ is a function-specific deviation term and $d \geq 0$ governs the severity of the fidelity discrepancy.

\paragraph{Test Details.} The total budget is $\mathcal{B} = 5D$ in HF-equivalent evaluations, with cost hierarchy $\mathcal{C}_{\mathrm{HF}}{:}\mathcal{C}_{\mathrm{MF}}{:}\mathcal{C}_{\mathrm{LF}} = 1.0{:}0.2{:}0.1$ following \citet{Vaiuso2025Multi-fidelityLearning}, allocated $70/30$ (HF/LF) in two-fidelity and $50/30/20$ (HF/MF/LF) in three-fidelity experiments. Each configuration is repeated for 25 random seeds with Latin hypercube training designs at each fidelity and fixed per-function test sets.

\paragraph{Performance Metrics.} Predictive accuracy is reported via RMSE, uncertainty calibration via the mean predictive density (mean~PDF), and goodness of fit via $R^2$ as a secondary diagnostic; all entries are normalised by the cost-equivalent HF-only baseline so RMSE values below~$1$ and mean-PDF values above~$1$ indicate improvement, and table cells report the mean~$\pm$~std of the per-seed paired ratio across the 25 repetitions. Per-function observation-noise specifications, baseline-specific implementation details (kernels, hyperparameter optimisation, deep-learning baselines), and formal metric definitions are deferred to Appendix~\ref{app:benchmarking}.

\section{Results and Discussion}
\label{sec:results}
\label{sec:discussion}
\subsection{Two-Fidelity Results}

Table~\ref{tab:results_combined} presents the normalised performance across ten two-fidelity synthetic benchmarks and the two real-world engineering benchmarks (Concrete and DrivAerNet), where RMSE values below unity and mean PDF values above unity indicate improvement over the HF-only baseline. Real-world benchmarks use the independent-sampling configuration in which HF and LF training points are drawn separately at each fidelity; the collocated-sampling counterpart for DrivAerNet is reported in Appendix~\ref{app:collocated_comparison}.

MAST achieves the best RMSE on 6 out of 12 and the best Mean PDF on 7 out of 12 benchmark functions, demonstrating consistent performance across diverse problem landscapes. On functions where MAST does not achieve the best result, it remains competitive: for instance, on Hartmann3 and Park2 where AR1 excels due to strong global linear correlation, MAST still achieves meaningful improvements. Importantly, it avoids severe miscalibrations on any of the benchmarks as seen in baselines such as recursive co-kriging (e.g., Park1) or MF-DGP (e.g., Rosenbrock).

Among competitors, AR1 co-kriging performs well on low-dimensional problems where inter-fidelity correlation is spatially constant (e.g., Hartmann3, Park2) but degrades sharply otherwise. BoTorch MF-GP shows consistent modest improvements on smooth functions (best on Branin, Hartmann6, Rastrigin). The deep learning approaches (NARGP, MF-DGP) consistently underperform, suggesting insufficient data under budget constraints ($\mathcal{B} = 5D$).

\begin{table*}[!t]
\centering
\caption{Two-fidelity benchmarks: surrogate model performance comparison. Each cell shows the mean of the per-seed paired ratio (metric / HF-only baseline) across 25 seeds. RMSE values $<1$ and Mean PDF $>1$ indicate improvement. Best per row and metric in \textbf{bold}. Per-cell standard deviations of the same paired ratios are in Tables~\ref{tab:results_rmse} (RMSE) and~\ref{tab:results_pdf} (Mean PDF) (Appendix~\ref{app:pdf_tables}).}
\label{tab:results_combined}
\resizebox{\textwidth}{!}{%
\begin{tabular}{lccccccccc|cccccccccc}
\toprule
& \multicolumn{9}{c}{\textbf{Normalised RMSE} $\downarrow$} & & \multicolumn{9}{c}{\textbf{Normalised Mean PDF} $\uparrow$} \\
\cmidrule(lr){2-10} \cmidrule(lr){12-20}
Test Function & MAST & AR1 & Recur. & Reif. & BoTorch & NARGP & MF-DGP & MISO & MFBNN & &
MAST & AR1 & Recur. & Reif. & BoTorch & NARGP & MF-DGP & MISO & MFBNN \\
\midrule
Branin (2D)      & 1.19 & 2.06 & 6.03 & 2.24 & \textbf{0.95} & 1.67 & 1.86 & 2.23 & 1.18 & & 1.28 & 0.63 & 0.62 & 1.37 & \textbf{1.43} & 0.89 & 0.91 & 1.35 & 1.05 \\
Hartmann3 (3D)   & 0.69 & \textbf{0.29} & 1.29 & 0.41 & 0.29 & 0.48 & 0.92 & 0.42 & 1.07 & & 1.83 & \textbf{3.97} & 0.72 & 2.52 & 2.48 & 2.45 & 0.94 & 2.35 & 1.08 \\
Ackley (4D)      & \textbf{0.94} & 1.63 & 1.19 & 1.19 & 1.02 & 1.59 & 1.57 & 1.11 & 1.62 & & \textbf{1.23} & 0.75 & 0.92 & 0.96 & 0.83 & 0.69 & 0.65 & 1.03 & 0.72 \\
Park1 (4D)       & \textbf{0.97} & 1.15 & 4.99 & 2.46 & 1.06 & 1.24 & 3.00 & 2.31 & 1.25 & & \textbf{1.46} & 1.38 & 0.48 & 0.37 & 0.72 & 1.35 & 0.26 & 0.42 & 0.87 \\
Park2 (4D)       & 0.64 & \textbf{0.35} & 1.46 & 2.82 & 0.51 & 1.31 & 1.35 & 2.75 & 0.82 & & 1.59 & \textbf{3.01} & 0.67 & 0.14 & 1.28 & 0.92 & 0.65 & 0.11 & 0.74 \\
Hartmann6 (6D)   & 0.88 & 0.67 & 1.00 & 0.84 & \textbf{0.64} & 0.93 & 1.09 & 0.81 & 1.02 & & \textbf{1.80} & 1.50 & 1.06 & 1.42 & 1.17 & 1.30 & 0.89 & 1.45 & 1.14 \\
Levy (7D)        & \textbf{0.78} & 1.03 & 1.31 & 0.87 & 0.81 & 1.10 & 1.92 & 0.85 & 1.15 & & \textbf{1.43} & 0.94 & 1.03 & 1.25 & 1.10 & 0.87 & 0.65 & 1.31 & 1.06 \\
Borehole (8D)    & \textbf{0.88} & 3.47 & 5.83 & 5.69 & 1.48 & 1.80 & 2.99 & 5.58 & 0.98 & & \textbf{1.58} & 1.00 & 0.78 & 0.04 & 0.65 & 0.81 & 0.39 & 0.05 & 1.22 \\
Rastrigin (15D)  & 0.95 & 1.06 & 0.91 & 0.96 & \textbf{0.83} & 1.29 & 1.83 & 0.91 & 1.67 & & 1.18 & 1.05 & 1.06 & 1.09 & 1.06 & 0.85 & 0.63 & \textbf{1.20} & 0.64 \\
Rosenbrock (20D) & \textbf{0.37} & 1.67 & 0.82 & 0.80 & 0.79 & 4.76 & 4.76 & 0.81 & 1.51 & & \textbf{3.23} & 0.39 & 1.27 & 1.38 & 1.14 & 0.17 & 0.00 & 1.40 & 0.84 \\
\midrule
Concrete (8D)    & \textbf{0.89} & 1.03 & 1.27 & 1.77 & 0.93 & 0.96 & 1.12 & 1.77 & 0.94 & & \textbf{1.27} & 0.94 & 0.84 & 0.68 & 1.11 & 1.18 & 0.88 & 0.68 & 1.13 \\
DrivAerNet (23D) & 1.17 & 1.24 & 2.59 & \textbf{0.97} & 1.37 & 0.99 & 0.99 & 0.97 & 7.29 & & 0.84 & \textbf{0.99} & 0.30 & 0.60 & 0.42 & 0.97 & 0.87 & 0.60 & 0.09 \\
\bottomrule
\end{tabular}%
}
\end{table*}

\subsection{Three-Fidelity Results}
\label{sec:res_three_fid}

The synthetic benchmarks are extended to three fidelities by inserting a mid-fidelity level between the existing HF and LF sources, with cost hierarchy $\mathcal{C}_{\mathrm{HF}} : \mathcal{C}_{\mathrm{MF}} : \mathcal{C}_{\mathrm{LF}} = 1.0 : 0.2 : 0.1$ and budget allocation $50\%/30\%/20\%$ across HF/MF/LF; the mid-fidelity uses the cost-controlled construction of Eq.~\eqref{eq:cost_controlled} at $d=0.5$. In addition, the HOIP perovskite band-gap dataset is used as the real-world three-fidelity benchmark (Appendix~\ref{app:rw_hoip}). Table~\ref{tab:three_fid_combined} reports normalised RMSE and normalised mean PDF on ten synthetic benchmarks plus HOIP; normalised $R^2$ for the same suite is in Appendix~\ref{app:r2_extra}.

\begin{table*}[!t]
\centering
\caption{Three-fidelity benchmarks: surrogate model performance comparison. Each cell shows the mean of the per-seed paired ratio (metric / HF-only baseline) across 25 seeds. RMSE values $<1$ and Mean PDF $>1$ indicate improvement. Best per row and metric in \textbf{bold}. Per-cell standard deviations of the same paired ratios are in Tables~\ref{tab:three_fid_rmse} (RMSE) and~\ref{tab:three_fid_pdf} (Mean PDF) (Appendix~\ref{app:pdf_tables}).}
\label{tab:three_fid_combined}
\resizebox{\textwidth}{!}{%
\begin{tabular}{lccccccccc|cccccccccc}
\toprule
& \multicolumn{9}{c}{\textbf{Normalised RMSE} $\downarrow$} & & \multicolumn{9}{c}{\textbf{Normalised Mean PDF} $\uparrow$} \\
\cmidrule(lr){2-10} \cmidrule(lr){12-20}
Test Function & MAST & AR1 & Recur. & Reif. & BoTorch & NARGP & MF-DGP & MISO & MFBNN & & MAST & AR1 & Recur. & Reif. & BoTorch & NARGP & MF-DGP & MISO & MFBNN \\
\midrule
Branin (2D)      & 1.06 & 2.04 & 1.96 & 1.83 & \textbf{0.64} & 1.91 & 1.25 & 1.78 & 1.37 & & 1.24 & 0.68 & 0.63 & 1.12 & \textbf{1.55} & 0.43 & 0.95 & 1.20 & 0.91 \\
Hartmann3 (3D)   & 0.76 & 0.55 & 1.06 & 9.01 & \textbf{0.30} & 0.97 & 0.97 & 0.58 & 1.42 & & 1.60 & 2.24 & 0.85 & 1.61 & \textbf{2.28} & 1.27 & 0.63 & 1.85 & 0.80 \\
Ackley (4D)      & \textbf{0.84} & 1.66 & 1.11 & 2.08 & 0.96 & 1.35 & 1.60 & 0.98 & 1.66 & & \textbf{1.37} & 0.77 & 0.91 & 1.05 & 0.82 & 0.78 & 0.65 & 1.19 & 0.68 \\
Park1 (4D)       & \textbf{1.15} & 1.17 & 2.10 & 3.68 & 1.23 & 1.57 & 2.20 & 1.84 & 2.80 & & \textbf{1.21} & 1.09 & 0.59 & 0.54 & 0.61 & 0.92 & 0.23 & 0.53 & 0.43 \\
Park2 (4D)       & 0.77 & 0.64 & 0.93 & 2.43 & \textbf{0.62} & 1.12 & 1.34 & 2.44 & 1.43 & & 1.37 & \textbf{1.82} & 1.07 & 0.21 & 1.09 & 1.17 & 0.43 & 0.13 & 0.56 \\
Hartmann6 (6D)   & 0.91 & 0.93 & 1.00 & 1.82 & \textbf{0.65} & 1.19 & 1.22 & 0.87 & 1.17 & & 1.61 & 1.30 & 1.11 & \textbf{2.30} & 1.21 & 1.07 & 0.83 & 1.53 & 1.03 \\
Levy (7D)        & \textbf{0.75} & 1.38 & 1.25 & 1.33 & 0.81 & 1.13 & 2.27 & 0.86 & 1.32 & & \textbf{1.49} & 0.86 & 0.91 & 1.18 & 1.10 & 0.83 & 0.65 & 1.32 & 0.93 \\
Borehole (8D)    & \textbf{1.27} & 3.48 & 4.79 & 4.38 & 1.60 & 1.40 & 2.71 & 4.20 & 2.26 & & \textbf{1.31} & 0.71 & 0.57 & 0.10 & 0.58 & 0.85 & 0.36 & 0.10 & 0.51 \\
Rastrigin (15D)  & 0.93 & 1.02 & 0.96 & 0.94 & \textbf{0.82} & 1.50 & 2.32 & 0.88 & 1.63 & & 1.23 & 1.08 & 1.04 & 1.15 & 1.09 & 0.75 & 0.55 & \textbf{1.30} & 0.66 \\
Rosenbrock (20D) & \textbf{0.36} & 1.53 & 0.90 & 0.91 & 0.82 & 4.75 & 4.75 & 0.85 & 1.73 & & \textbf{3.33} & 0.32 & 1.11 & 1.28 & 1.13 & 0.18 & 0.16 & 1.39 & 0.74 \\
\midrule
HOIP (3D)        & \textbf{1.06} & 2.37 & 1.10 & 2.06 & 1.43 & 1.18 & 1.12 & 2.06 & 1.75 & & \textbf{1.12} & 0.80 & 0.89 & 0.64 & 0.66 & 0.84 & 0.79 & 0.64 & 0.64 \\
\bottomrule
\end{tabular}%
}
\end{table*}

MAST achieves the best RMSE and Mean PDF on 6 out of 11 functions. Performance is consistent with the two-fidelity case, demonstrating that MAST's augmentation strategy effectively incorporates additional fidelity levels without loss of robustness. For all test functions where MAST is not best, it maintains competitive performance. Methods relying on strict hierarchical assumptions continue to struggle on multimodal and higher-dimensional problems, while BoTorch retains competitive performance primarily on low-dimensional smooth functions. Surprisingly, AR1's performance degrades with the addition of an intermediate fidelity level. Deep-learning methods (NARGP, MF-DGP) do not show any improvement on three-fidelities due to the constrained budget.

MAST demonstrates robustness across the benchmarking suite, compared to baseline cases which can suffer from degradation on some benchmarks. For example, Reification-based fusion exhibits a miscalibration on Hartmann3 (RMSE $9.01$, $R^2 \approx -3.12\!\times\!10^{2}$) where the precision-weighted fusion produces a grossly over-confident prediction. MAST processes each lower-fidelity dataset independently toward the highest fidelity (Appendix~\ref{app:algorithm_app}), so adding the mid-fidelity level does not compound errors along the hierarchy.

\subsection{Sensitivity to Budget Allocation}

\begin{figure}[H]
\centering
\includegraphics[width=0.9\textwidth]{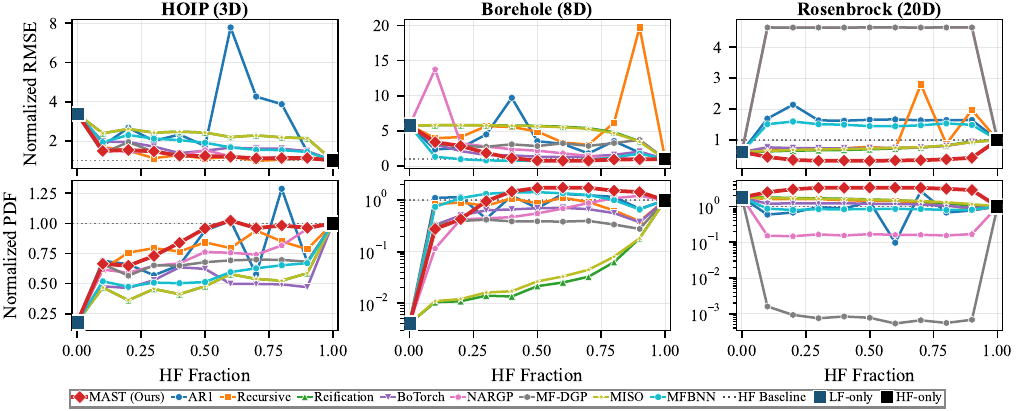}
\caption{Sensitivity to budget allocation. Top row: normalised RMSE; bottom row: normalised mean PDF on HOIP (3D), Borehole (8D), and Rosenbrock (20D), each divided by the cost-equivalent HF-only baseline (dotted line at 1.0). Full sweeps in Appendix~\ref{app:pa_combined}.}
\label{fig:ratio_sweep}
\end{figure}

To assess robustness to budget allocation choices, the HF cost fraction was varied at fixed total budget $\mathcal{B} = 5D$. Figure~\ref{fig:ratio_sweep} shows normalised performance across three representative functions. MAST maintains consistently strong performance across the entire allocation range, with substantial improvements even at HF fraction $=0.1$ where competitors degrade or become unstable. AR1 exhibits erratic behaviour, while NARGP and MF-DGP consistently underperform the baseline. The uncertainty calibration is particularly striking: MAST's mean PDF stays stable while other methods cluster near or below baseline. Quantitatively, MAST's per-method consistency rate (the seed-level fraction of interior sweep-points beating the cost-equivalent HF-only baseline) reaches 77\% on uncertainty calibration, leading the next-best calibration method by 23 percentage points across the sweep.

\subsection{Sensitivity to Total Budget}

\begin{figure}[H]
\centering
\includegraphics[width=0.9\textwidth]{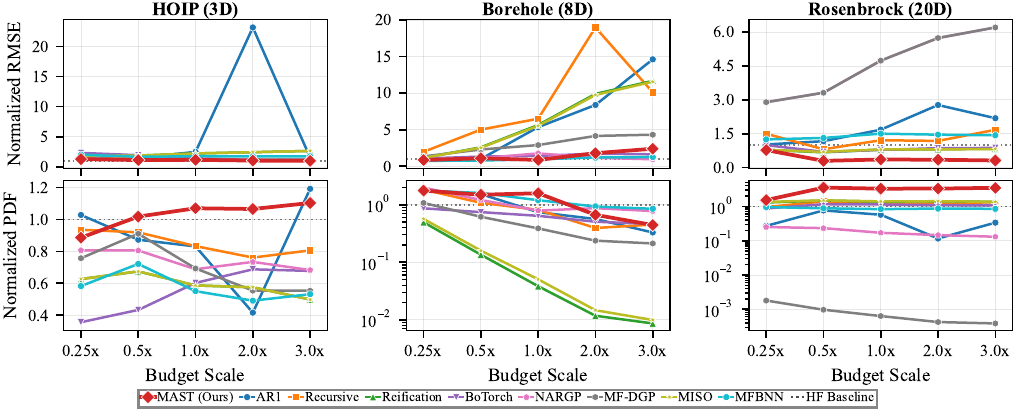}
\caption{Sensitivity to total budget. Axes as in Figure~\ref{fig:ratio_sweep}; total budget rescales by $\{0.5,1,2,3\}\times\mathcal{B}$.}
\label{fig:budget_scaling}
\end{figure}

To evaluate scaling with computational resources, the total budget was varied from $0.25\times$ to $3.0\times$ the base budget $\mathcal{B}=5D$ at fixed 70\%/30\% HF/LF cost split. Figure~\ref{fig:budget_scaling} shows normalised performance as budget scales. MAST stays stable across budget scales, particularly in data-scarce regimes; AR1 exhibits high variance, while MF-DGP remains at $2\times$ worse than baseline throughout. Even at $3\times$ budget, the data-hungry methods (NARGP, MF-DGP) fail to match MAST's performance at $0.25\times$ budget. The consistency rates reflect this robustness: MAST attains 76\% calibration consistency across budget scales, while every competitor falls below 50\%.

\subsection{Sensitivity to Fidelity Discrepancy}

\begin{figure}[H]
\centering
\includegraphics[width=0.9\textwidth]{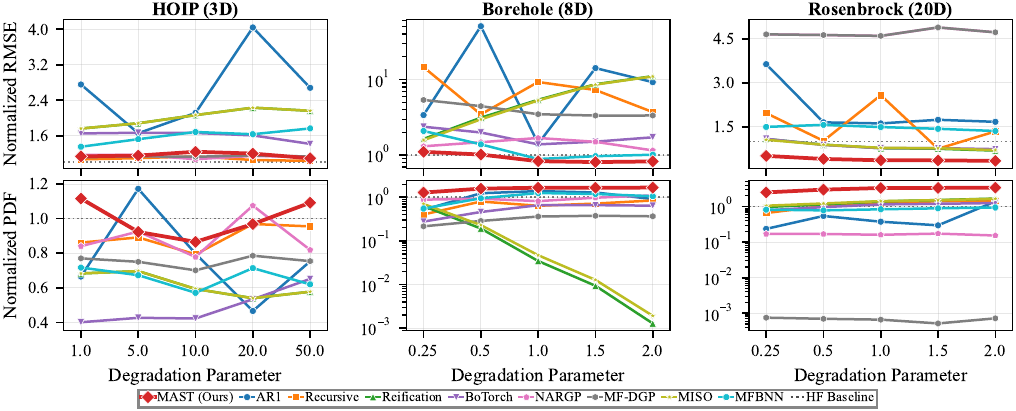}
\caption{Sensitivity to fidelity discrepancy. Axes as in Figure~\ref{fig:ratio_sweep}; synthetic $d$ of Eq.~\eqref{eq:cost_controlled} varies $0.25{-}2.0$, HOIP LF-noise cost-ratio $\mathcal{C}$ of Eq.~\eqref{eq:rw_pa_noise} varies $1.0{-}50.0$.}
\label{fig:cost_sweep}
\end{figure}

To assess robustness to varying inter-fidelity correlation, the degradation parameter $d$ in Eq.~\eqref{eq:cost_controlled} was swept from 0.25 (high correlation) to 2.0 (low correlation). Figure~\ref{fig:cost_sweep} shows normalised performance across three functions. MAST maintains strong RMSE reductions across the entire range of fidelity discrepancies, while most competing methods hover near or above baseline; AR1 degrades with increasing discrepancy on two benchmarks. MAST achieves this robustness by explicitly encoding the fidelity gap through its cost-aware weighting exponent rather than learning correlations that may not generalise. On the cost-controlled benchmarks, MAST exceeds the calibration baseline at 19/19 interior sweep-points on Rosenbrock and 14/19 on Borehole, with a mean calibration consistency rate of 80\% across the discrepancy sweep.

\paragraph{Ablation of MAST's heuristic components.} The default heuristics (square-root radius, cost-aware power-law weight, Euclidean distance) are sufficient to motivate the spatial-proximity mechanism on their own, and Appendix~\ref{app:ablation} shows that targeted variants can improve MAST further on landscapes where the default is structurally mismatched.

Two failure regimes account for every deviation: globally-structured discrepancy (Hartmann3, Hartmann6, Park2 --- where AR1 or BoTorch MF-GP is structurally matched and is best by construction) and moderate inter-fidelity correlation under independent sampling (DrivAerNet --- where the gap reverses under collocation, Appendix~\ref{app:collocated_comparison}). Even under failure, MAST is bounded: excluding the data-collapsed corners ($h_f \le 0.2$), its worst-case RMSE deviation across all 13 benchmarks and three sweeps is 2.04$\times$ HF-only, whereas competitor worst-cases exceed 7$\times$ (MISO), 8$\times$ (Reification), and 12$\times$ (NARGP, MF-DGP).

\paragraph{Limitations.}
We note three limitations. First, MAST inherits the cubic complexity of GP regression at each of its three stages, which becomes prohibitive beyond a few thousand HF points. Second, the cost-aware trust parameter $\alpha_m = \log_{10}(\mathcal{C}_M/\mathcal{C}_m)/2$ assumes known per-fidelity evaluation costs. Third, when the inter-fidelity correlation is near zero, the discrepancy GP extracts limited useful low-fidelity signal, and MAST's performance is comparable to mid-range baselines.

\section{Conclusion}
\label{sec:conclusion}
The results across the synthetic and real-world benchmarks support the central motivation of this work: when high-fidelity data are sparse and inter-fidelity relationships vary across the input space, relying solely on globally learned correlation structures can be limiting. In these settings, MAST provides an alternative inductive bias by using the geometric proximity of low-fidelity samples to high-fidelity calibration data to determine how strongly corrected low-fidelity information should contribute. This spatial trust mechanism helps reduce reliance on long-range extrapolation from sparse high-fidelity observations and leads to stable performance across different budget allocations, total evaluation budgets, and fidelity gaps.

Methodologically, MAST relaxes two common restrictions in multi-fidelity Gaussian-process modelling. First, rather than estimating a single global inter-fidelity correlation parameter or global cross-fidelity kernel structure, MAST assigns low-fidelity information a local contribution through proximity-based trust weighting. Second, unlike autoregressive and recursive formulations that often rely on nested sampling designs, MAST can incorporate independently sampled low-fidelity observations by treating each point as locally informative according to its spatial relationship with the available high-fidelity data. In doing so, the framework offers a practical route to spatially adaptive multi-fidelity fusion while retaining the simplicity of outputting a single heteroscedastic GP. These findings suggest that spatial trust weighting is a useful design principle for multi-fidelity surrogate modelling, particularly in sparse-data regimes where global correlation estimates may be misleading. Finally, while MAST is motivated with GPs in this work, the underlying spatial trust mechanism is surrogate‑agnostic, and an extension beyond GPs is outlined in Appendix~\ref{app:mast_icl}.

\clearpage

\bibliographystyle{unsrtnat}
\bibliography{referencesFixed}

\newpage
\appendix

\etocdepthtag.toc{app}
\etocsettagdepth{main}{none}
\etocsettagdepth{app}{subsection}
\etocsettocstyle
    {\section*{Appendix Contents}%
     \vspace{-4pt}\noindent\rule{\linewidth}{0.4pt}\par\vspace{6pt}}
    {\par\vspace{2pt}\noindent\rule{\linewidth}{0.4pt}\par\vspace{8pt}}
\tableofcontents

\section{Extended Methodology}
\label{app:algorithm_app}

\begin{figure}[h]
    \centering
    \includegraphics[trim={0cm 0.34cm 0cm 0cm}, clip,width=\linewidth]{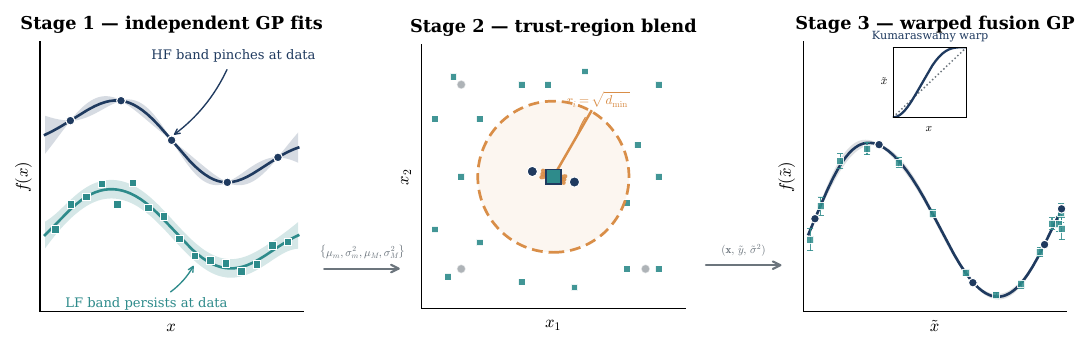}
    \caption{MAST pipeline. Stage~1: independent GPs at each fidelity. Stage~2: $N$-sphere trust region of radius $r_i = \sqrt{d_{\min}}$ around each LF query; HF samples inside the sphere contribute to $W_m^{(i)}$. Stage~3: fused GP on Kumaraswamy-warped inputs with per-point noise bars from the propagated Stage~2 variance.}
    \label{fig:mast_overview}
\end{figure}

This appendix provides a comprehensive description of the MAST algorithm, including the complete pseudocode and detailed implementation considerations for each stage. Figure~\ref{fig:mast_overview} gives an end-to-end schematic of the pipeline, and Algorithm~\ref{alg:mast_extended} below presents the full procedure.

\setcounter{algorithm}{0}
\begin{algorithm*}[h]
\caption{MAST: Multi-fidelity Augmented Surrogate with Trust-region Weighting (Extended)}
\label{alg:mast_extended}
\begin{algorithmic}[1]
\Require Data $\{\mathcal{D}_m\}_{m=1}^{M}$; Cost ratios $C_m = \mathcal{C}_M / \mathcal{C}_m$ for $m < M$
\Ensure Trained surrogate model $\mathcal{M}$
\State \textbf{Stage 1: Train independent GPs}
\State Normalise all inputs to $[0,1]^D$
\For{$m = 1$ to $M$}
    \State Train $\mathcal{GP}_m$ on $\mathcal{D}_m$ $\rightarrow$ obtain learned noise $\hat{\sigma}_m$
\EndFor
\State \textbf{Stage 2: Augment lower-fidelity data toward highest fidelity}
\For{$m = 1$ to $M-1$}
    \State \textit{Fit discrepancy GP on augmented inputs}
    \For{$j = 1$ to $N_M$}
        \State $\mu_m^{(j)}, \sigma_m^{2(j)} \leftarrow \mathcal{GP}_m(\mathbf{x}_j^{(M)})$ \Comment{LF posterior at HF input}
        \State $\delta_j \leftarrow y_j^{(M)} - \mu_m^{(j)}$ \Comment{Residuals at HF locations}
        \State $\mathbf{z}_j^{(m)} \leftarrow [\mathbf{x}_j^{(M)},\, \mu_m^{(j)},\, \sigma_m^{2(j)}]$ \Comment{Augmented input}
    \EndFor
    \State Train $\mathcal{GP}_{\delta_m}$ on $\{(\mathbf{z}_j^{(m)}, \delta_j)\}_{j=1}^{N_M}$ with ARD-RBF kernel

    \State \textit{Correct and blend each point}
    \State Compute cost-based exponent: $\alpha_m \leftarrow \frac{\log_{10}(C_m)}{2}$
    \For{$i = 1$ to $N_m$}
        \State $\mu_m^{(i)}, \sigma_m^{2(i)} \leftarrow \mathcal{GP}_m(\mathbf{x}_i^{(m)})$ \Comment{LF posterior at LF input}
        \State $\mathbf{z}_i^{(m)} \leftarrow [\mathbf{x}_i^{(m)},\, \mu_m^{(i)},\, \sigma_m^{2(i)}]$ \Comment{Augmented query}
        \State $\mu_{\delta_m}^{(i)}, \sigma_{\delta_m}^{2(i)} \leftarrow \mathcal{GP}_{\delta_m}(\mathbf{z}_i^{(m)})$ \Comment{Discrepancy prediction}
        \State $\mu_M^{(i)}, \sigma_M^{2(i)} \leftarrow \mathcal{GP}_M(\mathbf{x}_i^{(m)})$ \Comment{HF prediction}
        \For{$j = 1$ to $N_M$}
            \State $d_{ij} \leftarrow \|\tilde{\mathbf{x}}_i^{(m)} - \tilde{\mathbf{x}}_j^{(M)}\|_2$
        \EndFor
        \State $r_i \leftarrow g(\min_j d_{ij})$ \Comment{Trust region radius, e.g., $g(d) = \sqrt{d}$}
        \State $\mathcal{N}_i \leftarrow \{j : d_{ij} \leq r_i\}$ \Comment{Local neighbours}
        \For{$j \in \mathcal{N}_i$}
            \State $w_j \leftarrow h(d_{ij}; \alpha_m)$ \Comment{Distance weight, e.g., $h(d; \alpha) = 1 - d^{\alpha}$}
        \EndFor
        \State $W_m^{(i)} \leftarrow 1 - \frac{1}{|\mathcal{N}_i|} \sum_{j \in \mathcal{N}_i} w_j$
        \State $\tilde{y}^{(m)}_i \leftarrow W^{(i)}_m \cdot \left( y^{(m)}_i + \mu_{\delta_m}^{(i)} \right) + (1 - W^{(i)}_m) \cdot \mu_M^{(i)}$ \Comment{Eq.~\eqref{eq:augmented_mean}}
        \State $\tilde{\sigma}^2_i \leftarrow \left(W^{(i)}_m\right)^2 \left(\hat{\sigma}^2_m + \sigma^{2(i)}_{\delta_m}\right) + \left(1 - W^{(i)}_m\right)^2 \sigma^{2(i)}_M$ \Comment{Eq.~\eqref{eq:augmented_var}}
    \EndFor
\EndFor
\State \textbf{Stage 3: Train fusion GP with fixed observation variances}
\State $\mathcal{D}_{\mathrm{aug}} \leftarrow \{(\mathbf{x}_j^{(M)}, y_j^{(M)}, \hat{\sigma}_M^2)\}_{j=1}^{N_M} \cup \bigcup_{m=1}^{M-1} \{(\mathbf{x}_i^{(m)}, \tilde{y}_i, \tilde{\sigma}_i^2)\}_{i=1}^{N_m}$
\State Apply Kumaraswamy warp $\tilde{\mathbf{x}} = \Phi(\mathbf{x}; \boldsymbol{a}, \boldsymbol{b})$ to inputs on $[0,1]^D$ \Comment{Stage 3 input transform}
\State Train $\mathcal{GP}$ on $\{(\tilde{\mathbf{x}}, y, \sigma^2)\} \in \mathcal{D}_{\mathrm{aug}}$ with fixed per-point noise variances; $(\boldsymbol{a}, \boldsymbol{b})$ learned jointly with kernel hyperparameters
\State \textbf{return} $\mathcal{M} \leftarrow \mathcal{GP}$
\end{algorithmic}
\end{algorithm*}

\subsection{Stage 1: Independent Baseline Modelling}

The first stage establishes independent Gaussian process models for each fidelity level to capture their intrinsic characteristics and estimate observation noise levels.

\paragraph{Input normalisation.}
All input coordinates are normalised to the unit hypercube $[0,1]^D$ using the domain bounds. For each dimension $d$ and input $\mathbf{x}$:
\begin{equation}
    \tilde{x}_d = \frac{x_d - x_d^{\min}}{x_d^{\max} - x_d^{\min}},
\end{equation}
where $x_d^{\min}$ and $x_d^{\max}$ are the lower and upper bounds of the $d$-th input dimension. This normalisation ensures that distances computed in Stage~2 are comparable across dimensions and that the trust region radius scales appropriately.

\paragraph{GP training.}
For each fidelity level $m \in \{1, \ldots, M\}$, we fit an independent Gaussian process with a squared exponential (RBF) kernel with automatic relevance determination (ARD):
\begin{equation}
    k_m(\mathbf{x}, \mathbf{x}') = \sigma_{f,m}^2 \exp\left( -\frac{1}{2} \sum_{d=1}^{D} \frac{(x_d - x'_d)^2}{\ell_{m,d}^2} \right),
\end{equation}
where $\sigma_{f,m}^2$ is the output variance and $\boldsymbol{\ell}_m = (\ell_{m,1}, \ldots, \ell_{m,D})$ are the per-dimension lengthscales. The hyperparameters $\{\boldsymbol{\ell}_m, \sigma_{f,m}^2, \sigma_m^2\}$ are estimated via Type-II maximum likelihood estimation (MLE), maximising the log marginal likelihood:
\begin{equation}
    \log p(\mathbf{y}_m | \mathbf{X}_m, \boldsymbol{\theta}_m) = -\frac{1}{2}\mathbf{y}_m^\top \mathbf{K}_m^{-1} \mathbf{y}_m - \frac{1}{2}\log|\mathbf{K}_m| - \frac{N_m}{2}\log(2\pi),
\end{equation}
where $\mathbf{K}_m = \mathbf{K}_{mm} + \sigma_m^2 \mathbf{I}$ is the regularised covariance matrix. The learned noise variance $\hat{\sigma}_m^2$ serves as an estimate of the intrinsic measurement uncertainty at fidelity level $m$.

\begin{figure}[h]
    \centering
    \includegraphics[width=0.75\linewidth]{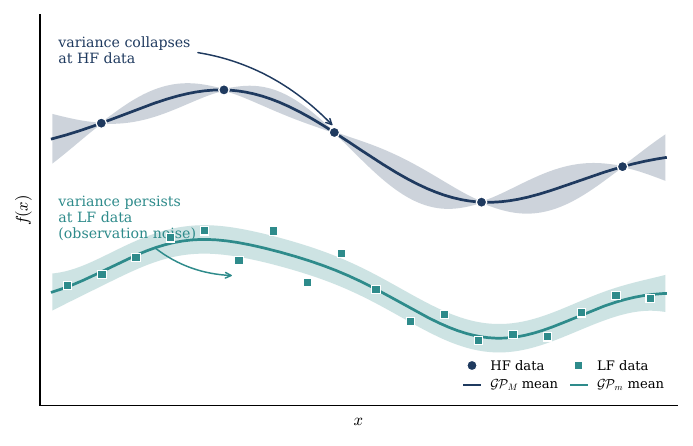}
    \caption{Stage~1 schematic. An independent Gaussian process is fitted at each fidelity. The predictive band at a training input equals the posterior latent variance plus the observation-noise variance, so the band pinches to near zero at every HF input (navy, low noise) but retains visible width at every LF input (teal, higher noise). The posterior reflects the noise on each observation rather than poor fit. The LF posterior is shifted down for visual separation only.}
    \label{fig:mast_stage1}
\end{figure}

\subsection{Stage 2: Discrepancy Modelling and Spatially-Weighted Augmentation}

The second stage transforms each lower-fidelity dataset into an augmented dataset by (i) learning the discrepancy between fidelities, (ii) computing spatially-adaptive weights, and (iii) blending corrected low-fidelity values with high-fidelity predictions.

\subsubsection{Discrepancy Modelling}

For each fidelity level $m < M$, we model the discrepancy $\delta_m(\mathbf{x}) = f_M(\mathbf{x}) - f_m(\mathbf{x})$ by fitting a GP to residuals computed at the high-fidelity observation locations.

\paragraph{Residual computation.}
Using the trained fidelity-$m$ GP, we predict at each high-fidelity location and compute residuals:
\begin{equation}
    \delta_j = y_j^{(M)} - \mu_m(\mathbf{x}_j^{(M)}), \quad j = 1, \ldots, N_M,
\end{equation}
where $\mu_m(\cdot)$ is the posterior mean of $\mathcal{GP}_m$. These residuals capture the systematic bias between fidelity levels at the high-fidelity observation locations.

\paragraph{Discrepancy GP.}
We fit a Gaussian process $\mathcal{GP}_{\delta_m}$ to the residual dataset $\{(\mathbf{x}_j^{(M)}, \delta_j)\}_{j=1}^{N_M}$, again using an RBF kernel with ARD lengthscales. This yields posterior predictions $\mu_{\delta_m}(\mathbf{x})$ and $\sigma_{\delta_m}^2(\mathbf{x})$ at any location $\mathbf{x}$, representing the expected discrepancy and its uncertainty.

\paragraph{LF-conditioned discrepancy modelling.}
In practice the discrepancy $\delta_m(\mathbf{x})$ often depends on how well the low-fidelity model is calibrated at $\mathbf{x}$, not only on location. We make this dependence explicit by conditioning $\mathcal{GP}_{\delta_m}$ on the LF posterior. For each training residual at an HF location $\mathbf{x}_j^{(M)}$, we form the augmented feature vector $\mathbf{z}_j^{(m)} = [\mathbf{x}_j^{(M)},\, \mu_m(\mathbf{x}_j^{(M)}),\, \sigma_m^2(\mathbf{x}_j^{(M)})] \in \mathbb{R}^{D+2}$. We then fit the discrepancy GP on $\{(\mathbf{z}_j^{(m)}, \delta_j)\}_{j=1}^{N_M}$ with an ARD-RBF kernel over the full $(D+2)$-dimensional feature space. Prediction at a query point $\mathbf{x}_*$ reuses the LF posterior: we evaluate $\mathcal{GP}_m$ at $\mathbf{x}_*$, construct $\mathbf{z}_*^{(m)}$, then query the discrepancy GP at $\mathbf{z}_*^{(m)}$. Because the kernel uses automatic relevance determination, the two distributional features compete with the spatial ones during hyperparameter optimisation and are down-weighted when not informative. Appendix~\ref{app:lf_conditioning_justification} discusses this construction.

\subsubsection{Spatially-Adaptive Weighting}

The weighting mechanism determines how much to trust the corrected low-fidelity value versus the high-fidelity GP prediction at each low-fidelity location. The key insight is that this trust should depend on spatial proximity to high-fidelity calibration data.

\paragraph{Distance computation.}
For each low-fidelity point $\mathbf{x}_i^{(m)}$, we compute the Euclidean distance to all high-fidelity points in the normalised input space:
\begin{equation}
    d_{ij} = \|\tilde{\mathbf{x}}_i^{(m)} - \tilde{\mathbf{x}}_j^{(M)}\|_2, \quad j = 1, \ldots, N_M.
\end{equation}

\paragraph{Adaptive trust region.}
A naive approach would aggregate contributions from all high-fidelity points, but this leads to problematic behaviour: a low-fidelity point with a nearby high-fidelity neighbour would still be influenced by distant high-fidelity observations elsewhere in the domain, artificially inflating the high-fidelity weight. To ensure that only \emph{locally relevant} high-fidelity information contributes, we introduce an adaptive trust region based on an $N$-sphere centred at $\mathbf{x}_i^{(m)}$.

The trust region radius is defined as a function of the minimum distance to any high-fidelity point:
\begin{equation}
    r_i = g(d_{\min}), \quad \text{where} \quad d_{\min} = \min_j d_{ij}.
\end{equation}
In principle, any monotonically increasing function $g: \mathbb{R}^+ \to \mathbb{R}^+$ can be used. This flexibility allows practitioners to adapt the trust region behaviour to domain-specific requirements. In this work, we adopt:
\begin{equation}
    g(d) = \sqrt{d},
\end{equation}
which we found empirically to provide robust performance across our benchmark suite. The square-root scaling ensures that the trust region expands sub-linearly with distance, providing a balance between local sensitivity and spatial coverage. Only high-fidelity points within this $N$-sphere contribute to the weight computation:
\begin{equation}
    \mathcal{N}_i = \{j : d_{ij} \leq r_i\}.
\end{equation}
When $d_{\min} \leq 1$ this guarantees the nearest high-fidelity point is included in $\mathcal{N}_i$. In dimensions $D \geq 2$, normalised Euclidean distances can exceed 1 (the unit hypercube has diameter $\sqrt{D}$), in which case $r_i = \sqrt{d_{\min}}$ may be smaller than $d_{\min}$ and the trust region would be empty. The implementation handles this by forcing the nearest high-fidelity point into $\mathcal{N}_i$ whenever the radius condition excludes it; Appendix~\ref{app:ablation_distance} reports a Normalised-Euclidean variant (D5, dividing distances by $\sqrt{D}$) and a Mahalanobis-ARD variant (D4) which both bypass the dimensional sensitivity by construction.

\paragraph{Distance-based weights.}
For each high-fidelity point $j \in \mathcal{N}_i$, we compute a distance-based weight. In general, this weight should be a decreasing function of distance, modulated by the fidelity cost ratio:
\begin{equation}
    w_j = h(d_{ij}; \alpha_m),
\end{equation}
where $h: \mathbb{R}^+ \times \mathbb{R}^+ \to [0, 1]$ is any function that is monotonically decreasing in its first argument (distance) and $\alpha_m$ is a fidelity-dependent exponent derived from the cost ratio. Alternative formulations are possible, including exponential decay $h(d; \alpha) = \exp(-\alpha d)$ or inverse-distance weighting $h(d; \alpha) = 1/(1 + d^\alpha)$. In this work, we adopt:
\begin{equation}
    h(d; \alpha) = 1 - d^{\alpha}, \quad \text{where} \quad \alpha_m = \frac{\log_{10}(C_m)}{2},
\end{equation}
which we found to work well empirically. The exponent $\alpha_m$ is determined by the cost ratio $C_m = \mathcal{C}_M / \mathcal{C}_m$, encoding the economic hierarchy of fidelity levels. When high-fidelity evaluations are substantially more expensive (large $C_m$), $\alpha_m$ increases and the weight decay becomes more gradual, extending the spatial influence of high-fidelity observations commensurate with their acquisition cost.

The individual weights are aggregated into a single low-fidelity weight:
\begin{equation}
    W_m^{(i)} = 1 - \frac{1}{|\mathcal{N}_i|} \sum_{j \in \mathcal{N}_i} w_j.
\end{equation}
By construction, $W_m^{(i)} \to 0$ when the low-fidelity point is very close to high-fidelity observations (favouring the high-fidelity prediction), and $W_m^{(i)} \to 1$ when distant (favouring the corrected low-fidelity value).

\begin{figure}[h]
    \centering
    \includegraphics[width=0.55\linewidth]{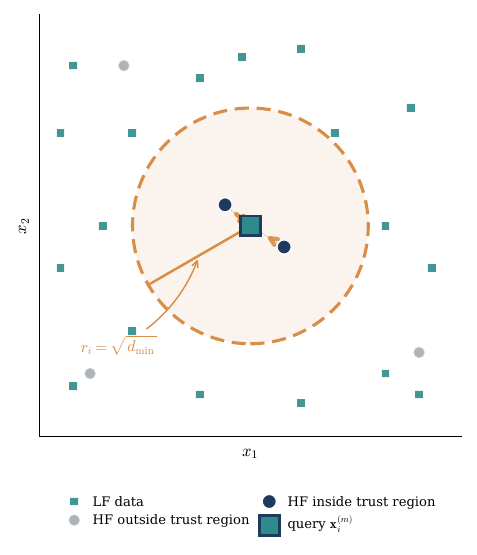}
    \caption{Stage~2 schematic. For a query low-fidelity point $\mathbf{x}_i^{(m)}$ (filled teal), the trust region is an $N$-sphere of radius $r_i = \sqrt{d_{\min}}$ centred on the query. High-fidelity samples inside the sphere contribute to the weight $W_m^{(i)}$ with magnitude decaying in distance; high-fidelity samples outside are excluded. Dashed orange: trust-region boundary; orange arrows: distance-weighted contributions.}
    \label{fig:mast_stage2}
\end{figure}

\subsubsection{Observations and Variance Propagation}

\paragraph{Blended mean.}
Each low-fidelity observation is first corrected by adding the predicted discrepancy, then blended with the high-fidelity GP prediction:
\begin{equation}
    \tilde{y}_i^{(m)} = W_m^{(i)} \cdot \underbrace{\left( y_i^{(m)} + \mu_{\delta_m}(\mathbf{z}_i^{(m)}) \right)}_{\text{corrected LF value}} + (1 - W_m^{(i)}) \cdot \underbrace{\mu_M(\mathbf{x}_i^{(m)})}_{\text{HF prediction}}.
\end{equation}

\paragraph{Variance propagation.}
Distributing the weights across the bracketed sum in Eq.~\eqref{eq:augmented_mean} writes $\tilde{y}_i^{(m)}$ as a sum of three elementary contributions: $T_1 = W_m^{(i)} y_i^{(m)}$ carrying the raw LF observation noise, $T_2 = W_m^{(i)} \mu_{\delta_m}(\mathbf{x}_i^{(m)})$ carrying the discrepancy GP's posterior uncertainty, and $T_3 = (1 - W_m^{(i)}) \mu_M(\mathbf{x}_i^{(m)})$ carrying the HF GP's posterior uncertainty. Their marginal variances, $(W_m^{(i)})^2 \hat\sigma_m^2$, $(W_m^{(i)})^2 \sigma_{\delta_m}^2$ and $(1 - W_m^{(i)})^2 \sigma_M^2$, are summed to give the total variance, the cross-covariances between $(T_1, T_2)$ and $(T_1, T_3)$ vanishing exactly and that between $(T_2, T_3)$ being neglected under the standard multi-fidelity simplification of \citet{Kennedy2000PredictingAvailable,LeGratiet2014RecursiveFidelity}. A step-by-step derivation is given in Appendix~\ref{app:variance_derivation}. The resulting propagated variance is
\begin{equation}
    \tilde{\sigma}_i^2 = \left(W_m^{(i)}\right)^2 \left(\hat{\sigma}_m^2 + \sigma_{\delta_m}^2(\mathbf{x}_i^{(m)})\right) + \left(1 - W_m^{(i)}\right)^2 \sigma_M^2(\mathbf{x}_i^{(m)}).
\end{equation}
This formulation ensures appropriate uncertainty quantification: augmented points near high-fidelity data inherit lower uncertainty from the high-fidelity predictions, while points far from high-fidelity observations retain higher uncertainty reflecting the extrapolated nature of the discrepancy correction.

\subsection{Stage 3: Heteroscedastic GP Fusion}

The final stage trains a single Gaussian process on the combined dataset consisting of original high-fidelity observations and augmented lower-fidelity observations from all levels:
\begin{equation}
    \mathcal{D}_{\mathrm{aug}} = \bigl\{(\mathbf{x}_j^{(M)}, y_j^{(M)}, \hat{\sigma}_M^2)\bigr\}_{j=1}^{N_M} \bigcup_{m=1}^{M-1} \bigl\{(\mathbf{x}_i^{(m)}, \tilde{y}_i^{(m)}, \tilde{\sigma}_i^2)\bigr\}_{i=1}^{N_m}.
\end{equation}

\paragraph{Fixed heteroscedastic noise.}
Crucially, the per-point noise variances $\{\hat{\sigma}_M^2, \tilde{\sigma}_i^2\}$ are \emph{fixed} during GP training---they are not treated as learnable hyperparameters. This design choice has several advantages:
\begin{enumerate}
    \item It avoids the numerical instabilities associated with jointly optimising heteroscedastic noise models.
    \item It allows the GP to weight observations according to their pre-determined reliability, naturally down-weighting uncertain augmented points.
    \item It preserves the uncertainty information propagated from Stage~2.
\end{enumerate}
Only the kernel hyperparameters (lengthscales and output variance) are optimised via maximum likelihood estimation.

\paragraph{Input warping.}
To handle non-stationarity introduced by blending high-fidelity and augmented low-fidelity data, we apply a per-dimension Kumaraswamy CDF warping to each input before evaluating the kernel. For each dimension $d \in \{1, \ldots, D\}$ the warp is
\begin{equation}
    \Phi_d(x_d; a_d, b_d) = 1 - (1 - x_d^{a_d})^{b_d}, \quad x_d \in [0, 1],
    \label{eq:kumaraswamy_cdf}
\end{equation}
with shape parameters $(a_d, b_d) > 0$ learned jointly with the kernel hyperparameters via maximum likelihood \citep{Snoek2014InputWarping}. The Stage~3 kernel then operates on $\tilde{\mathbf{x}} = (\Phi_1(x_1), \ldots, \Phi_D(x_D))$, so a stationary RBF kernel in warped space becomes a non-stationary kernel on the original inputs. The warp is applied only to the $D$ input dimensions; the fixed per-point noise variances are unaffected.

\begin{figure}[h]
    \centering
    \includegraphics[width=0.95\linewidth]{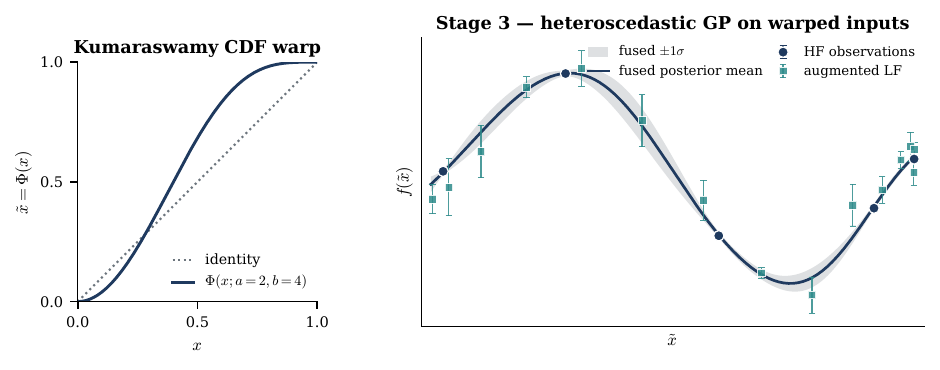}
    \caption{Stage~3 schematic. Left: per-dimension Kumaraswamy CDF $\Phi(x;\boldsymbol{a},\boldsymbol{b})$ warps $[0,1]$ onto itself (dotted: identity; navy: learned warp). Right: the Stage~3 GP is fitted on warped inputs with fixed per-point noise variances. High-fidelity observations (navy) have short noise bars; augmented low-fidelity points (teal) have longer bars reflecting the propagated Stage~2 variance. The fused posterior (navy line, shaded band) interpolates both sources.}
    \label{fig:mast_stage3}
\end{figure}

\paragraph{Prediction.}
At test time, the trained heteroscedastic GP provides posterior predictions $\mu(\mathbf{x}_*)$ and $\sigma^2(\mathbf{x}_*)$ that integrate information from all fidelity levels, with contributions weighted by both spatial proximity (via the kernel) and observation reliability (via the fixed noise variances).

\subsection{Extension to Multiple Fidelity Levels}

For $M > 2$ fidelity levels, MAST processes each lower-fidelity level independently toward the highest fidelity. This \emph{parallel} augmentation strategy differs from the \emph{sequential} approach used by recursive methods (e.g., recursive co-kriging), where fidelity $m$ is modelled conditional on fidelity $m-1$. The parallel approach has two advantages: (i) it avoids error accumulation across the fidelity hierarchy, and (ii) it allows each fidelity level to be calibrated directly against the target high-fidelity function.

\subsection{Computational Complexity}

The computational cost of MAST is dominated by GP training operations:
\begin{itemize}
    \item \textbf{Stage 1:} Training $M$ independent GPs with complexity $\mathcal{O}(N_m^3)$ each.
    \item \textbf{Stage 2:} For each fidelity $m < M$: training one discrepancy GP with $\mathcal{O}(N_M^3)$ complexity, plus $\mathcal{O}(N_m \cdot N_M)$ distance computations.
    \item \textbf{Stage 3:} Training one heteroscedastic GP with $\mathcal{O}(N_{\mathrm{total}}^3)$ complexity, where $N_{\mathrm{total}} = \sum_{m=1}^{M} N_m$.
\end{itemize}
The overall complexity is comparable to other GP-based multi-fidelity methods, with the main overhead being the discrepancy GP training and distance computations in Stage~2, which scale linearly with the number of fidelity levels.

\section{Theoretical Justifications}
\label{app:theory}

\subsection{Variance Propagation in Stage 2}
\label{app:variance_derivation}

This subsection derives the Stage~2 variance formula, Eq.~\eqref{eq:augmented_var}, from first principles by decomposing the predictive error on $\tilde{y}_i^{(m)}$ into independent uncertainty sources, stating each source's distribution, and propagating uncertainty through the deterministic combination that defines $\tilde{y}_i^{(m)}$. Let $\mathcal{D} = \bigcup_{m=1}^{M} \mathcal{D}_m$ denote the full training data. All variances and expectations are conditional on $\mathcal{D}$ and on the fitted GP hyperparameters; the notation ``$\mid \mathcal{D}$'' is suppressed where unambiguous. The variance computed below is the predictive-error variance evaluated at training inputs, since this is the quantity Stage~3 consumes as fixed observation noise; treating $y_i^{(m)}$ as a hypothetical fresh draw at $\mathbf{x}_i^{(m)}$ is the standard tractability device for arriving at this quantity.

\paragraph{Decompose the augmented observation.}
Starting from Eq.~\eqref{eq:augmented_mean}, distribute the weights across the bracketed sum:
\begin{align}
    \tilde{y}_i^{(m)}
    &= W_m^{(i)} \bigl( y_i^{(m)} + \mu_{\delta_m}(\mathbf{z}_i^{(m)}) \bigr) + \bigl( 1 - W_m^{(i)} \bigr) \mu_M(\mathbf{x}_i^{(m)}) \nonumber \\
    &= \underbrace{W_m^{(i)} \cdot y_i^{(m)}}_{T_1} + \underbrace{W_m^{(i)} \cdot \mu_{\delta_m}(\mathbf{z}_i^{(m)})}_{T_2} + \underbrace{\bigl( 1 - W_m^{(i)} \bigr) \cdot \mu_M(\mathbf{x}_i^{(m)})}_{T_3}.
    \label{eq:augmented_decomposed}
\end{align}
The predictive error on $\tilde y_i^{(m)}$ therefore decomposes as a weighted sum of three independent uncertainty sources.

\paragraph{Fix the weight.}
Conditional on $\mathcal{D}$, the weight $W_m^{(i)}$ is a deterministic function of pairwise distances in the normalised design space (Eqs.~\eqref{eq:nsphere_radius}, \eqref{eq:individual_weight}, \eqref{eq:aggregate_weight}); it does not fluctuate with the three random quantities appearing in Eq.~\eqref{eq:augmented_decomposed}. Treating $W_m^{(i)}$ as a constant scalar is thus exact under conditioning on $\mathcal{D}$.

\paragraph{Distribution of each elementary contribution.}
Given $\mathcal{D}$, each of $T_1$, $T_2$, $T_3$ has a simple Gaussian distribution.

\emph{$T_1$ (weighted raw LF observation).} The noise model of Eq.~\eqref{eq:disc} (and §\ref{sec:problem_setting}) sets $y_i^{(m)} = f_m(\mathbf{x}_i^{(m)}) + \varepsilon_i^{(m)}$ with $\varepsilon_i^{(m)} \sim \mathcal{N}(0, \hat\sigma_m^2)$. Treated as a predictive-error contribution, the residual at $\mathbf{x}_i^{(m)}$ inherits the noise variance, so
\begin{equation}
    \operatorname{Var}\bigl(T_1 \mid \mathcal{D}\bigr) = \bigl( W_m^{(i)} \bigr)^2 \cdot \operatorname{Var}\bigl(\varepsilon_i^{(m)}\bigr) = \bigl( W_m^{(i)} \bigr)^2 \, \hat\sigma_m^2.
    \label{eq:var_T1}
\end{equation}

\emph{$T_2$ (weighted discrepancy correction).} The posterior mean $\mu_{\delta_m}(\mathbf{x}_i^{(m)})$ is a Gaussian random variable with posterior variance $\sigma_{\delta_m}^2(\mathbf{x}_i^{(m)})$ (Stage~2 discrepancy GP). Therefore
\begin{equation}
    \operatorname{Var}\bigl(T_2 \mid \mathcal{D}\bigr) = \bigl( W_m^{(i)} \bigr)^2 \, \sigma_{\delta_m}^2(\mathbf{x}_i^{(m)}).
    \label{eq:var_T2}
\end{equation}

\emph{$T_3$ (weighted HF posterior prediction).} The HF GP posterior mean $\mu_M(\mathbf{x}_i^{(m)})$ is Gaussian with posterior variance $\sigma_M^2(\mathbf{x}_i^{(m)})$ (Stage~1 HF GP). Therefore
\begin{equation}
    \operatorname{Var}\bigl(T_3 \mid \mathcal{D}\bigr) = \bigl( 1 - W_m^{(i)} \bigr)^2 \, \sigma_M^2(\mathbf{x}_i^{(m)}).
    \label{eq:var_T3}
\end{equation}

\paragraph{Variance of the sum.}
For any three random quantities, the variance identity is
\begin{equation}
    \operatorname{Var}(T_1 + T_2 + T_3) = \sum_{k=1}^{3} \operatorname{Var}(T_k) + 2\!\!\sum_{1 \leq j < k \leq 3}\!\! \operatorname{Cov}(T_j, T_k).
    \label{eq:variance_expansion}
\end{equation}
The three marginal variances are already in Eqs.~\eqref{eq:var_T1}--\eqref{eq:var_T3}. We now evaluate the three cross-covariances and show that each is either exactly zero or negligible.

\paragraph{Cross-covariances.}
Write $\mu_\delta^{(i)} := \mu_{\delta_m}(\mathbf{x}_i^{(m)})$ and $\mu_M^{(i)} := \mu_M(\mathbf{x}_i^{(m)})$ for brevity.

\textbf{(A1)} $\operatorname{Cov}(T_1, T_2 \mid \mathcal{D}) = (W_m^{(i)})^2 \cdot \operatorname{Cov}(\varepsilon_i^{(m)}, \mu_\delta^{(i)} \mid \mathcal{D}) = 0$. The predictive-error noise term $\varepsilon_i^{(m)}$ is independent of the training data by the noise model of Eq.~\eqref{eq:disc}, while $\mu_\delta^{(i)}$ is a deterministic function of $\mathcal{D}$. The covariance vanishes exactly.

\textbf{(A2)} $\operatorname{Cov}(T_1, T_3 \mid \mathcal{D}) = W_m^{(i)} (1 - W_m^{(i)}) \cdot \operatorname{Cov}(\varepsilon_i^{(m)}, \mu_M^{(i)} \mid \mathcal{D}) = 0$. Identical reasoning: $\varepsilon_i^{(m)}$ is independent of training data, $\mu_M^{(i)}$ is deterministic in training data.

\textbf{(A3)} $\operatorname{Cov}(T_2, T_3 \mid \mathcal{D}) \approx 0$. This is the only approximation in the derivation. The neglected quantity is the conditional cross-covariance between the latent functions $\delta_m(\mathbf{x}_i^{(m)})$ and $f_M(\mathbf{x}_i^{(m)})$ given $\mathcal{D}$, which is non-zero because the HF GP is trained directly on $\mathbf{y}^{(M)}$ and the discrepancy GP is trained on residuals $\delta_j = y_j^{(M)} - \mu_m(\mathbf{x}_j^{(M)})$ whose first term is in $\mathbf{y}^{(M)}$. A more elaborate treatment would keep track of this induced cross-covariance via the joint posterior over the two GPs. We neglect it for tractability; the same independence simplification underpins the classical autoregressive formulation of \citet{Kennedy2000PredictingAvailable} and the recursive framework of \citet{LeGratiet2014RecursiveFidelity}, where inter-fidelity residuals and base predictors are treated as independent despite sharing data.

\paragraph{Assemble the propagated variance.}
Substituting (A1), (A2), (A3) into Eq.~\eqref{eq:variance_expansion} eliminates all cross-covariance terms, leaving
\begin{align}
    \operatorname{Var}\bigl(\tilde{y}_i^{(m)} \mid \mathcal{D}\bigr)
    &= \operatorname{Var}(T_1 \mid \mathcal{D}) + \operatorname{Var}(T_2 \mid \mathcal{D}) + \operatorname{Var}(T_3 \mid \mathcal{D}) \nonumber \\
    &= \bigl(W_m^{(i)}\bigr)^2 \hat\sigma_m^2 + \bigl(W_m^{(i)}\bigr)^2 \sigma_{\delta_m}^2(\mathbf{x}_i^{(m)}) + \bigl(1 - W_m^{(i)}\bigr)^2 \sigma_M^2(\mathbf{x}_i^{(m)}).
    \label{eq:variance_sum_explicit}
\end{align}
Grouping the first two terms (both carry weight $(W_m^{(i)})^2$) yields the Stage~2 variance formula,
\begin{equation}
    \tilde\sigma_i^2 = \bigl(W_m^{(i)}\bigr)^2 \Bigl[\hat\sigma_m^2 + \sigma_{\delta_m}^2(\mathbf{x}_i^{(m)})\Bigr] + \bigl(1 - W_m^{(i)}\bigr)^2 \sigma_M^2(\mathbf{x}_i^{(m)}),
    \label{eq:variance_sum_identity}
\end{equation}
which is Eq.~\eqref{eq:augmented_var}. The first bracket collects the variance of the corrected low-fidelity value, $\hat\sigma_m^2$ from the raw observation noise on $y_i^{(m)}$ and $\sigma_{\delta_m}^2$ from the discrepancy GP's posterior uncertainty about the latent $\delta_m(\mathbf{x}_i^{(m)})$, weighted by $(W_m^{(i)})^2$ because both are scaled by $W_m^{(i)}$ in Eq.~\eqref{eq:augmented_decomposed}. The second term is the high-fidelity posterior variance scaled by $(1 - W_m^{(i)})^2$. No other terms appear.

\paragraph{Variance floor.}
In practice we clamp the propagated variance using the floor
\begin{equation}
    \tilde\sigma^2_i \leftarrow \max\bigl(\,\tilde\sigma^2_i,\, \sigma^2_{\delta_m,\text{learned}},\, 10^{-6}\bigr),
    \label{eq:variance_floor}
\end{equation}
where $\sigma^2_{\delta_m,\text{learned}}$ is the noise variance learned by the discrepancy GP during hyperparameter fitting. The floor guards against numerical collapse when $W_m^{(i)}$ approaches $0$ or $1$ and a posterior variance is near zero. This is the regime in which the approximation A3 is most strained. The floor acts as a practical guardrail, not a theoretical upper bound, and has no effect where Eq.~\eqref{eq:variance_sum_identity} is accurate.

\subsection{Conditioning the Discrepancy GP on Low-fidelity Predictions}
\label{app:lf_conditioning_justification}

The standard multi-fidelity construction models the discrepancy between fidelities as a function of location alone: $\delta(\mathbf{x}) = f_M(\mathbf{x}) - f_m(\mathbf{x})$. This is a reasonable default when the low-fidelity model is a simple transformation of the high-fidelity one. It misses a useful signal: how well the low-fidelity model is calibrated at $\mathbf{x}$. Our LF-conditioning treatment makes this signal explicit.

\paragraph{Setup.}
For a fixed input $\mathbf{x}$ the low-fidelity posterior is Gaussian, $p(f_m(\mathbf{x}) \mid \mathcal{D}_m) = \mathcal{N}(\mu_m(\mathbf{x}), \sigma_m^2(\mathbf{x}))$. The pair $(\mu_m(\mathbf{x}), \sigma_m^2(\mathbf{x}))$ is the sufficient statistic of this posterior. We model the discrepancy as a function $\delta(\mathbf{x}, q(\mathbf{x}))$, where $q(\mathbf{x}) = (\mu_m(\mathbf{x}), \sigma_m^2(\mathbf{x}))$, and fit a Gaussian process on the augmented input space $\mathbb{R}^{D+2}$.

\paragraph{Why two moments are enough.}
When $f_m(\mathbf{x})$ is modelled by a GP, its posterior is Gaussian and therefore fully parameterised by its first two moments. Any statistic of the LF posterior that could inform discrepancy prediction is a function of $(\mu_m, \sigma_m^2)$. A common alternative, concatenating raw LF observations as additional features, throws away the posterior calibration learned in Stage~1 and introduces raw observation noise into the discrepancy model. Our formulation keeps the Stage~1 calibration intact.

\paragraph{Kernel interpretation.}
With an ARD-RBF kernel over the $(D+2)$-dimensional feature $(\mathbf{x}, \mu_m, \sigma_m^2)$, the lengthscales on the two distributional coordinates are learned separately from the spatial ones during maximum-likelihood fitting. A large learned lengthscale on $\sigma_m^2$ indicates that LF uncertainty carries little signal about the local discrepancy, and the augmented GP essentially reduces to an ordinary discrepancy GP over $\mathbf{x}$. A small learned lengthscale indicates the opposite: regions of large LF uncertainty predict systematically larger discrepancy magnitudes, and the augmented GP exploits that pattern. Automatic relevance determination thus turns LF conditioning into an opt-in behaviour: the data decides whether to use it.

\paragraph{Connection to distribution-conditioned transfer.}
Recent work on in-context multi-fidelity regression with tabular foundation models uses a related intuition, presenting the full low-fidelity distributional context to the learner to transfer across fidelity gaps \citep{Yu2026FIRE}.
Our construction is narrower and more classical: it is specific to Gaussian processes, where the LF posterior is Gaussian and therefore fully characterised by two numbers. We do not port the in-context formalism. We instead exploit the fact that the two-parameter Gaussian posterior is a natural and sufficient summary of the LF model at each query point.

\paragraph{Caveat.}
Augmenting the input space from $D$ to $D+2$ adds two hyperparameters and enlarges the optimisation landscape of the discrepancy GP. Because HF data is sparse, this could be a liability without ARD. Automatic relevance determination is what makes the augmentation safe under tight budgets: irrelevant features are learned to be flat and the effective number of active dimensions adapts to the training signal.


\section{Benchmarking Details}
\label{app:benchmarking}

\subsection{Evaluation Metrics}
\label{app:metrics}

All tables report three metrics, each normalised by the cost-equivalent HF-only baseline so that scores are directly comparable across benchmarks. We evaluate on $N_{\text{test}}$ held-out test inputs $\{\mathbf{x}_i\}_{i=1}^{N_{\text{test}}}$ with ground-truth values $f_M(\mathbf{x}_i)$. The surrogate's predictive mean and variance at $\mathbf{x}_i$ are $\mu(\mathbf{x}_i)$ and $\sigma^2(\mathbf{x}_i)$.

\paragraph{Root mean square error (RMSE).}
Standard pointwise accuracy metric. Lower is better; a normalised value below one beats the HF-only baseline.
\begin{equation}
    \mathrm{RMSE} = \sqrt{\frac{1}{N_{\text{test}}} \sum_{i=1}^{N_{\text{test}}} \bigl( \mu(\mathbf{x}_i) - f_M(\mathbf{x}_i) \bigr)^2}.
    \label{eq:rmse_def}
\end{equation}

\paragraph{Mean predictive density (mean PDF).}
The average Gaussian likelihood of the ground truth under the surrogate's predictive distribution. Evaluates predictive accuracy and uncertainty calibration jointly, so a surrogate that is simultaneously accurate and well-calibrated scores high. Higher is better; a normalised value above one beats the HF-only baseline.
\begin{equation}
    \text{mean PDF} = \frac{1}{N_{\text{test}}} \sum_{i=1}^{N_{\text{test}}} \mathcal{N}\!\Bigl( f_M(\mathbf{x}_i) \,\Big|\, \mu(\mathbf{x}_i),\, \sigma^2(\mathbf{x}_i) \Bigr).
    \label{eq:meanpdf_def}
\end{equation}
Mean PDF averages a Gaussian likelihood in the linear (not log) domain, where the contribution of any single overconfident prediction is finite even though the per-point density is unbounded; the arithmetic mean remains stable under the heavy-tailed residual outliers that dominate log-likelihood summaries. Reporting Mean PDF as a ratio relative to the cost-equivalent HF-only baseline further normalises the metric across benchmarks of different output scales.

\paragraph{Coefficient of determination ($R^2$).}
Secondary goodness-of-fit diagnostic. Measures the fraction of response variance explained by the surrogate. Higher is better; a normalised value above one beats the HF-only baseline, and values below zero indicate worse-than-mean-predictor fit.
\begin{equation}
    R^2 = 1 - \frac{\sum_{i=1}^{N_{\text{test}}} \bigl(\mu(\mathbf{x}_i) - f_M(\mathbf{x}_i)\bigr)^2}{\sum_{i=1}^{N_{\text{test}}} \bigl(f_M(\mathbf{x}_i) - \bar{f}_M\bigr)^2}, \qquad \bar{f}_M = \frac{1}{N_{\text{test}}} \sum_{i=1}^{N_{\text{test}}} f_M(\mathbf{x}_i).
    \label{eq:r2_def}
\end{equation}

\subsection{Baseline Implementations}
\label{app:baselines}

We benchmark MAST against seven multi-fidelity baselines spanning the taxonomy of §\ref{sec:relatedwork}, plus a single-fidelity reference. Except where noted, every GP baseline uses BoTorch's \texttt{SingleTaskGP} as the underlying class with ARD-RBF kernels and Type-II maximum-likelihood hyperparameter fitting. Inputs are normalised to $[0,1]^D$ prior to fitting. We use identical random seeds and identical shared training data across baselines for every benchmark--seed pair.

\paragraph{AR1 Co-kriging.}
The autoregressive cokriging of \citet{Kennedy2000PredictingAvailable}. A scalar correlation $\rho_s$ couples adjacent fidelities via $f^{(s)}(\mathbf{x}) = \rho_s f^{(s-1)}(\mathbf{x}) + \delta^{(s)}(\mathbf{x})$ with an independent discrepancy $\delta^{(s)} \sim \mathcal{GP}$. we implemented this via a custom made BoTorch model following \citet{Kennedy2000PredictingAvailable} implementation.

\paragraph{Recursive Co-kriging.}
The sequential nested-design variant of \citet{LeGratiet2014RecursiveFidelity}. Each fidelity is modelled as a GP whose posterior mean is passed as an extra feature to the next level, giving a closed-form recursion that decouples the MLE problems across levels. Implementation uses the same BoTorch scaffolding as AR1.

\paragraph{Reification-based Fusion.}
Model-reification fusion in the spirit of \citet{Thomison2017ASources}. Independent HF and LF GPs are fitted; each GP's predictive variance is augmented by a fidelity-specific floor ($\sigma^2_{\mathrm{LF}} \leftarrow \sigma^2_{\mathrm{LF}} + 10^{-2}$, $\sigma^2_{\mathrm{HF}} \leftarrow \sigma^2_{\mathrm{HF}} + 10^{-3}$) and the two predictions are fused by precision-weighted averaging, $\mu_{\mathrm{fused}}(\mathbf{x}) = \bigl(\mu_{\mathrm{LF}}/\sigma^2_{\mathrm{LF}} + \mu_{\mathrm{HF}}/\sigma^2_{\mathrm{HF}}\bigr) / \bigl(1/\sigma^2_{\mathrm{LF}} + 1/\sigma^2_{\mathrm{HF}}\bigr)$.

\paragraph{BoTorch SingleTask Multi-fidelity GP.}
BoTorch's \texttt{SingleTaskMultiFidelityGP} using the fidelity-augmented kernel of \citet{Poloczek2017Multi-informationOptimization, Wu2020PracticalTuning}. Fidelity is encoded as an additional input coordinate (LF $= 0.5$, HF $= 1.0$) and the kernel learns a fidelity-dependent lengthscale that modulates the spatial kernel.

\paragraph{NARGP.}
Nonlinear autoregressive GP \citep{Perdikaris2017NonlinearModelling} via the Emukit \texttt{NonLinearMultiFidelityModel}. Each fidelity-$s$ posterior is a GP over the augmented input $(\mathbf{x}, f^{(s-1)}(\mathbf{x}))$ with independent per-level RBF kernels. We use $5$ random optimisation restarts and $100$ Monte Carlo samples to propagate through the nonlinear hierarchy at prediction time.

\paragraph{MF-DGP.}
Deep multi-fidelity GP \citep{Cutajar2019DeepModeling} via a TensorFlow-Probability reimplementation. Layers are trained jointly for $5000$ iterations with the \texttt{multi\_step\_training} schedule and fixed inducing points. 

\paragraph{MISO Surrogate.}
Multi-information-source optimisation surrogate of \citet{Lam2015MultifidelitySources}. Independent HF and LF \texttt{SingleTaskGP}s are fitted; predictive variances are augmented with fixed fidelity noise floors ($10^{-2}$ for LF, $10^{-3}$ for HF) and the two posterior means are combined by precision weighting. Identical in final form to the reification baseline above but derived from a purely variance-driven multi-source argument rather than a model-reification one.

\paragraph{MFBNN.}
Three-stage multi-fidelity Bayesian neural network of \citet{Yi2024MFBML}. Stage one trains a deterministic deep neural network (DNN) on the abundant low-fidelity data. Stage two fits a linear transfer coefficient $\beta$ relating the DNN's prediction to the high-fidelity observations via L-BFGS-B optimisation of a least-squares objective. Stage three trains a Bayesian neural network on the high-fidelity residuals $y^{(M)} - \beta\,\widehat{f}_{\mathrm{LF}}(\mathbf{x})$ using preconditioned stochastic gradient Langevin dynamics for posterior sampling. Predictive distributions combine the LF DNN prediction scaled by $\beta$ with the Bayesian residual posterior. We use the authors' publicly available implementation with the default architecture (three hidden layers of 20 units, ReLU activations).

\paragraph{HF-only GP.}
Single-fidelity reference. A standard \texttt{SingleTaskGP} is fitted on a cost-normalised high-fidelity dataset of size $N_{\mathrm{HF}}^{\mathrm{eq}} = \lceil \mathcal{B} / \mathcal{C}_{\mathrm{HF}} \rceil$, matched to the total budget $\mathcal{B}$ consumed by each multi-fidelity method. Samples are fresh Latin hypercube draws so no training information is shared with the multi-fidelity baselines. All normalised-metric scores in the Results section are divided by this baseline's value; scores below $1$ (for RMSE) and above $1$ (for mean PDF) indicate improvement over the HF-only reference.

\subsection{Benchmark Test Functions and Real-World Datasets}
\label{app:test_functions}

This appendix provides detailed specifications for all benchmark functions used in the experimental evaluation. We first describe the cost-controlled multi-fidelity framework, then present each function with its mathematical formulation and discrepancy structure.

\subsubsection{Cost/degradation-Controlled Multi-Fidelity Framework}

All benchmark functions follow a unified degradation-controlled formulation that enables systematic variation of inter-fidelity correlation:
\begin{equation}
    f_{\mathrm{LF}}(\mathbf{x}) = f_{\mathrm{HF}}(\mathbf{x}) + d \cdot \delta(\mathbf{x}),
    \label{eq:cost_controlled_app}
\end{equation}
where $d \geq 0$ is the degradation parameter controlling the degree of low-fidelity deviation, and $\delta(\mathbf{x})$ is a function-specific discrepancy term capturing the systematic bias between fidelity levels. This formulation has several desirable properties:
\begin{itemize}
    \item When $d = 0$, the low-fidelity exactly matches the high-fidelity: $f_{\mathrm{LF}} = f_{\mathrm{HF}}$.
    \item Increasing $d$ monotonically decreases inter-fidelity correlation.
    \item The discrepancy structure $\delta(\mathbf{x})$ can be spatially varying or constant depending on the function.
\end{itemize}

Unless otherwise noted, all two-fidelity experiments use $d = 1$. For three-fidelity experiments, the intermediate mid-fidelity level uses $d = 0.5$ and the low-fidelity uses $d = 1$, establishing a hierarchy where fidelity quality degrades with increasing $d$.

\subsubsection{Observation Noise Model}
\label{app:noise_model}

The benchmark pipeline injects Gaussian noise at each fidelity before any surrogate sees the data. The motivation is twofold. First, cheaper simulations carry larger rounding, truncation and mesh-resolution error than their expensive counterparts, so treating both fidelities as noiseless would remove a realistic failure mode for uncertainty-aware surrogates. Second, the low-fidelity noise should inherit the high-fidelity baseline: in a real measurement pipeline the two sources share electronics, calibration floors or background variability, and tightening the high-fidelity noise should automatically tighten the low-fidelity observations rather than leaving an independent LF floor fixed by construction. The cost/degradation-controlled formulation in \S\ref{sec:syn} makes both properties explicit.

\paragraph{Formula.}
Each fidelity carries a specified noise fraction $\sigma_{m,\mathrm{spec}}$. High-fidelity samples are perturbed by zero-mean Gaussian noise whose standard deviation is $\sigma_{m,\mathrm{spec}}$ times the magnitude of the fidelity mean output. The low-fidelity noise shares the HF baseline and adds a cost-scaled increment on top. Formally,
\begin{align}
    \sigma_{\mathrm{HF}}^{\mathrm{std}} &= \sigma_{\mathrm{HF,spec}} \cdot \bigl|\mathbb{E}[f_{\mathrm{HF}}]\bigr|, \label{eq:noise_hf} \\
    \sigma_{\mathrm{LF}}^{\mathrm{std}} &= \bigl(\sigma_{\mathrm{HF,spec}} + d \cdot \sigma_{\mathrm{LF,spec}}\bigr) \cdot \bigl|\mathbb{E}[f_{\mathrm{LF}}]\bigr|, \label{eq:noise_lf}
\end{align}
where $d$ is the cost/degradation parameter introduced in \S\ref{sec:syn} and $|\mathbb{E}[f_m]|$ is the magnitude of the fidelity mean on its evaluation grid. The multiplicative coupling to the fidelity mean means a noise fraction of $0.01$ corresponds to roughly $1\%$ of the output scale at that fidelity rather than an absolute magnitude; the two fidelities therefore see noise at the same relative scale even when their raw output magnitudes differ.

\paragraph{Per-function values.}
Ackley, the function with identically-zero discrepancy ($\delta(\mathbf{x}) \equiv 0$), uses elevated noise values to stress the uncertainty calibration of every method. Every other function uses a lighter default. Table~\ref{tab:noise_specs} gives the values and the effective LF noise at the experimental $d = 1$.

\begin{table}[h]
\centering
\caption{Per-function observation noise specifications. The low-fidelity effective noise at $d = 1$ is $\sigma_{\mathrm{HF,spec}} + \sigma_{\mathrm{LF,spec}}$, applied as a fraction of $|\mathbb{E}[f_{\mathrm{LF}}]|$ per Eq.~\eqref{eq:noise_lf}.}
\label{tab:noise_specs}
\begin{tabular}{lccc}
\toprule
Function class & $\sigma_{\mathrm{HF,spec}}$ & $\sigma_{\mathrm{LF,spec}}$ & Effective $\sigma_{\mathrm{LF}}$ at $c = 1$ \\
\midrule
Ackley                               & 0.02 & 0.13 & 0.15 \\
All other functions                  & 0.01 & 0.02 & 0.03 \\
\bottomrule
\end{tabular}
\end{table}

Both values in each row are dimensionless fractions of the fidelity mean magnitude, not absolute standard deviations. Under Eqs.~\eqref{eq:noise_hf} and~\eqref{eq:noise_lf} the effective HF noise is $0.02\,|\mathbb{E}[f_{\mathrm{HF}}]|$ for Ackley and $0.01\,|\mathbb{E}[f_{\mathrm{HF}}]|$ for the rest; the effective LF noise is $0.15\,|\mathbb{E}[f_{\mathrm{LF}}]|$ and $0.03\,|\mathbb{E}[f_{\mathrm{LF}}]|$ respectively. The coupling through $\sigma_{\mathrm{HF,spec}}$ in Eq.~\eqref{eq:noise_lf} means the LF noise floor inherits any change in HF noise, so noisy-data ablations that perturb one end of the hierarchy propagate consistently to the other. The main-body summary in \S\ref{sec:syn} points back to this subsection; values used in the variance derivation of Appendix~\ref{app:variance_derivation} are consistent with this table.

\subsubsection{Discrepancy Type Classification}

The benchmark functions exhibit different discrepancy structures, which we classify into four categories:

\paragraph{Spatially-varying discrepancy.}
The discrepancy $\delta(\mathbf{x})$ varies across the input domain, meaning the low-fidelity approximation quality differs by region. This is common in engineering applications where simplified models may be accurate in some operating regimes but not others.

\paragraph{Oscillatory discrepancy.}
The discrepancy takes the form $\delta(\mathbf{x}) = A \sin(\omega_1 x_1 + \omega_2 x_2)$, introducing periodic errors that challenge methods assuming smooth inter-fidelity relationships.

\paragraph{Linear scaling discrepancy.}
The discrepancy is a linear function of the high-fidelity response: $\delta(\mathbf{x}) = (\gamma - 1) f_{\mathrm{HF}}(\mathbf{x}) + \beta$. This represents the classical autoregressive assumption where linear inter-fidelity correlation holds globally. Function: Park2.

\paragraph{Noise-only discrepancy.}
The discrepancy is identically zero: $\delta(\mathbf{x}) = 0$. Fidelity levels differ only in observation noise, testing a method's ability to appropriately weight data sources based on noise characteristics alone.

\subsubsection{Branin (2D)}
\label{app:branin}

The Branin function \citep{Forrester2008EngineeringGuide} is a classical 2D optimisation benchmark with three global minima.

\textbf{Domain:} $x_1 \in [-5, 10]$, $x_2 \in [0, 15]$

\textbf{High-fidelity:}
\begin{equation}
    f_{\mathrm{HF}}(\mathbf{x}) = a\left(x_2 - bx_1^2 + cx_1 - r\right)^2 + s(1-t)\cos(x_1) + s
\end{equation}
with parameters $a = 1$, $b = \frac{5.1}{4\pi^2}$, $c = \frac{5}{\pi}$, $r = 6$, $s = 10$, $t = \frac{1}{8\pi}$.

\textbf{Discrepancy term:} Based on perturbing the $b$ coefficient:
\begin{equation}
    \delta(\mathbf{x}) = f_{\mathrm{HF}}(\mathbf{x}; b' = b - 0.1) - f_{\mathrm{HF}}(\mathbf{x})
\end{equation}

\textbf{Global minima:} $f(\mathbf{x}^*) \approx 0.398$ at $\mathbf{x}^* \in \{(-\pi, 12.275), (\pi, 2.275), (9.425, 2.475)\}$

\subsubsection{Rosenbrock ($n$D, $n \geq 2$)}
\label{app:rosenbrock}

A variant of the classic Rosenbrock valley function \citep{Molga2005TestNeeds, Lam2015MultifidelitySources} with an oscillatory low-fidelity component.

\textbf{Domain:} $\mathbf{x} \in [-5, 10]^n$

\textbf{High-fidelity:}
\begin{equation}
    f_{\mathrm{HF}}(\mathbf{x}) = \sum_{i=1}^{n-1} \left[100(x_{i+1} - x_i^2)^2 + (1 - x_i)^2\right]
\end{equation}

\textbf{Discrepancy term:}
\begin{equation}
    \delta(\mathbf{x}) = A \sin(\omega_1 x_1 + \omega_2 x_2)
\end{equation}
with default parameters $A = 0.1$, $\omega_1 = 10$, $\omega_2 = 5$.

\textbf{Global minimum:} $f(\mathbf{1}) = 0$

\subsubsection{Rastrigin ($n$D)}
\label{app:rastrigin}

A highly multimodal function commonly used for testing global optimisation algorithms \citep{Molga2005TestNeeds}.

\textbf{Domain:} $\mathbf{x} \in [-5.12, 5.12]^n$

\textbf{High-fidelity:}
\begin{equation}
    f_{\mathrm{HF}}(\mathbf{x}) = An + \sum_{i=1}^{n} \left[x_i^2 - A\cos(2\pi x_i)\right]
\end{equation}
with $A = 10$.

\textbf{Discrepancy term:}
\begin{equation}
    \delta(\mathbf{x}) = A \sin(\omega_1 x_1 + \omega_2 x_2)
\end{equation}
with default parameters $A = 0.1$, $\omega_1 = 10$, $\omega_2 = 5$.

\textbf{Global minimum:} $f(\mathbf{0}) = 0$

\subsubsection{Ackley (\texorpdfstring{$n$}{n}D)}
\label{app:ackley}

A widely-used multimodal function with many local minima surrounding a global minimum at the origin \citep{Molga2005TestNeeds}.

\textbf{Domain:} $\mathbf{x} \in [-5, 5]^n$

\textbf{High-fidelity:}
\begin{equation}
    f_{\mathrm{HF}}(\mathbf{x}) = -a \exp\left(-b\sqrt{\frac{1}{n}\sum_{i=1}^{n} x_i^2}\right) - \exp\left(\frac{1}{n}\sum_{i=1}^{n} \cos(cx_i)\right) + a + e
\end{equation}
with parameters $a = 20$, $b = 0.2$, $c = 2\pi$, and $e = \exp(1)$.

\textbf{Discrepancy term:}
\begin{equation}
    \delta(\mathbf{x}) = 0
\end{equation}
The low-fidelity differs from high-fidelity only through observation noise levels.

\textbf{Global minimum:} $f(\mathbf{0}) = 0$

\subsubsection{Levy ($n$D)}
\label{app:levy}

A multimodal function with sinusoidal components \citep{Laguna2005ExperimentalFunctions}.

\textbf{Domain:} $\mathbf{x} \in [-10, 10]^n$

\textbf{High-fidelity:}
\begin{align}
    f_{\mathrm{HF}}(\mathbf{x}) &= \sin^2(\pi w_1) + \sum_{i=1}^{n-1} (w_i - 1)^2 \left[1 + 10\sin^2(\pi w_i + 1)\right] \nonumber \\
    &\quad + (w_n - 1)^2 \left[1 + \sin^2(2\pi w_n)\right]
\end{align}
where $w_i = 1 + \frac{x_i - 1}{4}$.

\textbf{Discrepancy term:}
\begin{equation}
    \delta(\mathbf{x}) = A \sin(\omega_1 x_1 + \omega_2 x_2)
\end{equation}
with default parameters $A = 0.1$, $\omega_1 = 10$, $\omega_2 = 5$.

\textbf{Global minimum:} $f(\mathbf{1}) = 0$

\subsubsection{Hartmann3 (3D)}
\label{app:hartmann3}

A 3D function with four local minima, commonly used for testing global optimisation \citep{Kim2020BenchmarkOptimization}.

\textbf{Domain:} $\mathbf{x} \in [0, 1]^3$

\textbf{High-fidelity:}
\begin{equation}
    f_{\mathrm{HF}}(\mathbf{x}) = \sum_{i=1}^{4} \alpha_i \exp\left(-\sum_{j=1}^{3} A_{ij}(x_j - P_{ij})^2\right)
\end{equation}
with parameters:
\begin{equation*}
    \boldsymbol{\alpha} = (1.0, 1.2, 3.0, 3.2)^\top
\end{equation*}
\begin{equation*}
    \mathbf{A} = \begin{pmatrix} 3.0 & 10.0 & 30.0 \\ 0.1 & 10.0 & 35.0 \\ 3.0 & 10.0 & 30.0 \\ 0.1 & 10.0 & 35.0 \end{pmatrix}, \quad
    \mathbf{P} = \begin{pmatrix} 0.3689 & 0.1170 & 0.2673 \\ 0.4699 & 0.4387 & 0.7470 \\ 0.1091 & 0.8732 & 0.5547 \\ 0.0381 & 0.5743 & 0.8828 \end{pmatrix}
\end{equation*}

\textbf{Discrepancy term:} Based on perturbing the $\boldsymbol{\alpha}$ vector:
\begin{equation}
    \delta(\mathbf{x}) = f_{\mathrm{HF}}(\mathbf{x}; \boldsymbol{\alpha} + \Delta\boldsymbol{\alpha}) - f_{\mathrm{HF}}(\mathbf{x})
\end{equation}
with $\Delta\boldsymbol{\alpha} = (0.01, -0.01, -0.1, 0.1)^\top$.

\subsubsection{Hartmann6 (6D)}
\label{app:hartmann6}

A 6D extension of the Hartmann function with four local minima \citep{Picheny2013AOptimization}.

\textbf{Domain:} $\mathbf{x} \in [0, 1]^6$

\textbf{High-fidelity:}
\begin{equation}
    f_{\mathrm{HF}}(\mathbf{x}) = \sum_{i=1}^{4} \alpha_i \exp\left(-\sum_{j=1}^{6} A_{ij}(x_j - P_{ij})^2\right)
\end{equation}
with parameters:
\begin{equation*}
    \boldsymbol{\alpha} = (1.0, 1.2, 3.0, 3.2)^\top
\end{equation*}
\begin{equation*}
    \mathbf{A} = \begin{pmatrix} 10.0 & 3.0 & 17.0 & 3.5 & 1.7 & 8.0 \\ 0.05 & 10.0 & 17.0 & 0.1 & 8.0 & 14.0 \\ 3.0 & 3.5 & 1.7 & 10.0 & 17.0 & 8.0 \\ 17.0 & 8.0 & 0.05 & 10.0 & 0.1 & 14.0 \end{pmatrix}
\end{equation*}
\begin{equation*}
    \mathbf{P} = 10^{-4} \times \begin{pmatrix} 1312 & 1696 & 5569 & 124 & 8283 & 5886 \\ 2329 & 4135 & 8307 & 3736 & 1004 & 9991 \\ 2348 & 1451 & 3522 & 2883 & 3047 & 6650 \\ 4047 & 8828 & 8732 & 5743 & 1091 & 381 \end{pmatrix}
\end{equation*}

\textbf{Discrepancy term:} Same perturbation as Hartmann3:
\begin{equation}
    \delta(\mathbf{x}) = f_{\mathrm{HF}}(\mathbf{x}; \boldsymbol{\alpha} + \Delta\boldsymbol{\alpha}) - f_{\mathrm{HF}}(\mathbf{x})
\end{equation}
with $\Delta\boldsymbol{\alpha} = (0.01, -0.01, -0.1, 0.1)^\top$.

\subsubsection{Park1 (4D)}
\label{app:park1}

The first Park function with a combination of rational and exponential terms \citep{Park1991TuningDesigns}.

\textbf{Domain:} $\mathbf{x} \in [0, 1]^4$

\textbf{High-fidelity:}
\begin{equation}
    f_{\mathrm{HF}}(\mathbf{x}) = \frac{x_1}{2}\left(\sqrt{1 + (x_2 + x_3^2)\frac{x_4}{x_1^2}} - 1\right) + (x_1 + 3x_4)\exp(1 + \sin(x_3))
\end{equation}

\textbf{Discrepancy term:}
\begin{equation}
    \delta(\mathbf{x}) = \frac{\sin(x_1)}{10} \cdot f_{\mathrm{HF}}(\mathbf{x}) - 2x_1 + x_2^2 + x_3^2 + 0.5
\end{equation}

\subsubsection{Park2 (4D)}
\label{app:park2}

The second Park function with exponential and trigonometric terms \citep{Park1991TuningDesigns}.

\textbf{Domain:} $\mathbf{x} \in [0, 1]^4$

\textbf{High-fidelity:}
\begin{equation}
    f_{\mathrm{HF}}(\mathbf{x}) = \frac{2}{3}\exp(x_1 + x_2) - x_4\sin(x_3) + x_3
\end{equation}

\textbf{Discrepancy term:}
\begin{equation}
    \delta(\mathbf{x}) = (\gamma - 1) f_{\mathrm{HF}}(\mathbf{x}) + \beta
\end{equation}
with default parameters $\gamma = 1.2$ and $\beta = -1.0$.

\textbf{Low-fidelity (at $d=1$):}
\begin{equation}
    f_{\mathrm{LF}}(\mathbf{x}) = \gamma \cdot f_{\mathrm{HF}}(\mathbf{x}) + \beta
\end{equation}

\subsubsection{Borehole (8D)}
\label{app:borehole}

A physics-based function modelling water flow rate through a borehole \citep{Morris1993BayesianPrediction}. This function has practical significance in groundwater modelling.

\textbf{Domain and physical parameters:}
\begin{center}
\begin{tabular}{clcc}
\toprule
Variable & Description & Lower & Upper \\
\midrule
$r_w$ & Borehole radius (m) & 0.05 & 0.15 \\
$r$ & Radius of influence (m) & 100 & 50,000 \\
$T_u$ & Upper aquifer transmissivity (m$^2$/yr) & 63,070 & 115,600 \\
$H_u$ & Upper aquifer potentiometric head (m) & 990 & 1,110 \\
$T_l$ & Lower aquifer transmissivity (m$^2$/yr) & 63.1 & 116 \\
$H_l$ & Lower aquifer potentiometric head (m) & 700 & 820 \\
$L$ & Borehole length (m) & 1,120 & 1,680 \\
$K_w$ & Borehole hydraulic conductivity (m/yr) & 9,855 & 12,045 \\
\bottomrule
\end{tabular}
\end{center}

\textbf{High-fidelity:}
\begin{equation}
    f_{\mathrm{HF}}(\mathbf{x}) = \frac{2\pi T_u (H_u - H_l)}{\ln(r/r_w)\left(1 + \dfrac{2LT_u}{\ln(r/r_w) \cdot r_w^2 K_w} + \dfrac{T_u}{T_l}\right)}
\end{equation}

\textbf{Discrepancy term:} Based on coefficient simplifications:
\begin{equation}
    \delta(\mathbf{x}) = \frac{5 T_u (H_u - H_l)}{\ln(r/r_w)\left(1.5 + \dfrac{2LT_u}{\ln(r/r_w) \cdot r_w^2 K_w} + \dfrac{T_u}{T_l}\right)} - f_{\mathrm{HF}}(\mathbf{x})
\end{equation}
The low-fidelity uses coefficient $5$ instead of $2\pi \approx 6.28$ and $1.5$ instead of $1$ in the denominator.

\subsubsection{Summary Table}
\label{app:summary_table}

Table~\ref{tab:function_summary} summarises all benchmark functions used in this study, listed by ascending dimensionality of the active configuration.

\begin{table}[ht]
\centering
\caption{Summary of benchmark test functions in the active evaluation roster.}
\label{tab:function_summary}
\begin{tabular}{llll}
\toprule
Function & Dim. & Domain & Discrepancy Type \\
\midrule
Branin           & 2   & $[-5, 10] \times [0, 15]$  & Coefficient perturbation        \\
Hartmann3        & 3   & $[0, 1]^3$                 & Parameter perturbation          \\
Ackley           & 4   & $[-5, 5]^4$                & Noise only                      \\
Park1            & 4   & $[0, 1]^4$                 & Multiplicative + polynomial     \\
Park2            & 4   & $[0, 1]^4$                 & Linear scaling + offset         \\
Hartmann6        & 6   & $[0, 1]^6$                 & Parameter perturbation          \\
Levy             & 7   & $[-10, 10]^7$              & Oscillatory                     \\
Borehole         & 8   & Physical bounds            & Coefficient simplification      \\
Rastrigin        & 15  & $[-5.12, 5.12]^{15}$       & Oscillatory                     \\
Rosenbrock       & 20  & $[-5, 10]^{20}$            & Oscillatory                     \\
\bottomrule
\end{tabular}
\end{table}

\subsubsection{Real-World Engineering Datasets}
\label{app:rw_datasets}

The three real-world engineering datasets span materials science and aerodynamics. Each combines a high-fidelity source (physical measurements or high-resolution simulations) with a cheaper low-fidelity source (analytical models or physics-inspired surrogates). The benchmarks span 3 to 23 input dimensions and a range of inter-fidelity correlation regimes, from strongly correlated analytical simplifications (Concrete, HOIP) to a moderately correlated CFD pairing (DrivAerNet).

\subsubsection{Concrete Compressive Strength (8D)}
\label{app:rw_concrete}

\paragraph{Description.}
Predictive modelling of concrete compressive strength $f_c$ (MPa) as a function of mixture composition (cement, slag, fly ash, water, superplasticizer, coarse aggregate, fine aggregate) and curing age. The high fidelity is laboratory-measured compressive strength from the UCI concrete dataset \citep{Yeh1998Concrete}. The low fidelity is a generalised Abrams' law \citep{Yu2026FIRE}, a physics-inspired multiplicative power-law fitted by least squares on log-strength:
\begin{align}
    \log f_c^{\mathrm{LF}}(\mathbf{x}) &= 2.56 - 0.815 \log(\mathrm{w/c} + \epsilon) - 0.0380 \log(\mathrm{cement} + \epsilon) \notag \\
    &\quad + 0.0161 \log(\mathrm{slag} + \epsilon) + 0.00231 \log(\mathrm{fly\,ash} + \epsilon) \notag \\
    &\quad + 0.0148 \log(\mathrm{superplasticizer} + \epsilon) + 0.292 \log(\mathrm{age} + \epsilon),
\end{align}
where $\mathrm{w/c} = \mathrm{water} / (\mathrm{cement} + \epsilon)$ and $\epsilon = 1$ is an offset handling zero components. Coarse and fine aggregate do not appear in the LF formula.

\paragraph{Setup.}
High-fidelity: 1{,}030 laboratory measurements loaded from the UCI dataset. Low-fidelity: $N_{\mathrm{LF}} = 120$ fresh Latin hypercube samples per experiment, evaluated through the Abrams' law formula above (not CSV rows). Cost ratio $\mathcal{C}_{\mathrm{HF}} : \mathcal{C}_{\mathrm{LF}} = 10 : 1$, budget $\mathcal{B} = 5D = 40$ cost-equivalent HF evaluations with a $70\%/30\%$ HF/LF split yielding $N_{\mathrm{HF}} = 28$, $N_{\mathrm{LF}} = 120$.

Concrete results are reported alongside the synthetic two-fidelity benchmarks in main-body Table~\ref{tab:results_combined}; normalised $R^2$ is in Table~\ref{tab:results_r2}.

\subsubsection{HOIP Perovskite Band Gap (3D, three-fidelity)}
\label{app:rw_hoip}

\paragraph{Description.}
Multi-fidelity prediction of band-gap energy (eV) for hybrid organic-inorganic perovskite compositions across three tiers of DFT accuracy. The underlying dataset of 1{,}346 HOIP compositions with DFT-computed band gaps is due to \citet{Kim2017HOIP}. The input is a 3D categorical specification (cation index, anion index, organic-group index) encoded ordinally to $[0, 1]$. The three-tier multi-fidelity pairing and cost hierarchy (§5.4) follow the GP+ library \citep{Yousefpour2024GPPlus}.

\paragraph{Setup.}
The HF, MF and LF pools contain 480, 179 and 240 rows respectively. The cost hierarchy is $\mathcal{C}_{\mathrm{HF}} : \mathcal{C}_{\mathrm{MF}} : \mathcal{C}_{\mathrm{LF}} = 40 : 10 : 1$, normalising $\mathcal{C}_{\mathrm{HF}} = 1$ to give $1 : 0.25 : 0.025$. At budget $\mathcal{B} = 5D = 15$ with a $50\%/30\%/20\%$ HF/MF/LF split, the training samples drawn from these pools are $N_{\mathrm{HF}} = 8$, $N_{\mathrm{MF}} = 18$, $N_{\mathrm{LF}} = 120$.

\paragraph{Distance metric.}
Because the inputs are categorical, Euclidean distance on the ordinal $[0,1]$ encoding is not meaningful. We therefore replace the Euclidean metric in Eq.~\eqref{eq:disc} -- the Stage~2 distance used by MAST to form the trust region and distance weights -- with the normalised Hamming distance $d(\mathbf{a}, \mathbf{b}) = \frac{1}{D}\sum_{d=1}^{D} \mathbf{1}\bigl[a_d \neq b_d\bigr]$ on the raw categorical indices. The GP kernels themselves continue to operate on the ordinal $[0,1]$ encoding. This swap demonstrates that MAST's spatial-weighting mechanism is agnostic to the choice of metric: only a monotonically-increasing-in-distance weight function is required (see Eq.~\eqref{eq:individual_weight} and Appendix~\ref{app:ablation_distance}), so any metric suited to the input space -- Euclidean on continuous inputs, Hamming on categorical inputs, Mahalanobis on anisotropic spaces -- can be substituted without changing any other aspect of the pipeline.

HOIP results are reported in the main-body three-fidelity table (Table~\ref{tab:three_fid_combined}), since the HF/MF/LF construction is the natural fit for this dataset; normalised $R^2$ is in Table~\ref{tab:three_fid_r2}. LF training rows are drawn without replacement from the 240-row pool, so the requested LF count is capped at 240 in budget-sweep settings where the cost-budget formula would otherwise demand more; this caps the effective $N_{\mathrm{LF}}$ at the high-budget end of Appendix~\ref{app:pa_combined_budget} (sweep points beyond $\mathcal{B} \approx 12$ share the same $N_{\mathrm{LF}} = 240$).

\subsubsection{DrivAerNet Car Aerodynamics (23D)}
\label{app:rw_drivaernet}

\paragraph{Description.}
Drag coefficient $C_d$ for automotive body designs across a 23-dimensional geometric design space (angles, lengths, curvatures, offsets, thicknesses). The high fidelity is CFD (OpenFOAM with the $k$-$\omega$ SST turbulence model, 24M cells per case) from the DrivAerNet dataset \citep{Elrefaie2024DrivAerNet}. The low fidelity is the quasi-2D analytical surrogate released alongside the FIRE benchmark suite \citep{Yu2026FIRE}.

\paragraph{Setup.}
Dataset contains 3{,}991 records. Input: 23 geometric parameters covering ramp, diffuser and trunklid angles; mirror rotation and translation; window, pillar, trunklid, bumper, door-handle, and fender parameters; and overall car length, width, roof height, and greenhouse angle. Cost ratio: $10 : 1$. HF-LF correlation $\approx 0.59$, a moderate regime between the strongly-correlated analytical benchmarks and the more weakly-correlated multi-fidelity pairings. Gaussian noise is injected during training ($\sigma_{\mathrm{HF}} = 0.01$, $\sigma_{\mathrm{LF}} = 0.02$) to mimic CFD numerical error and analytical bias. DrivAerNet results are reported alongside the synthetic benchmarks in main-body Table~\ref{tab:results_combined}; normalised $R^2$ is in Table~\ref{tab:results_r2}.


\section{Ablation Study}
\label{app:ablation}

To probe the sensitivity of MAST to its three main design choices, we run an axis-at-a-time ablation covering the trust-region radius function $g(\cdot)$, the distance-based weight function $h(\cdot)$, and the distance metric used in Stage~2. The study uses the ten-function benchmark roster, 25 independent seeds per configuration, and $\mathcal{B} = 5D$ budget with $70\%/30\%$ HF/LF cost split as the main synthetic evaluation (see §\ref{sec:syn}). The vanilla MAST configuration (square-root radius, cost-aware power-law weight, Euclidean distance) is held as the fixed baseline and each axis is perturbed independently. Normalised RMSE and normalised mean PDF, each divided by the cost-equivalent HF-only baseline, are reported below; values below 1 (for RMSE) and above 1 (for PDF) indicate improvement.

\subsection{Trust Region Radius Function}
\label{app:ablation_radius}

The trust-region radius $r_i = g(d_{\min})$ controls how far the local neighbourhood $\mathcal{N}_i$ extends around a low-fidelity query point. We sweep five candidate functions:
\begin{enumerate}
    \item $g(d_{\min}) = \sqrt{d_{\min}}$ (square root, default),
    \item $g(d_{\min}) = d_{\min}$ (linear),
    \item $g(d_{\min}) = \sqrt[3]{d_{\min}}$ (cube root),
    \item $g(d_{\min}) = 10^{-6}$ (vanishing radius; trust region collapses to the nearest HF point),
    \item $g(d_{\min}) = \infty$ (global, disables the trust region).
\end{enumerate}
The default square-root form expands the neighbourhood sub-linearly with distance, balancing local sensitivity against spatial coverage. R2 and R3 vary the scaling exponent; R4 collapses the trust region to the single nearest HF point (a degenerate local case); R5 eliminates locality entirely.

\begin{table}[h]
\centering
\caption{Trust-region radius function ablation --- normalised RMSE per benchmark, mean $\pm$ std across 25 seeds; lower is better. Best variant per function in bold.}
\label{tab:ablation_radius}
\resizebox{\textwidth}{!}{%
\begin{tabular}{l c c c c c}
\toprule
Test function & R1: $\sqrt{d_{\min}}$ (default) & R2: $d_{\min}$ & R3: $d_{\min}^{1/3}$ & R4: $10^{-6}$ (zero+) & R5: $\infty$ (global) \\
\midrule
Branin (2D)       & $1.192{\scriptstyle\pm0.397}$ & $\mathbf{1.163{\scriptstyle\pm0.393}}$ & $1.222{\scriptstyle\pm0.419}$ & $\mathbf{1.163{\scriptstyle\pm0.393}}$ & $1.334{\scriptstyle\pm0.488}$ \\
Hartmann-3 (3D)   & $0.692{\scriptstyle\pm0.188}$ & $0.799{\scriptstyle\pm0.221}$ & $0.623{\scriptstyle\pm0.168}$ & $0.799{\scriptstyle\pm0.221}$ & $\mathbf{0.464{\scriptstyle\pm0.125}}$ \\
Ackley (4D)       & $0.945{\scriptstyle\pm0.256}$ & $0.954{\scriptstyle\pm0.276}$ & $\mathbf{0.941{\scriptstyle\pm0.247}}$ & $0.954{\scriptstyle\pm0.276}$ & $0.945{\scriptstyle\pm0.233}$ \\
Park-1 (4D)       & $0.970{\scriptstyle\pm0.475}$ & $0.978{\scriptstyle\pm0.474}$ & $\mathbf{0.963{\scriptstyle\pm0.486}}$ & $0.978{\scriptstyle\pm0.474}$ & $1.108{\scriptstyle\pm0.498}$ \\
Park-2 (4D)       & $0.644{\scriptstyle\pm0.177}$ & $0.704{\scriptstyle\pm0.198}$ & $0.605{\scriptstyle\pm0.161}$ & $0.704{\scriptstyle\pm0.198}$ & $\mathbf{0.560{\scriptstyle\pm0.133}}$ \\
Hartmann-6 (6D)   & $0.882{\scriptstyle\pm0.193}$ & $0.881{\scriptstyle\pm0.188}$ & $\mathbf{0.873{\scriptstyle\pm0.180}}$ & $0.881{\scriptstyle\pm0.188}$ & $0.874{\scriptstyle\pm0.176}$ \\
Levy (7D)         & $0.783{\scriptstyle\pm0.080}$ & $0.789{\scriptstyle\pm0.082}$ & $\mathbf{0.779{\scriptstyle\pm0.080}}$ & $0.789{\scriptstyle\pm0.082}$ & $0.793{\scriptstyle\pm0.097}$ \\
Borehole (8D)     & $0.888{\scriptstyle\pm0.240}$ & $\mathbf{0.869{\scriptstyle\pm0.256}}$ & $0.889{\scriptstyle\pm0.234}$ & $\mathbf{0.869{\scriptstyle\pm0.256}}$ & $1.001{\scriptstyle\pm0.244}$ \\
Rastrigin (15D)   & $0.947{\scriptstyle\pm0.061}$ & $\mathbf{0.945{\scriptstyle\pm0.058}}$ & $0.945{\scriptstyle\pm0.062}$ & $\mathbf{0.945{\scriptstyle\pm0.058}}$ & $0.947{\scriptstyle\pm0.059}$ \\
Rosenbrock (20D)  & $0.356{\scriptstyle\pm0.025}$ & $0.356{\scriptstyle\pm0.025}$ & $\mathbf{0.356{\scriptstyle\pm0.025}}$ & $0.356{\scriptstyle\pm0.025}$ & $0.356{\scriptstyle\pm0.025}$ \\
\midrule
\textbf{\# Wins}  & 0              & 3              & 5              & 3              & 2 \\
\bottomrule
\end{tabular}%
}
\end{table}

\begin{table}[h]
\centering
\caption{Trust-region radius function ablation --- normalised mean PDF per benchmark, mean $\pm$ std across 25 seeds; higher is better. Best variant per function in bold.}
\label{tab:ablation_radius_pdf}
\resizebox{\textwidth}{!}{%
\begin{tabular}{l c c c c c}
\toprule
Test function & R1: $\sqrt{d_{\min}}$ (default) & R2: $d_{\min}$ & R3: $d_{\min}^{1/3}$ & R4: $10^{-6}$ (zero+) & R5: $\infty$ (global) \\
\midrule
Branin (2D)       & $1.279{\scriptstyle\pm0.427}$ & $\mathbf{1.325{\scriptstyle\pm0.480}}$ & $1.284{\scriptstyle\pm0.423}$ & $\mathbf{1.325{\scriptstyle\pm0.480}}$ & $1.228{\scriptstyle\pm0.438}$ \\
Hartmann-3 (3D)   & $1.826{\scriptstyle\pm0.500}$ & $1.663{\scriptstyle\pm0.475}$ & $1.996{\scriptstyle\pm0.536}$ & $1.663{\scriptstyle\pm0.475}$ & $\mathbf{2.496{\scriptstyle\pm0.637}}$ \\
Ackley (4D)       & $1.229{\scriptstyle\pm0.321}$ & $\mathbf{1.241{\scriptstyle\pm0.326}}$ & $1.218{\scriptstyle\pm0.314}$ & $\mathbf{1.241{\scriptstyle\pm0.326}}$ & $1.166{\scriptstyle\pm0.289}$ \\
Park-1 (4D)       & $1.460{\scriptstyle\pm0.494}$ & $1.445{\scriptstyle\pm0.487}$ & $\mathbf{1.464{\scriptstyle\pm0.484}}$ & $1.445{\scriptstyle\pm0.487}$ & $1.285{\scriptstyle\pm0.387}$ \\
Park-2 (4D)       & $1.590{\scriptstyle\pm0.284}$ & $1.488{\scriptstyle\pm0.264}$ & $1.663{\scriptstyle\pm0.294}$ & $1.488{\scriptstyle\pm0.264}$ & $\mathbf{1.795{\scriptstyle\pm0.338}}$ \\
Hartmann-6 (6D)   & $1.792{\scriptstyle\pm0.523}$ & $\mathbf{1.806{\scriptstyle\pm0.523}}$ & $1.791{\scriptstyle\pm0.531}$ & $\mathbf{1.806{\scriptstyle\pm0.523}}$ & $1.749{\scriptstyle\pm0.534}$ \\
Levy (7D)         & $1.435{\scriptstyle\pm0.162}$ & $1.429{\scriptstyle\pm0.164}$ & $\mathbf{1.440{\scriptstyle\pm0.162}}$ & $1.429{\scriptstyle\pm0.164}$ & $1.399{\scriptstyle\pm0.172}$ \\
Borehole (8D)     & $1.585{\scriptstyle\pm0.239}$ & $\mathbf{1.642{\scriptstyle\pm0.258}}$ & $1.554{\scriptstyle\pm0.241}$ & $\mathbf{1.642{\scriptstyle\pm0.258}}$ & $1.288{\scriptstyle\pm0.209}$ \\
Rastrigin (15D)   & $1.186{\scriptstyle\pm0.072}$ & $\mathbf{1.190{\scriptstyle\pm0.069}}$ & $1.188{\scriptstyle\pm0.070}$ & $\mathbf{1.190{\scriptstyle\pm0.069}}$ & $1.183{\scriptstyle\pm0.071}$ \\
Rosenbrock (20D)  & $3.418{\scriptstyle\pm0.149}$ & $3.418{\scriptstyle\pm0.149}$ & $\mathbf{3.418{\scriptstyle\pm0.149}}$ & $3.418{\scriptstyle\pm0.149}$ & $3.411{\scriptstyle\pm0.151}$ \\
\midrule
\textbf{\# Wins}  & 0              & 5              & 3              & 5              & 2 \\
\bottomrule
\end{tabular}%
}
\end{table}

\subsection{Distance Weight Function}
\label{app:ablation_weight}

The weight $w_j = h(d_{ij}; \alpha_m)$ on each nearby high-fidelity point controls how quickly trust in the HF prediction decays with distance. We sweep ten candidate forms:
\begin{enumerate}
    \item $h(d; \alpha) = 1 - d^\alpha$ (cost-aware power law, default; $\alpha_m = \log_{10}(C_m)/2$),
    \item $h(d; \alpha) = \exp(-\alpha d)$ (exponential decay),
    \item $h(d; \alpha) = \exp(-\alpha d^2)$ (Gaussian in distance),
    \item $h(d; \alpha) = 1/(1 + d^\alpha)$ (inverse power),
    \item $h(d) = \max(1 - d/r_i, 0)$ (triangular, cost-independent),
    \item $h(d) = 1 - d$ (linear, cost-independent),
    \item $h(d; \alpha) = 1 - d^{2\alpha}$ (steeper cost-aware power law),
    \item $h(d) = \exp(-d^2)$ (RBF-kernel form, cost-independent),
    \item $h(d) = 1$ (identity; collapses Stage~2 to HF-only).
\end{enumerate}
Options W1, W3, W4, W7 retain the fidelity-cost coupling through $\alpha_m$; the remainder test whether that coupling is essential to MAST's behaviour. W8 is equivalent to a standard Gaussian RBF kernel on distance. W9 is a genuine uniform-trust limit ($h = 1$), which sets every per-neighbour weight to one and collapses Stage~2 to the HF-only fall-back.

\begin{table}[h]
\centering
\caption{Distance weight function ablation --- normalised RMSE per benchmark, mean $\pm$ std across 25 seeds; lower is better. Best variant per function in bold.}
\label{tab:ablation_weight}
\resizebox{\textwidth}{!}{%
\begin{tabular}{l c c c c c c c c c}
\toprule
Test function & W1: $1{-}d^\alpha$ (def.) & W2: $e^{-\alpha d}$ & W3: $e^{-\alpha d^2}$ & W4: $(1{+}d^\alpha)^{-1}$ & W5: triangular & W6: $1{-}d$ & W7: $1{-}d^{2\alpha}$ & W8: $e^{-d^2}$ & W9: $h{=}1$ \\
\midrule
Branin (2D)       & $1.192{\scriptstyle\pm0.397}$ & $1.206{\scriptstyle\pm0.427}$ & $1.284{\scriptstyle\pm0.485}$ & $\mathbf{1.112{\scriptstyle\pm0.339}}$ & $1.236{\scriptstyle\pm0.417}$ & $1.113{\scriptstyle\pm0.344}$ & $1.113{\scriptstyle\pm0.344}$ & $1.231{\scriptstyle\pm0.444}$ & $1.346{\scriptstyle\pm0.524}$ \\
Hartmann-3 (3D)   & $0.692{\scriptstyle\pm0.188}$ & $1.321{\scriptstyle\pm0.431}$ & $1.417{\scriptstyle\pm0.479}$ & $1.066{\scriptstyle\pm0.325}$ & $\mathbf{0.576{\scriptstyle\pm0.150}}$ & $1.035{\scriptstyle\pm0.308}$ & $1.035{\scriptstyle\pm0.308}$ & $1.341{\scriptstyle\pm0.437}$ & $1.481{\scriptstyle\pm0.524}$ \\
Ackley (4D)       & $0.945{\scriptstyle\pm0.256}$ & $1.204{\scriptstyle\pm0.389}$ & $1.271{\scriptstyle\pm0.414}$ & $1.026{\scriptstyle\pm0.305}$ & $\mathbf{0.931{\scriptstyle\pm0.243}}$ & $1.005{\scriptstyle\pm0.304}$ & $1.005{\scriptstyle\pm0.304}$ & $1.208{\scriptstyle\pm0.398}$ & $1.327{\scriptstyle\pm0.422}$ \\
Park-1 (4D)       & $0.970{\scriptstyle\pm0.475}$ & $1.362{\scriptstyle\pm0.570}$ & $1.482{\scriptstyle\pm0.613}$ & $1.145{\scriptstyle\pm0.506}$ & $\mathbf{0.969{\scriptstyle\pm0.483}}$ & $1.060{\scriptstyle\pm0.479}$ & $1.059{\scriptstyle\pm0.479}$ & $1.353{\scriptstyle\pm0.560}$ & $1.614{\scriptstyle\pm0.677}$ \\
Park-2 (4D)       & $0.644{\scriptstyle\pm0.177}$ & $1.285{\scriptstyle\pm0.404}$ & $1.408{\scriptstyle\pm0.451}$ & $1.020{\scriptstyle\pm0.306}$ & $\mathbf{0.591{\scriptstyle\pm0.154}}$ & $0.880{\scriptstyle\pm0.249}$ & $0.880{\scriptstyle\pm0.249}$ & $1.269{\scriptstyle\pm0.391}$ & $1.566{\scriptstyle\pm0.522}$ \\
Hartmann-6 (6D)   & $\mathbf{0.882{\scriptstyle\pm0.193}}$ & $1.049{\scriptstyle\pm0.154}$ & $1.090{\scriptstyle\pm0.160}$ & $0.962{\scriptstyle\pm0.156}$ & $0.882{\scriptstyle\pm0.183}$ & $0.915{\scriptstyle\pm0.169}$ & $0.915{\scriptstyle\pm0.169}$ & $1.024{\scriptstyle\pm0.153}$ & $1.129{\scriptstyle\pm0.156}$ \\
Levy (7D)         & $0.783{\scriptstyle\pm0.080}$ & $0.970{\scriptstyle\pm0.095}$ & $1.005{\scriptstyle\pm0.103}$ & $0.880{\scriptstyle\pm0.095}$ & $\mathbf{0.781{\scriptstyle\pm0.081}}$ & $0.814{\scriptstyle\pm0.082}$ & $0.814{\scriptstyle\pm0.082}$ & $0.936{\scriptstyle\pm0.094}$ & $1.067{\scriptstyle\pm0.112}$ \\
Borehole (8D)     & $\mathbf{0.888{\scriptstyle\pm0.240}}$ & $1.335{\scriptstyle\pm0.415}$ & $1.409{\scriptstyle\pm0.443}$ & $1.162{\scriptstyle\pm0.354}$ & $0.902{\scriptstyle\pm0.239}$ & $0.899{\scriptstyle\pm0.282}$ & $0.899{\scriptstyle\pm0.282}$ & $1.209{\scriptstyle\pm0.376}$ & $1.662{\scriptstyle\pm0.536}$ \\
Rastrigin (15D)   & $0.947{\scriptstyle\pm0.061}$ & $0.987{\scriptstyle\pm0.071}$ & $0.985{\scriptstyle\pm0.071}$ & $0.958{\scriptstyle\pm0.063}$ & $0.943{\scriptstyle\pm0.058}$ & $\mathbf{0.938{\scriptstyle\pm0.058}}$ & $0.940{\scriptstyle\pm0.056}$ & $0.941{\scriptstyle\pm0.054}$ & $1.085{\scriptstyle\pm0.093}$ \\
Rosenbrock (20D)  & $\mathbf{0.356{\scriptstyle\pm0.025}}$ & $0.623{\scriptstyle\pm0.086}$ & $0.547{\scriptstyle\pm0.071}$ & $0.564{\scriptstyle\pm0.074}$ & $0.356{\scriptstyle\pm0.025}$ & $0.356{\scriptstyle\pm0.027}$ & $0.356{\scriptstyle\pm0.027}$ & $0.388{\scriptstyle\pm0.032}$ & $1.086{\scriptstyle\pm0.181}$ \\
\midrule
\textbf{\# Wins}  & 3              & 0     & 0     & 1              & 5              & 1              & 0     & 0     & 0 \\
\bottomrule
\end{tabular}%
}
\end{table}

\begin{table}[h]
\centering
\caption{Distance weight function ablation --- normalised mean PDF per benchmark, mean $\pm$ std across 25 seeds; higher is better. Best variant per function in bold.}
\label{tab:ablation_weight_pdf}
\resizebox{\textwidth}{!}{%
\begin{tabular}{l c c c c c c c c c}
\toprule
Test function & W1: $1{-}d^\alpha$ (def.) & W2: $e^{-\alpha d}$ & W3: $e^{-\alpha d^2}$ & W4: $(1{+}d^\alpha)^{-1}$ & W5: triangular & W6: $1{-}d$ & W7: $1{-}d^{2\alpha}$ & W8: $e^{-d^2}$ & W9: $h{=}1$ \\
\midrule
Branin (2D)       & $1.279{\scriptstyle\pm0.427}$ & $1.203{\scriptstyle\pm0.471}$ & $1.131{\scriptstyle\pm0.458}$ & $\mathbf{1.334{\scriptstyle\pm0.458}}$ & $1.245{\scriptstyle\pm0.408}$ & $1.287{\scriptstyle\pm0.462}$ & $1.287{\scriptstyle\pm0.462}$ & $1.167{\scriptstyle\pm0.465}$ & $1.089{\scriptstyle\pm0.458}$ \\
Hartmann-3 (3D)   & $1.826{\scriptstyle\pm0.500}$ & $1.228{\scriptstyle\pm0.418}$ & $1.156{\scriptstyle\pm0.418}$ & $1.411{\scriptstyle\pm0.428}$ & $\mathbf{2.079{\scriptstyle\pm0.539}}$ & $1.413{\scriptstyle\pm0.426}$ & $1.413{\scriptstyle\pm0.426}$ & $1.206{\scriptstyle\pm0.419}$ & $1.124{\scriptstyle\pm0.409}$ \\
Ackley (4D)       & $\mathbf{1.229{\scriptstyle\pm0.321}}$ & $1.051{\scriptstyle\pm0.232}$ & $0.996{\scriptstyle\pm0.227}$ & $1.189{\scriptstyle\pm0.259}$ & $1.224{\scriptstyle\pm0.319}$ & $1.214{\scriptstyle\pm0.285}$ & $1.214{\scriptstyle\pm0.285}$ & $1.048{\scriptstyle\pm0.238}$ & $0.951{\scriptstyle\pm0.209}$ \\
Park-1 (4D)       & $\mathbf{1.460{\scriptstyle\pm0.494}}$ & $1.086{\scriptstyle\pm0.359}$ & $1.015{\scriptstyle\pm0.349}$ & $1.259{\scriptstyle\pm0.399}$ & $1.457{\scriptstyle\pm0.488}$ & $1.324{\scriptstyle\pm0.421}$ & $1.324{\scriptstyle\pm0.421}$ & $1.089{\scriptstyle\pm0.361}$ & $0.947{\scriptstyle\pm0.346}$ \\
Park-2 (4D)       & $1.590{\scriptstyle\pm0.284}$ & $0.944{\scriptstyle\pm0.180}$ & $0.870{\scriptstyle\pm0.168}$ & $1.147{\scriptstyle\pm0.204}$ & $\mathbf{1.700{\scriptstyle\pm0.304}}$ & $1.263{\scriptstyle\pm0.222}$ & $1.263{\scriptstyle\pm0.222}$ & $0.947{\scriptstyle\pm0.181}$ & $0.797{\scriptstyle\pm0.158}$ \\
Hartmann-6 (6D)   & $\mathbf{1.792{\scriptstyle\pm0.523}}$ & $1.732{\scriptstyle\pm0.578}$ & $1.692{\scriptstyle\pm0.575}$ & $1.757{\scriptstyle\pm0.550}$ & $1.789{\scriptstyle\pm0.532}$ & $1.790{\scriptstyle\pm0.554}$ & $1.791{\scriptstyle\pm0.554}$ & $1.712{\scriptstyle\pm0.553}$ & $1.602{\scriptstyle\pm0.557}$ \\
Levy (7D)         & $\mathbf{1.435{\scriptstyle\pm0.162}}$ & $1.194{\scriptstyle\pm0.138}$ & $1.155{\scriptstyle\pm0.146}$ & $1.309{\scriptstyle\pm0.170}$ & $1.432{\scriptstyle\pm0.168}$ & $1.389{\scriptstyle\pm0.159}$ & $1.388{\scriptstyle\pm0.159}$ & $1.232{\scriptstyle\pm0.151}$ & $1.092{\scriptstyle\pm0.145}$ \\
Borehole (8D)     & $1.585{\scriptstyle\pm0.239}$ & $1.283{\scriptstyle\pm0.295}$ & $1.216{\scriptstyle\pm0.284}$ & $1.452{\scriptstyle\pm0.296}$ & $1.539{\scriptstyle\pm0.223}$ & $\mathbf{1.675{\scriptstyle\pm0.279}}$ & $1.673{\scriptstyle\pm0.279}$ & $1.392{\scriptstyle\pm0.299}$ & $1.045{\scriptstyle\pm0.262}$ \\
Rastrigin (15D)   & $1.186{\scriptstyle\pm0.072}$ & $1.202{\scriptstyle\pm0.081}$ & $1.202{\scriptstyle\pm0.081}$ & $1.224{\scriptstyle\pm0.074}$ & $1.189{\scriptstyle\pm0.067}$ & $1.195{\scriptstyle\pm0.068}$ & $1.193{\scriptstyle\pm0.065}$ & $\mathbf{1.229{\scriptstyle\pm0.073}}$ & $1.115{\scriptstyle\pm0.085}$ \\
Rosenbrock (20D)  & $3.418{\scriptstyle\pm0.149}$ & $2.069{\scriptstyle\pm0.285}$ & $2.340{\scriptstyle\pm0.285}$ & $2.283{\scriptstyle\pm0.279}$ & $3.414{\scriptstyle\pm0.154}$ & $\mathbf{3.422{\scriptstyle\pm0.176}}$ & $3.421{\scriptstyle\pm0.176}$ & $3.220{\scriptstyle\pm0.216}$ & $1.187{\scriptstyle\pm0.222}$ \\
\midrule
\textbf{\# Wins}  & 4              & 0     & 0     & 1              & 2              & 2              & 0              & 1              & 0 \\
\bottomrule
\end{tabular}%
}
\end{table}

\subsection{Distance Metric}
\label{app:ablation_distance}

The trust region and weights depend on a distance metric applied to inputs in the normalised design space. We sweep six metrics:
\begin{enumerate}
    \item Euclidean, $\|\mathbf{a} - \mathbf{b}\|_2$ (default),
    \item Manhattan, $\|\mathbf{a} - \mathbf{b}\|_1$,
    \item Chebyshev, $\|\mathbf{a} - \mathbf{b}\|_\infty$,
    \item Mahalanobis with ARD-learned lengthscales $\boldsymbol{\ell}$: $\sqrt{\sum_d (a_d - b_d)^2 / \ell_d^2}$,
    \item Normalised Euclidean, $\|\mathbf{a} - \mathbf{b}\|_2 / \sqrt{D}$,
    \item Domain-relative, $\sqrt{\sum_d (a_d - b_d)^2 / w_d^2}$, where $w_d$ is the width of the original domain in dimension $d$.
\end{enumerate}
The Mahalanobis variant targets the high-dimensional pathology of plain Euclidean distance. ARD lengthscales give dimensions with weak functional variation a smaller effective weight, focusing the trust-weighting on directions where the HF signal actually varies. The distance-metric slot is a drop-in replacement, and a categorical-input benchmark in Appendix~\ref{app:rw_hoip} uses the Hamming distance instead -- demonstrating that MAST extends beyond continuous Euclidean inputs without any change to the remainder of the pipeline.

\begin{table}[h]
\centering
\caption{Distance-metric ablation --- normalised RMSE per benchmark, mean $\pm$ std across 25 seeds; lower is better. Best variant per function in bold. D6 (Domain-relative) reduces to D1 after Stage-1 input normalisation to $[0,1]^D$ and is included as an implementation sanity check; its column matches D1 by construction.}
\label{tab:ablation_distance}
\resizebox{\textwidth}{!}{%
\begin{tabular}{l c c c c c c}
\toprule
Test function & D1: Euclidean (default) & D2: Manhattan & D3: Chebyshev & D4: Mahalanobis (ARD) & D5: Norm.\ Euclidean & D6: Domain-relative \\
\midrule
Branin (2D)       & $1.192{\scriptstyle\pm0.397}$ & $1.228{\scriptstyle\pm0.422}$ & $1.184{\scriptstyle\pm0.392}$ & $1.342{\scriptstyle\pm0.503}$ & $\mathbf{1.162{\scriptstyle\pm0.378}}$ & $1.192{\scriptstyle\pm0.397}$ \\
Hartmann-3 (3D)   & $0.692{\scriptstyle\pm0.188}$ & $0.589{\scriptstyle\pm0.153}$ & $0.745{\scriptstyle\pm0.205}$ & $\mathbf{0.430{\scriptstyle\pm0.139}}$ & $0.849{\scriptstyle\pm0.253}$ & $0.692{\scriptstyle\pm0.188}$ \\
Ackley (4D)       & $0.945{\scriptstyle\pm0.256}$ & $\mathbf{0.936{\scriptstyle\pm0.230}}$ & $0.941{\scriptstyle\pm0.266}$ & $0.954{\scriptstyle\pm0.251}$ & $0.970{\scriptstyle\pm0.279}$ & $0.945{\scriptstyle\pm0.256}$ \\
Park-1 (4D)       & $\mathbf{0.970{\scriptstyle\pm0.475}}$ & $0.989{\scriptstyle\pm0.483}$ & $0.982{\scriptstyle\pm0.479}$ & $0.977{\scriptstyle\pm0.474}$ & $1.014{\scriptstyle\pm0.477}$ & $\mathbf{0.970{\scriptstyle\pm0.475}}$ \\
Park-2 (4D)       & $0.644{\scriptstyle\pm0.177}$ & $\mathbf{0.576{\scriptstyle\pm0.141}}$ & $0.710{\scriptstyle\pm0.212}$ & $0.667{\scriptstyle\pm0.175}$ & $0.796{\scriptstyle\pm0.228}$ & $0.644{\scriptstyle\pm0.177}$ \\
Hartmann-6 (6D)   & $0.882{\scriptstyle\pm0.193}$ & $\mathbf{0.873{\scriptstyle\pm0.177}}$ & $0.887{\scriptstyle\pm0.183}$ & $0.888{\scriptstyle\pm0.193}$ & $0.909{\scriptstyle\pm0.168}$ & $0.882{\scriptstyle\pm0.193}$ \\
Levy (7D)         & $\mathbf{0.783{\scriptstyle\pm0.080}}$ & $0.790{\scriptstyle\pm0.092}$ & $0.804{\scriptstyle\pm0.084}$ & $0.790{\scriptstyle\pm0.092}$ & $0.825{\scriptstyle\pm0.087}$ & $\mathbf{0.783{\scriptstyle\pm0.080}}$ \\
Borehole (8D)     & $\mathbf{0.888{\scriptstyle\pm0.240}}$ & $0.986{\scriptstyle\pm0.245}$ & $0.901{\scriptstyle\pm0.278}$ & $0.971{\scriptstyle\pm0.288}$ & $0.995{\scriptstyle\pm0.304}$ & $0.996{\scriptstyle\pm0.249}$ \\
Rastrigin (15D)   & $0.947{\scriptstyle\pm0.061}$ & $0.947{\scriptstyle\pm0.059}$ & $\mathbf{0.934{\scriptstyle\pm0.052}}$ & $0.946{\scriptstyle\pm0.057}$ & $0.940{\scriptstyle\pm0.053}$ & $0.947{\scriptstyle\pm0.061}$ \\
Rosenbrock (20D)  & $0.356{\scriptstyle\pm0.025}$ & $0.356{\scriptstyle\pm0.025}$ & $0.372{\scriptstyle\pm0.029}$ & $\mathbf{0.354{\scriptstyle\pm0.026}}$ & $0.474{\scriptstyle\pm0.056}$ & $0.356{\scriptstyle\pm0.025}$ \\
\midrule
\textbf{\# Wins}  & 3              & 3              & 1              & 2              & 1              & 2 \\
\bottomrule
\end{tabular}%
}
\end{table}

\begin{table}[h]
\centering
\caption{Distance-metric ablation --- normalised mean PDF per benchmark, mean $\pm$ std across 25 seeds; higher is better. Best variant per function in bold. D6 (Domain-relative) reduces to D1 after Stage-1 input normalisation to $[0,1]^D$ and is included as an implementation sanity check; its column matches D1 by construction.}
\label{tab:ablation_distance_pdf}
\resizebox{\textwidth}{!}{%
\begin{tabular}{l c c c c c c}
\toprule
Test function & D1: Euclidean (default) & D2: Manhattan & D3: Chebyshev & D4: Mahalanobis (ARD) & D5: Norm.\ Euclidean & D6: Domain-relative \\
\midrule
Branin (2D)       & $1.279{\scriptstyle\pm0.427}$ & $1.278{\scriptstyle\pm0.434}$ & $1.287{\scriptstyle\pm0.424}$ & $1.236{\scriptstyle\pm0.433}$ & $\mathbf{1.299{\scriptstyle\pm0.432}}$ & $1.279{\scriptstyle\pm0.427}$ \\
Hartmann-3 (3D)   & $1.826{\scriptstyle\pm0.500}$ & $2.058{\scriptstyle\pm0.562}$ & $1.742{\scriptstyle\pm0.483}$ & $\mathbf{2.697{\scriptstyle\pm0.695}}$ & $1.614{\scriptstyle\pm0.453}$ & $1.826{\scriptstyle\pm0.500}$ \\
Ackley (4D)       & $1.229{\scriptstyle\pm0.321}$ & $1.196{\scriptstyle\pm0.282}$ & $\mathbf{1.251{\scriptstyle\pm0.317}}$ & $1.156{\scriptstyle\pm0.286}$ & $1.235{\scriptstyle\pm0.306}$ & $1.229{\scriptstyle\pm0.321}$ \\
Park-1 (4D)       & $1.460{\scriptstyle\pm0.494}$ & $1.428{\scriptstyle\pm0.485}$ & $1.452{\scriptstyle\pm0.485}$ & $\mathbf{1.462{\scriptstyle\pm0.506}}$ & $1.408{\scriptstyle\pm0.460}$ & $1.460{\scriptstyle\pm0.494}$ \\
Park-2 (4D)       & $1.590{\scriptstyle\pm0.284}$ & $\mathbf{1.725{\scriptstyle\pm0.318}}$ & $1.499{\scriptstyle\pm0.268}$ & $1.548{\scriptstyle\pm0.272}$ & $1.373{\scriptstyle\pm0.235}$ & $1.590{\scriptstyle\pm0.284}$ \\
Hartmann-6 (6D)   & $\mathbf{1.792{\scriptstyle\pm0.523}}$ & $1.741{\scriptstyle\pm0.530}$ & $1.785{\scriptstyle\pm0.515}$ & $1.731{\scriptstyle\pm0.533}$ & $1.771{\scriptstyle\pm0.517}$ & $\mathbf{1.792{\scriptstyle\pm0.523}}$ \\
Levy (7D)         & $\mathbf{1.435{\scriptstyle\pm0.162}}$ & $1.410{\scriptstyle\pm0.167}$ & $1.404{\scriptstyle\pm0.166}$ & $1.408{\scriptstyle\pm0.173}$ & $1.378{\scriptstyle\pm0.166}$ & $\mathbf{1.435{\scriptstyle\pm0.162}}$ \\
Borehole (8D)     & $1.585{\scriptstyle\pm0.239}$ & $1.312{\scriptstyle\pm0.211}$ & $\mathbf{1.695{\scriptstyle\pm0.291}}$ & $1.636{\scriptstyle\pm0.300}$ & $1.616{\scriptstyle\pm0.300}$ & $1.281{\scriptstyle\pm0.206}$ \\
Rastrigin (15D)   & $1.186{\scriptstyle\pm0.072}$ & $1.183{\scriptstyle\pm0.071}$ & $1.219{\scriptstyle\pm0.068}$ & $1.186{\scriptstyle\pm0.069}$ & $\mathbf{1.232{\scriptstyle\pm0.071}}$ & $1.186{\scriptstyle\pm0.072}$ \\
Rosenbrock (20D)  & $3.418{\scriptstyle\pm0.149}$ & $3.411{\scriptstyle\pm0.151}$ & $3.345{\scriptstyle\pm0.178}$ & $\mathbf{3.440{\scriptstyle\pm0.180}}$ & $2.688{\scriptstyle\pm0.268}$ & $3.418{\scriptstyle\pm0.149}$ \\
\midrule
\textbf{\# Wins}  & 2              & 1              & 2              & 3              & 2              & 2 \\
\bottomrule
\end{tabular}%
}
\end{table}

\subsection{Aggregated Findings}
\label{app:ablation_aggregate}

The ten-function ablation isolates which of MAST's three design knobs carries the surrogate. The weight function is by far the most influential axis, with a mean swing of 0.53 in normalised RMSE and 0.64 in mean PDF across its ten variants. Distance metric and trust-region radius are both 4 to 5$\times$ smaller in RMSE influence (mean swings of 0.12 and 0.10). The counterintuitive lead is that the most-influential axis carries the most-robust default. The cost-aware power-law weight sits within 10\% of the best non-default on 9 of 10 functions and within 30\% on all 10, with a mean gap-to-best of 3.9\%. The default Euclidean distance and square-root radius are each within 10\% of best on 8 of 10 functions. The ablation also identifies where non-defaults are decisive rather than marginal. Mahalanobis-ARD against Euclidean is a near-no-op on most of the roster but cuts Hartmann3 RMSE by 38\% and boosts its mean PDF by 48\%, the function whose discrepancy is a fixed shift of the $\boldsymbol\alpha$ vector and the case where ARD lengthscales most cleanly identify the active directions. The HF-only collapse weight (zero geometric content) is the worst non-default on 10 of 10 functions for both metrics, and the Gaussian RBF weight underperforms the default on 9 of 10. These two anti-cases separate the value of the spatial-weighting mechanism itself from the choice of weight family.

A practitioner installing MAST with the published default configuration (square-root radius, cost-aware power-law weight, Euclidean distance) is in a sensible place. The default is within 10\% of the best non-default on 25 of 30 (axis $\times$ function) RMSE pairs and 26 of 30 PDF pairs across the suite. When MAST under-performs on a new problem, the first knob to swap is the distance metric. Mahalanobis-ARD is the right substitution when the Stage-1 ARD lengthscales differ markedly across input dimensions; on Hartmann3 this swap recovers a 38\% RMSE reduction, while on more isotropic landscapes it costs up to 13\% (Branin) or 9\% (Borehole) and is best avoided. The second knob is the weight function: the cost-independent linear variant (W5) is the most reliable non-default, strictly beating the default RMSE on 6 of 10 functions with a mean gain of 2\% and a single-function high of 17\% on Hartmann3. The trust-region radius does not warrant tuning except when the HF sample count is very small, in which case the global trust-region variant is the right substitution. Two variants are anti-recommendations: never use the HF-only collapse weight (worst on every function, both metrics) or the Gaussian RBF weight (below default on 9 of 10).


\section{Performance Analysis on Synthetic and Real-World Benchmarks}
\label{app:pa_combined}

\subsection{Setup}
\label{app:pa_combined_setup}

This appendix presents the full performance-analysis sweeps that underpin the three-function summaries in \S\ref{sec:results}. The synthetic suite covers the ten test functions of Appendix~\ref{app:test_functions} organised into four function groups: Branin (2D), Hartmann3 (3D), Ackley (4D); Park1 (4D), Park2 (4D), Hartmann6 (6D); Levy (7D), Borehole (8D), Rastrigin (15D); and Rosenbrock (20D) on its own. The real-world suite covers Concrete (8D), HOIP (3D, three-fidelity), and DrivAerNet (23D), described in Appendix~\ref{app:rw_datasets}. Each benchmark is stress-tested under three independent perturbations -- budget allocation, total budget, and fidelity discrepancy or LF noise -- each holding all other settings fixed at the §5 baseline (cost ratio $\mathcal{C}_{\mathrm{HF}}:\mathcal{C}_{\mathrm{LF}} = 10:1$, base budget $\mathcal{B} = 5D$, 25 seeds, identical baselines). Synthetic figures and real-world figures are interleaved within each stress-test subsection so the two evaluation modes can be read against each other.


\subsection{Sensitivity to Budget Allocation}
\label{app:pa_combined_ratio}

\begin{figure}[!htb]
\centering
\includegraphics[width=\textwidth]{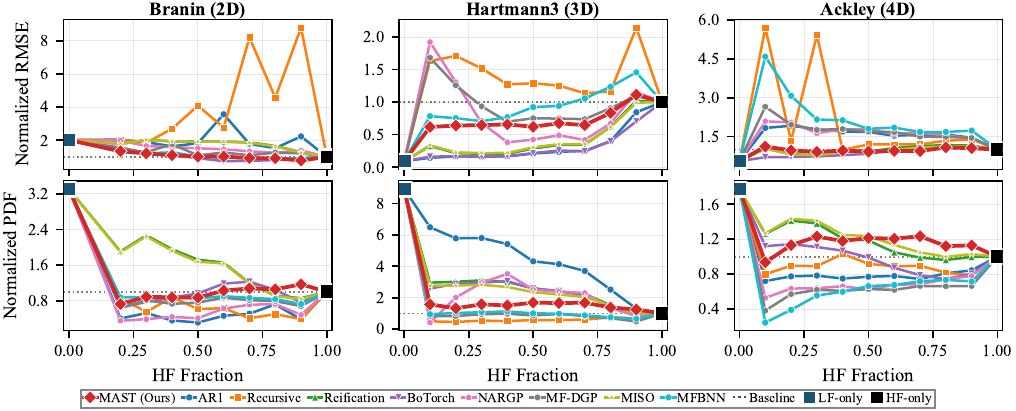}
\caption{Synthetic budget-allocation sweep, group 1: Branin (2D), Hartmann3 (3D), Ackley (4D). Top: normalised RMSE; bottom: normalised mean PDF.}
\label{fig:pa_ratio_1}
\end{figure}

\begin{figure}[!htb]
\centering
\includegraphics[width=\textwidth]{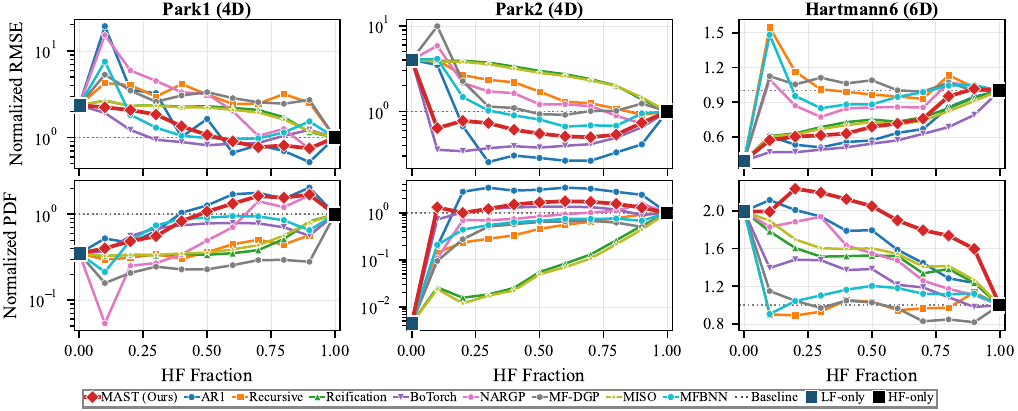}
\caption{Synthetic budget-allocation sweep, group 2: Park1 (4D), Park2 (4D), Hartmann6 (6D).}
\label{fig:pa_ratio_2}
\end{figure}

\begin{figure}[!htb]
\centering
\includegraphics[width=\textwidth]{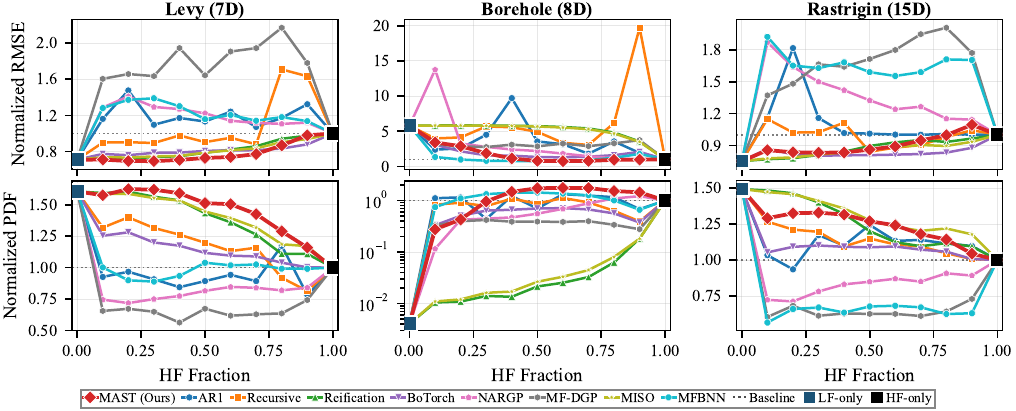}
\caption{Synthetic budget-allocation sweep, group 3: Levy (7D), Borehole (8D), Rastrigin (15D).}
\label{fig:pa_ratio_3}
\end{figure}

\begin{figure}[!htb]
\centering
\includegraphics[width=\textwidth]{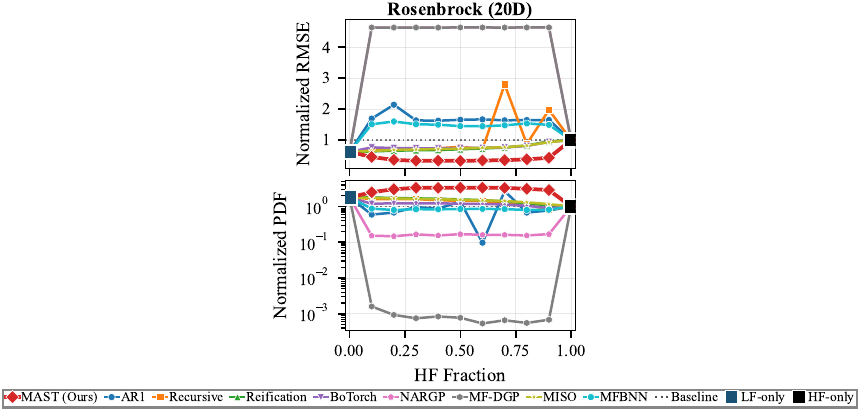}
\caption{Synthetic budget-allocation sweep, group 4: Rosenbrock (20D).}
\label{fig:pa_ratio_4}
\end{figure}

\begin{figure}[!htb]
\centering
\includegraphics[width=\textwidth]{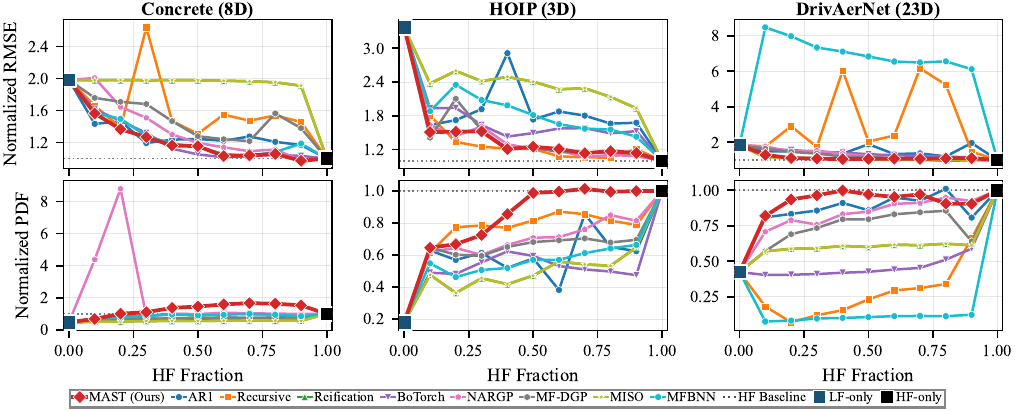}
\caption{Real-world budget-allocation sweep: Concrete (8D), HOIP (3D), DrivAerNet (23D).}
\label{fig:rw_pa_ratio_1}
\end{figure}

\begin{figure}[!htb]
\centering
\includegraphics[width=\textwidth]{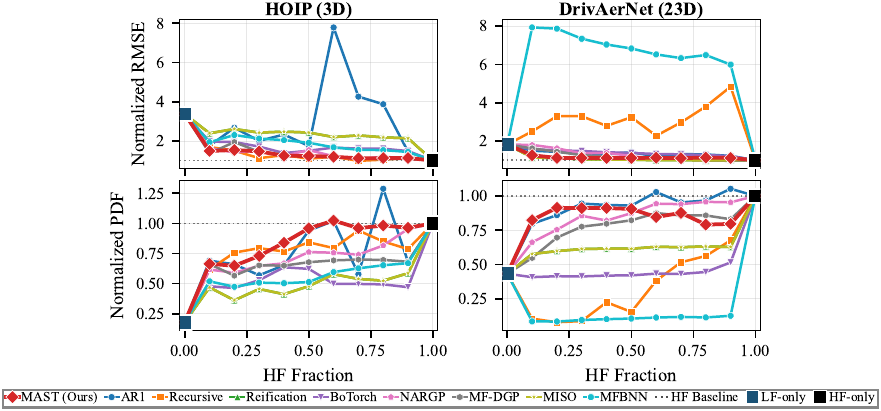}
\caption{Real-world budget-allocation sweep: HOIP and DrivAerNet drawn from independent training pools at each fidelity, in contrast to the collocated pairing in Figure~\ref{fig:rw_pa_ratio_1}.}
\label{fig:rw_pa_ratio_2}
\end{figure}

The HF cost fraction is varied over $\{0.0, 0.1, \ldots, 1.0\}$ at fixed total budget $\mathcal{B}=5D$, with $\mathrm{HF}=0$ and $\mathrm{HF}=1$ retained as pure-LF and pure-HF anchors. Figures~\ref{fig:pa_ratio_1}--\ref{fig:pa_ratio_4} show the synthetic sweep across the four function groups; Figures~\ref{fig:rw_pa_ratio_1}--\ref{fig:rw_pa_ratio_2} show the corresponding real-world sweep.

\FloatBarrier

Across the synthetic suite MAST stays at-or-below baseline RMSE on the higher-dimensional, multimodal benchmarks (Borehole, Rastrigin, Rosenbrock) regardless of the HF allocation, while the learned-correlation methods (AR1, NARGP, MF-DGP) oscillate strongly. On the real-world side, the strongly-correlated analytical-LF benchmarks (Concrete, HOIP) replicate the synthetic narrative: MAST tracks at-or-below baseline on RMSE and dominates calibration. DrivAerNet sits in the leading cluster but does not separate as cleanly from the alternatives. The independent-sampling variants in Figure~\ref{fig:rw_pa_ratio_2} preserve the same qualitative ordering, confirming that MAST's spatial-trust mechanism does not depend on collocation between the HF and LF training pools.

\subsection{Sensitivity to Total Budget}
\label{app:pa_combined_budget}

The total budget is rescaled by $\{0.25, 0.5, 1, 2, 3\}\times\mathcal{B}$ at fixed 70/30 HF/LF cost split, holding all other settings at the §5 defaults. The synthetic and real-world figure groupings mirror \S\ref{app:pa_combined_ratio}.

\begin{figure}[!htb]
\centering
\includegraphics[width=\textwidth]{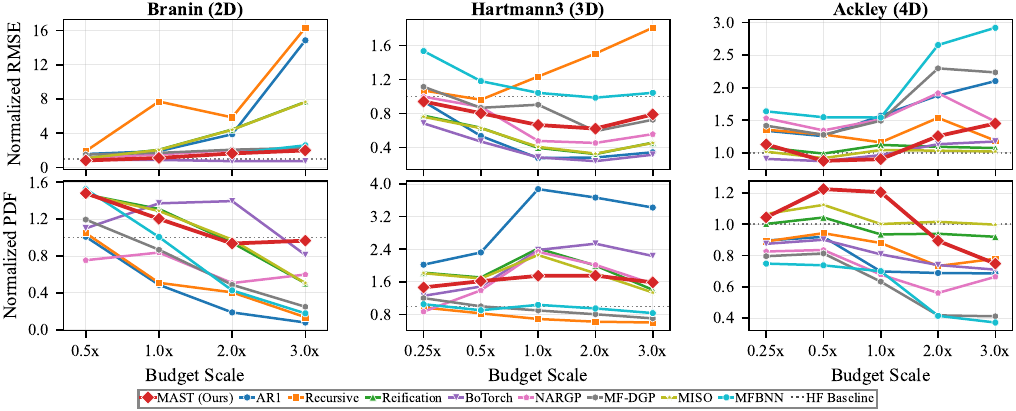}
\caption{Synthetic total-budget sweep, group 1: Branin, Hartmann3, Ackley.}
\label{fig:pa_budget_1}
\end{figure}

\begin{figure}[!htb]
\centering
\includegraphics[width=\textwidth]{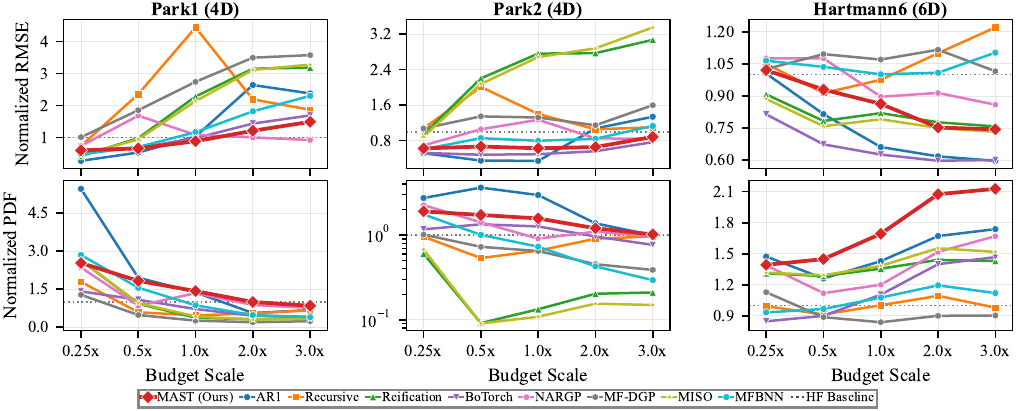}
\caption{Synthetic total-budget sweep, group 2: Park1, Park2, Hartmann6.}
\label{fig:pa_budget_2}
\end{figure}

\begin{figure}[!htb]
\centering
\includegraphics[width=\textwidth]{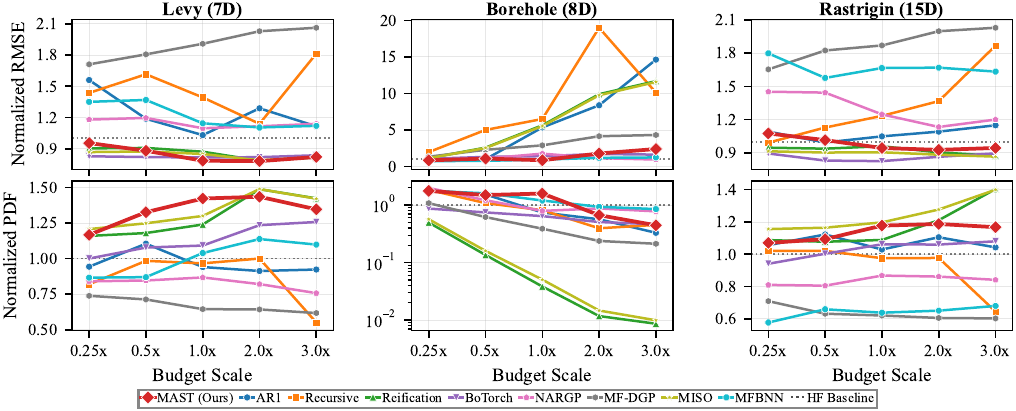}
\caption{Synthetic total-budget sweep, group 3: Levy, Borehole, Rastrigin.}
\label{fig:pa_budget_3}
\end{figure}

\begin{figure}[!htb]
\centering
\includegraphics[width=\textwidth]{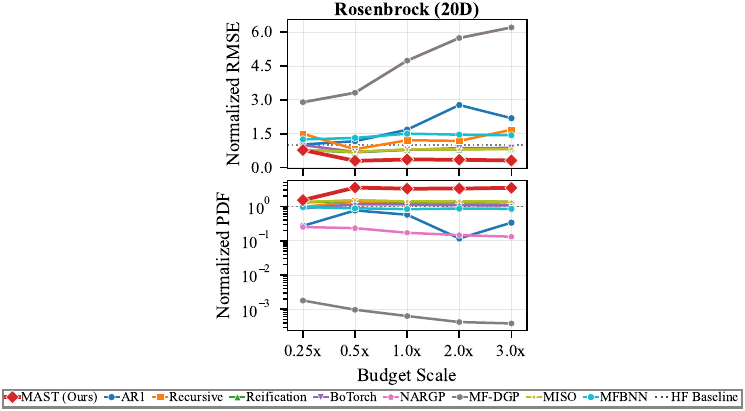}
\caption{Synthetic total-budget sweep, group 4: Rosenbrock.}
\label{fig:pa_budget_4}
\end{figure}

\begin{figure}[!htb]
\centering
\includegraphics[width=\textwidth]{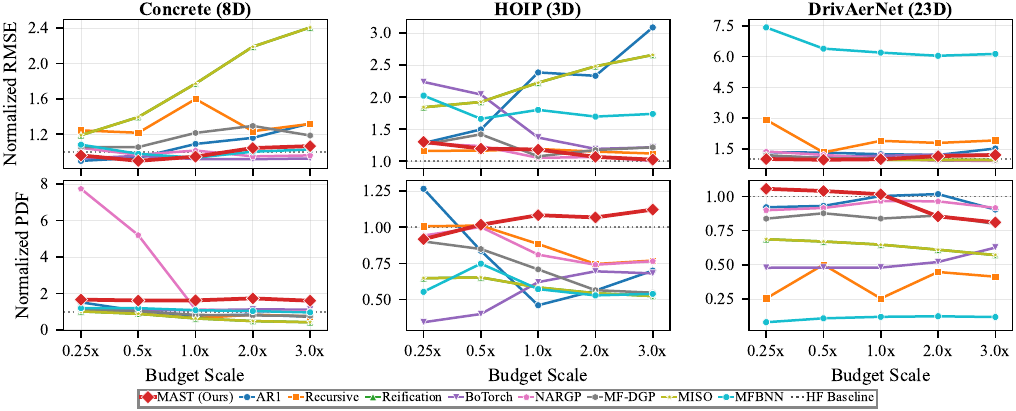}
\caption{Real-world total-budget sweep: Concrete, HOIP, DrivAerNet.}
\label{fig:rw_pa_budget_1}
\end{figure}

\begin{figure}[!htb]
\centering
\includegraphics[width=\textwidth]{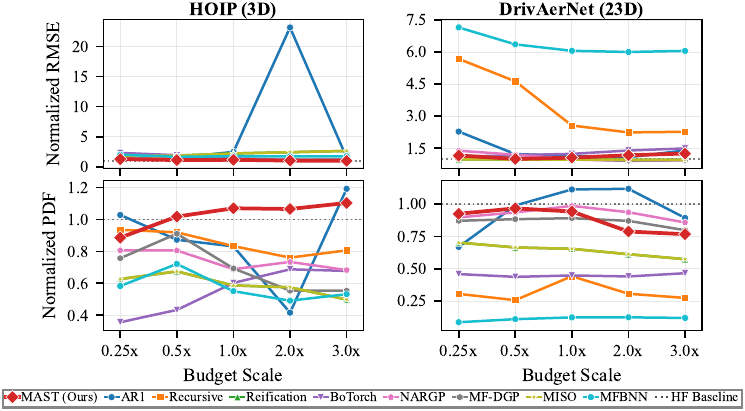}
\caption{Real-world total-budget sweep: HOIP and DrivAerNet drawn from independent pools.}
\label{fig:rw_pa_budget_2}
\end{figure}

\FloatBarrier

On the synthetic suite, MAST holds a 64--70\% RMSE reduction on Rosenbrock from $0.5\times$ budget onwards and stays within the leading cluster on the other higher-dimensional functions; AR1 and the deep-learning methods exhibit much larger budget-driven variance. On Concrete and HOIP MAST tracks the HF-only baseline tightly across the sweep, with the calibration advantage strongest at small budgets. On DrivAerNet the absolute spread between methods narrows as budget grows because the HF-only fit itself becomes accurate; MAST remains within the leading cluster. The synthetic and real-world responses to budget scaling are aligned: the data-scarce edge is where MAST's geometric inductive bias has the largest payoff, and the gap shrinks as the HF-only baseline becomes adequate on its own.

\subsection{Sensitivity to Fidelity Discrepancy}
\label{app:pa_combined_cost}

\begin{figure}[!htb]
\centering
\includegraphics[width=\textwidth]{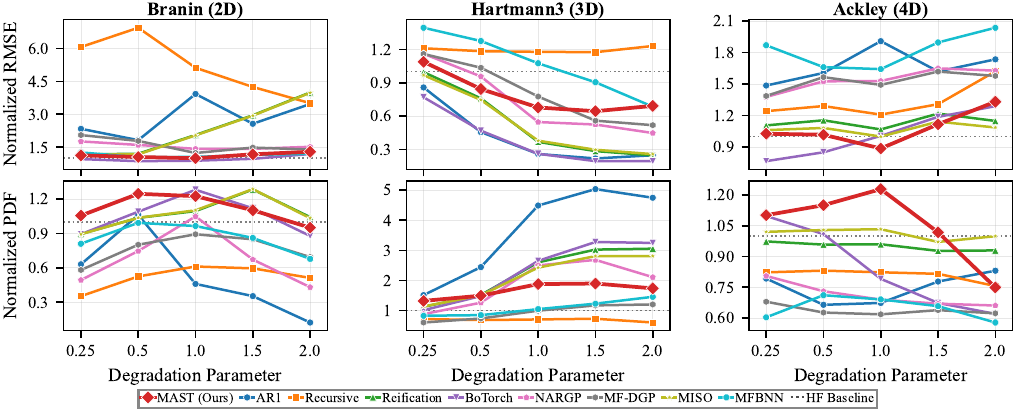}
\caption{Synthetic discrepancy sweep against $d$, group 1: Branin, Hartmann3, Ackley.}
\label{fig:pa_cost_1}
\end{figure}

\begin{figure}[!htb]
\centering
\includegraphics[width=\textwidth]{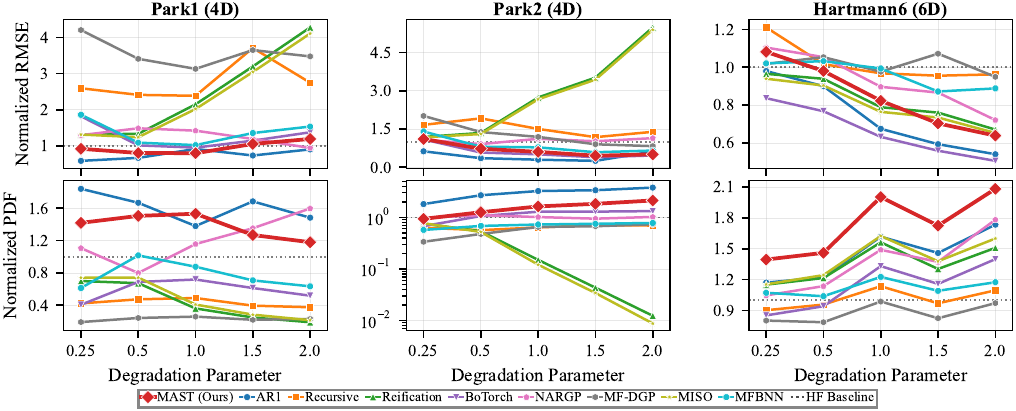}
\caption{Synthetic discrepancy sweep against $d$, group 2: Park1, Park2, Hartmann6.}
\label{fig:pa_cost_2}
\end{figure}

\begin{figure}[!htb]
\centering
\includegraphics[width=\textwidth]{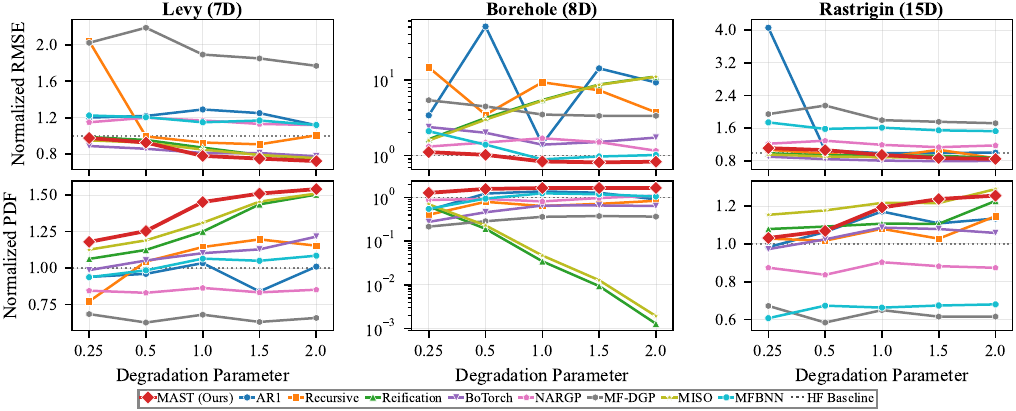}
\caption{Synthetic discrepancy sweep against $d$, group 3: Levy, Borehole, Rastrigin.}
\label{fig:pa_cost_3}
\end{figure}

\begin{figure}[!htb]
\centering
\includegraphics[width=\textwidth]{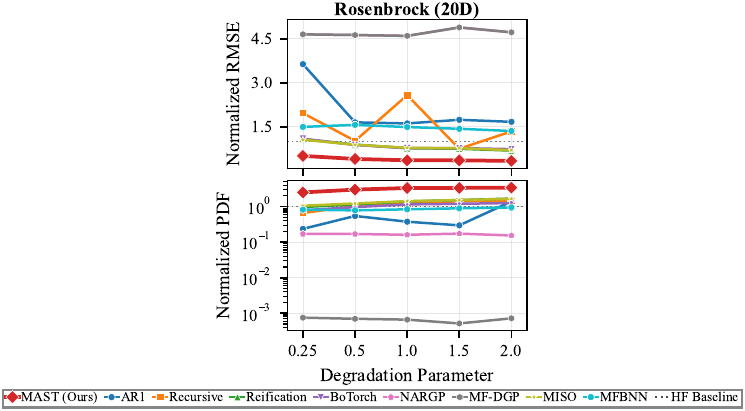}
\caption{Synthetic discrepancy sweep against $d$, group 4: Rosenbrock.}
\label{fig:pa_cost_4}
\end{figure}

\begin{figure}[!htb]
\centering
\includegraphics[width=\textwidth]{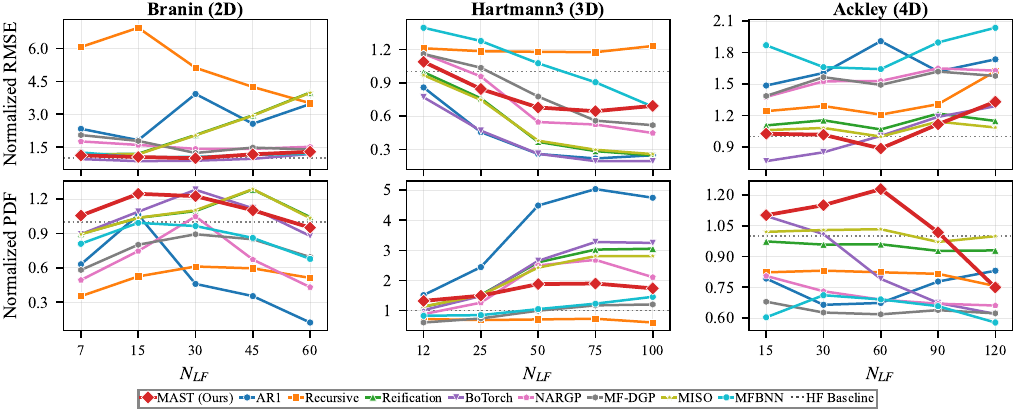}
\caption{Same data as Figure~\ref{fig:pa_cost_1}, with $N_{\mathrm{LF}}$ on the x-axis instead of $d$. The two axes are linked by $N_{\mathrm{LF}}/N_{\mathrm{HF}} = 10\,d$.}
\label{fig:pa_cost_nlf_1}
\end{figure}

\begin{figure}[!htb]
\centering
\includegraphics[width=\textwidth]{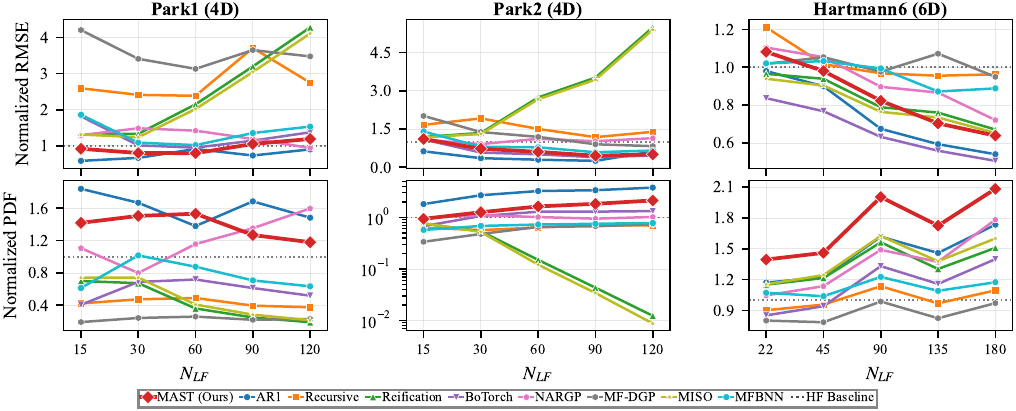}
\caption{Same data as Figure~\ref{fig:pa_cost_2}, plotted against $N_{\mathrm{LF}}$.}
\label{fig:pa_cost_nlf_2}
\end{figure}

\begin{figure}[!htb]
\centering
\includegraphics[width=\textwidth]{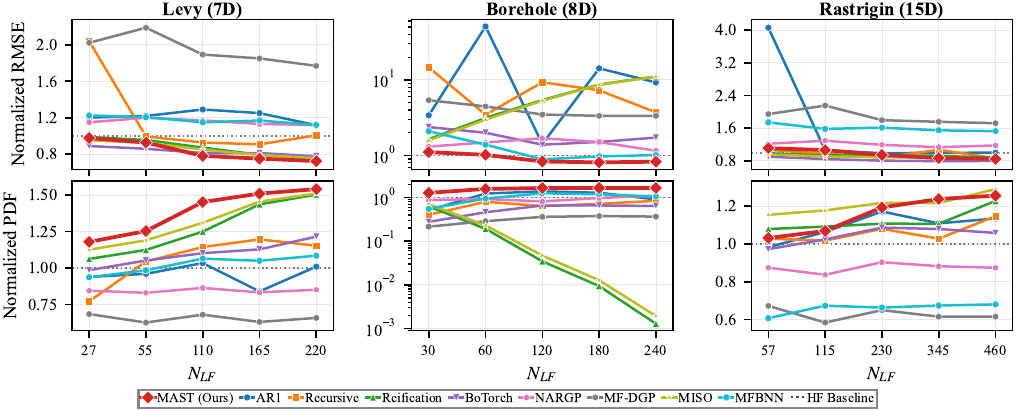}
\caption{Same data as Figure~\ref{fig:pa_cost_3}, plotted against $N_{\mathrm{LF}}$.}
\label{fig:pa_cost_nlf_3}
\end{figure}

\begin{figure}[!htb]
\centering
\includegraphics[width=\textwidth]{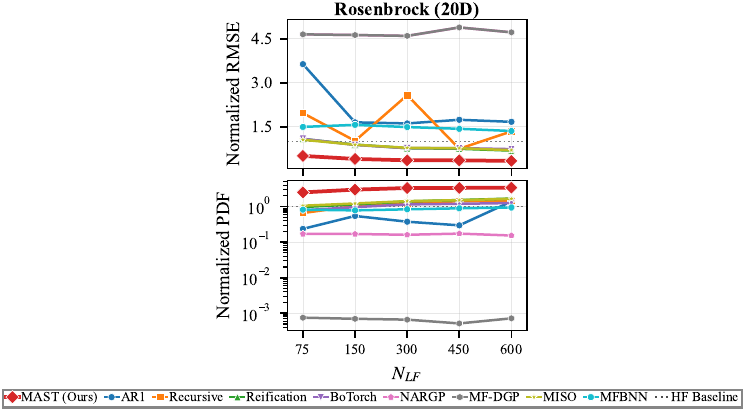}
\caption{Same data as Figure~\ref{fig:pa_cost_4}, plotted against $N_{\mathrm{LF}}$.}
\label{fig:pa_cost_nlf_4}
\end{figure}

On the synthetic side, the cost-controlled degradation parameter $d$ of Eq.~\eqref{eq:cost_controlled} is swept over $\{0.25, 0.5, 1.0, 1.5, 2.0\}$. Increasing $d$ does two things simultaneously: it amplifies the discrepancy between LF and HF, and it scales the LF sample multiplier $N_{\mathrm{LF}}/N_{\mathrm{HF}} = 10\,d$, so the LF budget grows with $d$. Figures~\ref{fig:pa_cost_1}--\ref{fig:pa_cost_4} plot performance against $d$; Figures~\ref{fig:pa_cost_nlf_1}--\ref{fig:pa_cost_nlf_4} replot the same data against $N_{\mathrm{LF}}$ on the x-axis. On the real-world side, the cost-ratio sweep degrades the LF observations by injecting additional Gaussian noise that grows with the cost ratio:
\begin{equation}
\sigma_{\mathrm{LF}}^{\mathrm{extra}}(\mathcal{C}) = \alpha \, \ln(\mathcal{C}) \, \mathrm{std}(y_{\mathrm{LF}}),
\qquad \alpha = 0.1,
\label{eq:rw_pa_noise}
\end{equation}
where $\mathcal{C}$ is the swept HF/LF cost ratio. The two stress tests are conceptually distinct: the synthetic version amplifies a bias term in the LF generator while the real-world version adds noise to existing LF observations.

\begin{figure}[!htb]
\centering
\includegraphics[width=\textwidth]{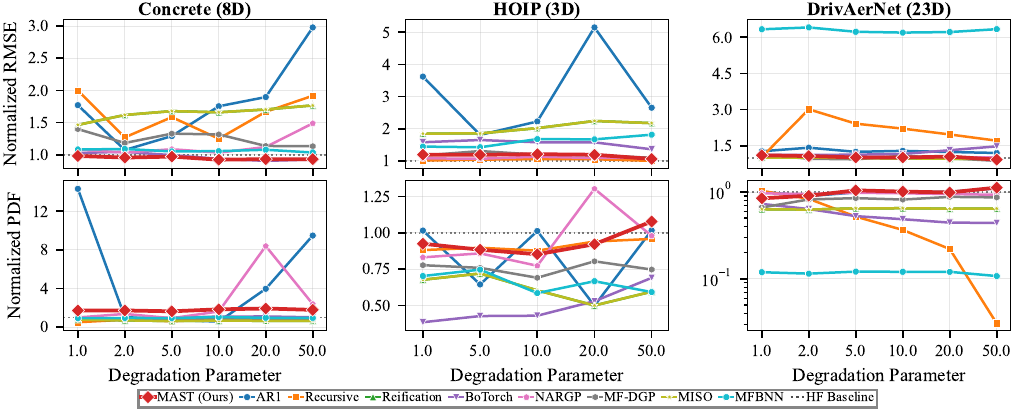}
\caption{Real-world LF-noise-degradation sweep: Concrete, HOIP, DrivAerNet. The x-axis is the HF/LF cost ratio $\mathcal{C}$; the LF observations are corrupted with extra Gaussian noise per Eq.~\eqref{eq:rw_pa_noise}.}
\label{fig:rw_pa_cost_1}
\end{figure}

\begin{figure}[!htb]
\centering
\includegraphics[width=\textwidth]{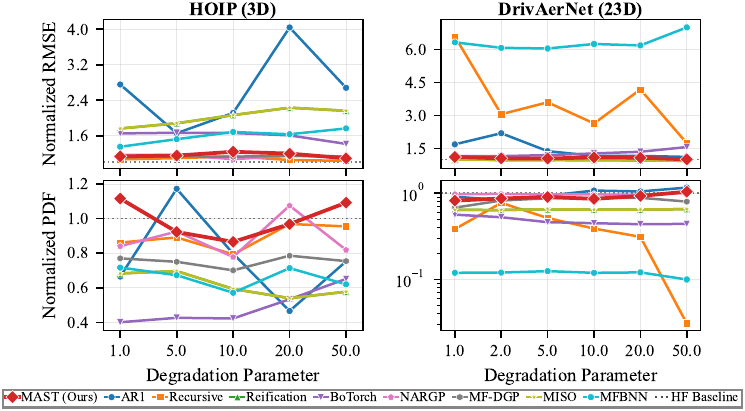}
\caption{Real-world LF-noise-degradation sweep: HOIP and DrivAerNet drawn from independent pools.}
\label{fig:rw_pa_cost_2}
\end{figure}

\FloatBarrier

On the synthetic suite MAST is robust across the full $d$ range: a 49--66\% RMSE reduction on Rosenbrock (deepest at high $d$), $\le 20\%$ improvement on Borehole at $d \ge 1$, and at-or-below baseline on the higher-dimensional benchmarks throughout. AR1 sits well above baseline at every $d$, often by a factor of 2--4$\times$ on Borehole and Rosenbrock with sporadic outlier spikes. On uncertainty calibration MAST exceeds the baseline on every Borehole and Rosenbrock combination across the sweep. On the real-world side, MAST stays within the leading cluster on every benchmark as the LF noise grows; methods that lean heavily on the LF tier degrade quickly, while those that defer to the HF GP recover. MAST's spatial-trust mechanism interpolates between the two extremes and approaches but does not exceed baseline calibration on HOIP under the most aggressive noise injection (where the LF tier becomes too noisy for the discrepancy GP to extract a useful correction).

\subsection{Cross-Cutting Findings}
\label{app:pa_combined_summary}

Three patterns recur across the three stress tests and the two evaluation modes. (i) MAST's RMSE advantage is largest on the higher-dimensional, multimodal synthetic functions (Rosenbrock, Borehole, Rastrigin) and on the strongly-correlated real-world pair (Concrete, HOIP); the advantage shrinks on benchmarks where the HF-only fit is already adequate. (ii) The deep-learning baselines (NARGP, MF-DGP) and the precision-weighted reification fusion are unstable across all three sweeps, with NARGP/MF-DGP often 2--6$\times$ worse than baseline at the data-scarce edge. AR1 is mid-pack but exhibits the largest variance across sweep points, indicating that its scalar correlation parameter struggles to absorb the regime changes. (iii) The synthetic and real-world responses are qualitatively aligned -- MAST's geometric inductive bias transfers from controlled-discrepancy benchmarks to genuine engineering datasets without re-tuning. The fidelity-discrepancy and LF-noise asymmetry between the two evaluation modes (synthetic amplifies a bias term, real-world adds Gaussian noise) is the only structural difference; both stress tests degrade the LF tier, and MAST's response is qualitatively similar in both.

\subsection{Full Tables with Standard Deviations}
\label{app:pdf_tables}

For statistical-significance reporting, we provide the full per-seed standard deviations of the per-row paired ratios summarised in main-body Tables~\ref{tab:results_combined} and~\ref{tab:three_fid_combined}. Standard deviations are computed across the 25 seeds and indicate the cross-seed variability of the normalised metric. Tables~\ref{tab:results_rmse} and~\ref{tab:three_fid_rmse} report normalised RMSE (Eq.~\eqref{eq:rmse_def}) on the two-fidelity and three-fidelity benchmark suites respectively, with values below $1$ indicating improvement over the cost-equivalent HF-only baseline. Tables~\ref{tab:results_pdf} and~\ref{tab:three_fid_pdf} report the corresponding normalised mean predictive density (mean PDF, Eq.~\eqref{eq:meanpdf_def}); values above $1$ indicate improved uncertainty calibration. Row order and column order are identical across all four tables and across the main-body combined tables. The normalised $R^2$ counterparts (Tables~\ref{tab:results_r2} and~\ref{tab:three_fid_r2}) are reported in Appendix~\ref{app:r2_extra}.

\begin{table}[!t]
\centering
\caption{Two-fidelity benchmarks: \textbf{normalised RMSE} ($\downarrow$, lower is better; $<1$ beats HF-only baseline). Mean $\pm$ std of the per-seed paired ratio across 25 seeds. Best per row in \textbf{bold}. Mean PDF counterpart: Table~\ref{tab:results_pdf}; combined-mean main-body view: Table~\ref{tab:results_combined}.}
\label{tab:results_rmse}
\label{tab:results_rmse_std}
\resizebox{\textwidth}{!}{%
\begin{tabular}{lccccccccc}
\toprule
Test Function & MAST & AR1 & Recur. & Reif. & BoTorch & NARGP & MF-DGP & MISO & MFBNN \\
\midrule
Branin (2D)      & $1.19{\scriptstyle\pm0.40}$ & $2.06{\scriptstyle\pm1.07}$ & $6.03{\scriptstyle\pm10.61}$ & $2.24{\scriptstyle\pm0.76}$ & $\mathbf{0.95{\scriptstyle\pm0.41}}$ & $1.67{\scriptstyle\pm0.80}$ & $1.86{\scriptstyle\pm1.65}$ & $2.23{\scriptstyle\pm0.76}$ & $1.18{\scriptstyle\pm0.35}$ \\
Hartmann3 (3D)   & $0.69{\scriptstyle\pm0.19}$ & $\mathbf{0.29{\scriptstyle\pm0.09}}$ & $1.29{\scriptstyle\pm0.37}$ & $0.41{\scriptstyle\pm0.14}$ & $0.29{\scriptstyle\pm0.08}$ & $0.48{\scriptstyle\pm0.25}$ & $0.92{\scriptstyle\pm0.58}$ & $0.42{\scriptstyle\pm0.14}$ & $1.07{\scriptstyle\pm0.28}$ \\
Ackley (4D)      & $\mathbf{0.94{\scriptstyle\pm0.26}}$ & $1.63{\scriptstyle\pm0.47}$ & $1.19{\scriptstyle\pm0.39}$ & $1.19{\scriptstyle\pm0.42}$ & $1.02{\scriptstyle\pm0.34}$ & $1.59{\scriptstyle\pm0.52}$ & $1.57{\scriptstyle\pm0.44}$ & $1.11{\scriptstyle\pm0.38}$ & $1.62{\scriptstyle\pm0.69}$ \\
Park1 (4D)       & $\mathbf{0.97{\scriptstyle\pm0.48}}$ & $1.15{\scriptstyle\pm1.15}$ & $4.99{\scriptstyle\pm9.24}$ & $2.46{\scriptstyle\pm0.70}$ & $1.06{\scriptstyle\pm0.30}$ & $1.24{\scriptstyle\pm0.94}$ & $3.00{\scriptstyle\pm1.65}$ & $2.31{\scriptstyle\pm0.67}$ & $1.25{\scriptstyle\pm0.38}$ \\
Park2 (4D)       & $0.64{\scriptstyle\pm0.18}$ & $\mathbf{0.35{\scriptstyle\pm0.11}}$ & $1.46{\scriptstyle\pm0.46}$ & $2.82{\scriptstyle\pm0.44}$ & $0.51{\scriptstyle\pm0.11}$ & $1.31{\scriptstyle\pm0.56}$ & $1.35{\scriptstyle\pm0.57}$ & $2.75{\scriptstyle\pm0.43}$ & $0.82{\scriptstyle\pm0.25}$ \\
Hartmann6 (6D)   & $0.88{\scriptstyle\pm0.19}$ & $0.67{\scriptstyle\pm0.11}$ & $1.00{\scriptstyle\pm0.10}$ & $0.84{\scriptstyle\pm0.14}$ & $\mathbf{0.64{\scriptstyle\pm0.09}}$ & $0.93{\scriptstyle\pm0.21}$ & $1.09{\scriptstyle\pm0.15}$ & $0.81{\scriptstyle\pm0.13}$ & $1.02{\scriptstyle\pm0.18}$ \\
Levy (7D)        & $\mathbf{0.78{\scriptstyle\pm0.08}}$ & $1.03{\scriptstyle\pm0.11}$ & $1.31{\scriptstyle\pm1.42}$ & $0.87{\scriptstyle\pm0.09}$ & $0.81{\scriptstyle\pm0.05}$ & $1.10{\scriptstyle\pm0.14}$ & $1.92{\scriptstyle\pm0.45}$ & $0.85{\scriptstyle\pm0.08}$ & $1.15{\scriptstyle\pm0.12}$ \\
Borehole (8D)    & $\mathbf{0.88{\scriptstyle\pm0.24}}$ & $3.47{\scriptstyle\pm6.95}$ & $5.83{\scriptstyle\pm9.09}$ & $5.69{\scriptstyle\pm1.27}$ & $1.48{\scriptstyle\pm0.39}$ & $1.80{\scriptstyle\pm0.65}$ & $2.99{\scriptstyle\pm1.08}$ & $5.58{\scriptstyle\pm1.25}$ & $0.98{\scriptstyle\pm0.23}$ \\
Rastrigin (15D)  & $0.95{\scriptstyle\pm0.06}$ & $1.06{\scriptstyle\pm0.09}$ & $0.91{\scriptstyle\pm0.08}$ & $0.96{\scriptstyle\pm0.08}$ & $\mathbf{0.83{\scriptstyle\pm0.04}}$ & $1.29{\scriptstyle\pm0.26}$ & $1.83{\scriptstyle\pm0.18}$ & $0.91{\scriptstyle\pm0.06}$ & $1.67{\scriptstyle\pm0.35}$ \\
Rosenbrock (20D) & $\mathbf{0.37{\scriptstyle\pm0.04}}$ & $1.67{\scriptstyle\pm0.05}$ & $0.82{\scriptstyle\pm0.22}$ & $0.80{\scriptstyle\pm0.06}$ & $0.79{\scriptstyle\pm0.06}$ & $4.76{\scriptstyle\pm0.30}$ & $4.76{\scriptstyle\pm0.30}$ & $0.81{\scriptstyle\pm0.07}$ & $1.51{\scriptstyle\pm0.14}$ \\
\midrule
Concrete (8D)    & $\mathbf{0.89{\scriptstyle\pm0.15}}$ & $1.03{\scriptstyle\pm0.13}$ & $1.27{\scriptstyle\pm0.76}$ & $1.77{\scriptstyle\pm0.25}$ & $0.93{\scriptstyle\pm0.10}$ & $0.96{\scriptstyle\pm0.16}$ & $1.12{\scriptstyle\pm0.26}$ & $1.77{\scriptstyle\pm0.25}$ & $0.94{\scriptstyle\pm0.19}$ \\
DrivAerNet (23D) & $1.17{\scriptstyle\pm0.12}$ & $1.24{\scriptstyle\pm0.19}$ & $2.59{\scriptstyle\pm2.90}$ & $\mathbf{0.97{\scriptstyle\pm0.01}}$ & $1.37{\scriptstyle\pm0.12}$ & $0.99{\scriptstyle\pm0.11}$ & $0.99{\scriptstyle\pm0.06}$ & $0.97{\scriptstyle\pm0.01}$ & $7.29{\scriptstyle\pm1.14}$ \\
\bottomrule
\end{tabular}%
}
\end{table}

\begin{table}[!t]
\centering
\caption{Three-fidelity benchmarks: \textbf{normalised RMSE} ($\downarrow$, lower is better; $<1$ beats HF-only baseline). Mean $\pm$ std of the per-seed paired ratio across 25 seeds. Best per row in \textbf{bold}. Mean PDF counterpart: Table~\ref{tab:three_fid_pdf}; combined-mean main-body view: Table~\ref{tab:three_fid_combined}.}
\label{tab:three_fid_rmse}
\label{tab:three_fid_rmse_std}
\resizebox{\textwidth}{!}{%
\begin{tabular}{lccccccccc}
\toprule
Test Function & MAST & AR1 & Recur. & Reif. & BoTorch & NARGP & MF-DGP & MISO & MFBNN \\
\midrule
Branin (2D)      & $1.06{\scriptstyle\pm0.29}$ & $2.04{\scriptstyle\pm1.32}$ & $1.96{\scriptstyle\pm0.63}$ & $1.83{\scriptstyle\pm0.59}$ & $\mathbf{0.64{\scriptstyle\pm0.23}}$ & $1.91{\scriptstyle\pm0.64}$ & $1.25{\scriptstyle\pm0.61}$ & $1.78{\scriptstyle\pm0.57}$ & $1.37{\scriptstyle\pm0.47}$ \\
Hartmann3 (3D)   & $0.76{\scriptstyle\pm0.18}$ & $0.55{\scriptstyle\pm0.22}$ & $1.06{\scriptstyle\pm0.31}$ & $9.01{\scriptstyle\pm35.37}$ & $\mathbf{0.30{\scriptstyle\pm0.08}}$ & $0.97{\scriptstyle\pm0.42}$ & $0.97{\scriptstyle\pm0.35}$ & $0.58{\scriptstyle\pm0.16}$ & $1.42{\scriptstyle\pm0.39}$ \\
Ackley (4D)      & $\mathbf{0.84{\scriptstyle\pm0.23}}$ & $1.66{\scriptstyle\pm0.37}$ & $1.11{\scriptstyle\pm0.39}$ & $2.08{\scriptstyle\pm1.68}$ & $0.96{\scriptstyle\pm0.29}$ & $1.35{\scriptstyle\pm0.33}$ & $1.60{\scriptstyle\pm0.52}$ & $0.98{\scriptstyle\pm0.30}$ & $1.66{\scriptstyle\pm0.60}$ \\
Park1 (4D)       & $\mathbf{1.15{\scriptstyle\pm0.40}}$ & $1.17{\scriptstyle\pm0.49}$ & $2.10{\scriptstyle\pm1.27}$ & $3.68{\scriptstyle\pm4.96}$ & $1.23{\scriptstyle\pm0.42}$ & $1.57{\scriptstyle\pm1.36}$ & $2.20{\scriptstyle\pm0.92}$ & $1.84{\scriptstyle\pm0.54}$ & $2.80{\scriptstyle\pm1.18}$ \\
Park2 (4D)       & $0.77{\scriptstyle\pm0.23}$ & $0.64{\scriptstyle\pm0.24}$ & $0.93{\scriptstyle\pm0.59}$ & $2.43{\scriptstyle\pm0.40}$ & $\mathbf{0.62{\scriptstyle\pm0.14}}$ & $1.12{\scriptstyle\pm0.52}$ & $1.34{\scriptstyle\pm0.49}$ & $2.44{\scriptstyle\pm0.37}$ & $1.43{\scriptstyle\pm0.42}$ \\
Hartmann6 (6D)   & $0.91{\scriptstyle\pm0.19}$ & $0.93{\scriptstyle\pm0.27}$ & $1.00{\scriptstyle\pm0.18}$ & $1.82{\scriptstyle\pm1.99}$ & $\mathbf{0.65{\scriptstyle\pm0.10}}$ & $1.19{\scriptstyle\pm0.28}$ & $1.22{\scriptstyle\pm0.28}$ & $0.87{\scriptstyle\pm0.16}$ & $1.17{\scriptstyle\pm0.27}$ \\
Levy (7D)        & $\mathbf{0.75{\scriptstyle\pm0.09}}$ & $1.38{\scriptstyle\pm0.84}$ & $1.25{\scriptstyle\pm0.72}$ & $1.33{\scriptstyle\pm2.11}$ & $0.81{\scriptstyle\pm0.07}$ & $1.13{\scriptstyle\pm0.15}$ & $2.27{\scriptstyle\pm0.64}$ & $0.86{\scriptstyle\pm0.08}$ & $1.32{\scriptstyle\pm0.18}$ \\
Borehole (8D)    & $\mathbf{1.27{\scriptstyle\pm0.47}}$ & $3.48{\scriptstyle\pm3.70}$ & $4.79{\scriptstyle\pm6.42}$ & $4.38{\scriptstyle\pm1.09}$ & $1.60{\scriptstyle\pm0.43}$ & $1.40{\scriptstyle\pm1.00}$ & $2.71{\scriptstyle\pm0.89}$ & $4.20{\scriptstyle\pm0.96}$ & $2.26{\scriptstyle\pm0.64}$ \\
Rastrigin (15D)  & $0.93{\scriptstyle\pm0.07}$ & $1.02{\scriptstyle\pm0.09}$ & $0.96{\scriptstyle\pm0.09}$ & $0.94{\scriptstyle\pm0.09}$ & $\mathbf{0.82{\scriptstyle\pm0.04}}$ & $1.50{\scriptstyle\pm0.52}$ & $2.32{\scriptstyle\pm0.42}$ & $0.88{\scriptstyle\pm0.05}$ & $1.63{\scriptstyle\pm0.27}$ \\
Rosenbrock (20D) & $\mathbf{0.36{\scriptstyle\pm0.03}}$ & $1.53{\scriptstyle\pm0.12}$ & $0.90{\scriptstyle\pm0.17}$ & $0.91{\scriptstyle\pm0.23}$ & $0.82{\scriptstyle\pm0.06}$ & $4.75{\scriptstyle\pm0.31}$ & $4.75{\scriptstyle\pm0.31}$ & $0.85{\scriptstyle\pm0.07}$ & $1.73{\scriptstyle\pm0.16}$ \\
\midrule
HOIP (3D)        & $\mathbf{1.06{\scriptstyle\pm0.16}}$ & $2.37{\scriptstyle\pm2.76}$ & $1.10{\scriptstyle\pm0.32}$ & $2.06{\scriptstyle\pm0.43}$ & $1.43{\scriptstyle\pm0.39}$ & $1.18{\scriptstyle\pm0.28}$ & $1.12{\scriptstyle\pm0.20}$ & $2.06{\scriptstyle\pm0.43}$ & $1.75{\scriptstyle\pm0.49}$ \\
\bottomrule
\end{tabular}%
}
\end{table}

\begin{table}[!t]
\centering
\caption{Two-fidelity benchmarks: \textbf{normalised mean PDF} ($\uparrow$, higher is better; $>1$ beats HF-only baseline). Mean $\pm$ std of the per-seed paired ratio across 25 seeds. Best per row in \textbf{bold}. RMSE counterpart: Table~\ref{tab:results_rmse}; combined-mean main-body view: Table~\ref{tab:results_combined}.}
\label{tab:results_pdf}
\label{tab:results_pdf_std}
\resizebox{\textwidth}{!}{%
\begin{tabular}{lccccccccc}
\toprule
Test Function & MAST & AR1 & Recur. & Reif. & BoTorch & NARGP & MF-DGP & MISO & MFBNN \\
\midrule
Branin (2D)      & $1.28{\scriptstyle\pm0.43}$ & $0.63{\scriptstyle\pm0.75}$ & $0.62{\scriptstyle\pm0.47}$ & $1.37{\scriptstyle\pm0.44}$ & $\mathbf{1.43{\scriptstyle\pm0.46}}$ & $0.89{\scriptstyle\pm0.68}$ & $0.91{\scriptstyle\pm0.46}$ & $1.35{\scriptstyle\pm0.44}$ & $1.05{\scriptstyle\pm0.40}$ \\
Hartmann3 (3D)   & $1.83{\scriptstyle\pm0.50}$ & $\mathbf{3.97{\scriptstyle\pm0.74}}$ & $0.72{\scriptstyle\pm0.18}$ & $2.52{\scriptstyle\pm0.63}$ & $2.48{\scriptstyle\pm0.51}$ & $2.45{\scriptstyle\pm0.77}$ & $0.94{\scriptstyle\pm0.28}$ & $2.35{\scriptstyle\pm0.56}$ & $1.08{\scriptstyle\pm0.34}$ \\
Ackley (4D)      & $\mathbf{1.23{\scriptstyle\pm0.32}}$ & $0.75{\scriptstyle\pm0.20}$ & $0.92{\scriptstyle\pm0.23}$ & $0.96{\scriptstyle\pm0.23}$ & $0.83{\scriptstyle\pm0.21}$ & $0.69{\scriptstyle\pm0.16}$ & $0.65{\scriptstyle\pm0.13}$ & $1.03{\scriptstyle\pm0.24}$ & $0.72{\scriptstyle\pm0.24}$ \\
Park1 (4D)       & $\mathbf{1.46{\scriptstyle\pm0.49}}$ & $1.38{\scriptstyle\pm0.64}$ & $0.48{\scriptstyle\pm0.32}$ & $0.37{\scriptstyle\pm0.06}$ & $0.72{\scriptstyle\pm0.08}$ & $1.35{\scriptstyle\pm0.67}$ & $0.26{\scriptstyle\pm0.07}$ & $0.42{\scriptstyle\pm0.06}$ & $0.87{\scriptstyle\pm0.15}$ \\
Park2 (4D)       & $1.59{\scriptstyle\pm0.28}$ & $\mathbf{3.01{\scriptstyle\pm0.69}}$ & $0.67{\scriptstyle\pm0.23}$ & $0.14{\scriptstyle\pm0.06}$ & $1.28{\scriptstyle\pm0.15}$ & $0.92{\scriptstyle\pm0.21}$ & $0.65{\scriptstyle\pm0.19}$ & $0.11{\scriptstyle\pm0.06}$ & $0.74{\scriptstyle\pm0.09}$ \\
Hartmann6 (6D)   & $\mathbf{1.80{\scriptstyle\pm0.52}}$ & $1.50{\scriptstyle\pm0.46}$ & $1.06{\scriptstyle\pm0.41}$ & $1.42{\scriptstyle\pm0.40}$ & $1.17{\scriptstyle\pm0.29}$ & $1.30{\scriptstyle\pm0.59}$ & $0.89{\scriptstyle\pm0.36}$ & $1.45{\scriptstyle\pm0.38}$ & $1.14{\scriptstyle\pm0.36}$ \\
Levy (7D)        & $\mathbf{1.43{\scriptstyle\pm0.16}}$ & $0.94{\scriptstyle\pm0.28}$ & $1.03{\scriptstyle\pm0.39}$ & $1.25{\scriptstyle\pm0.15}$ & $1.10{\scriptstyle\pm0.07}$ & $0.87{\scriptstyle\pm0.12}$ & $0.65{\scriptstyle\pm0.13}$ & $1.31{\scriptstyle\pm0.13}$ & $1.06{\scriptstyle\pm0.13}$ \\
Borehole (8D)    & $\mathbf{1.58{\scriptstyle\pm0.20}}$ & $1.00{\scriptstyle\pm0.67}$ & $0.78{\scriptstyle\pm0.53}$ & $0.04{\scriptstyle\pm0.01}$ & $0.65{\scriptstyle\pm0.07}$ & $0.81{\scriptstyle\pm0.34}$ & $0.39{\scriptstyle\pm0.10}$ & $0.05{\scriptstyle\pm0.01}$ & $1.22{\scriptstyle\pm0.27}$ \\
Rastrigin (15D)  & $1.18{\scriptstyle\pm0.07}$ & $1.05{\scriptstyle\pm0.34}$ & $1.06{\scriptstyle\pm0.06}$ & $1.09{\scriptstyle\pm0.06}$ & $1.06{\scriptstyle\pm0.05}$ & $0.85{\scriptstyle\pm0.11}$ & $0.63{\scriptstyle\pm0.06}$ & $\mathbf{1.20{\scriptstyle\pm0.07}}$ & $0.64{\scriptstyle\pm0.14}$ \\
Rosenbrock (20D) & $\mathbf{3.23{\scriptstyle\pm0.23}}$ & $0.39{\scriptstyle\pm0.55}$ & $1.27{\scriptstyle\pm0.17}$ & $1.38{\scriptstyle\pm0.07}$ & $1.14{\scriptstyle\pm0.06}$ & $0.17{\scriptstyle\pm0.04}$ & $0.00{\scriptstyle\pm0.00}$ & $1.40{\scriptstyle\pm0.07}$ & $0.84{\scriptstyle\pm0.07}$ \\
\midrule
Concrete (8D)    & $\mathbf{1.27{\scriptstyle\pm0.22}}$ & $0.94{\scriptstyle\pm0.19}$ & $0.84{\scriptstyle\pm0.33}$ & $0.68{\scriptstyle\pm0.13}$ & $1.11{\scriptstyle\pm0.13}$ & $1.18{\scriptstyle\pm0.29}$ & $0.88{\scriptstyle\pm0.12}$ & $0.68{\scriptstyle\pm0.13}$ & $1.13{\scriptstyle\pm0.21}$ \\
DrivAerNet (23D) & $0.84{\scriptstyle\pm0.10}$ & $\mathbf{0.99{\scriptstyle\pm0.11}}$ & $0.30{\scriptstyle\pm0.44}$ & $0.60{\scriptstyle\pm0.03}$ & $0.42{\scriptstyle\pm0.03}$ & $0.97{\scriptstyle\pm0.05}$ & $0.87{\scriptstyle\pm0.05}$ & $0.60{\scriptstyle\pm0.03}$ & $0.09{\scriptstyle\pm0.02}$ \\
\bottomrule
\end{tabular}%
}
\end{table}

\begin{table}[!t]
\centering
\caption{Three-fidelity benchmarks: \textbf{normalised mean PDF} ($\uparrow$, higher is better; $>1$ beats HF-only baseline). Mean $\pm$ std of the per-seed paired ratio across 25 seeds. Best per row in \textbf{bold}. RMSE counterpart: Table~\ref{tab:three_fid_rmse}; combined-mean main-body view: Table~\ref{tab:three_fid_combined}.}
\label{tab:three_fid_pdf}
\label{tab:three_fid_pdf_std}
\resizebox{\textwidth}{!}{%
\begin{tabular}{lccccccccc}
\toprule
Test Function & MAST & AR1 & Recur. & Reif. & BoTorch & NARGP & MF-DGP & MISO & MFBNN \\
\midrule
Branin (2D)      & $1.24{\scriptstyle\pm0.47}$ & $0.68{\scriptstyle\pm0.51}$ & $0.63{\scriptstyle\pm0.37}$ & $1.12{\scriptstyle\pm0.27}$ & $\mathbf{1.55{\scriptstyle\pm0.50}}$ & $0.43{\scriptstyle\pm0.28}$ & $0.95{\scriptstyle\pm0.30}$ & $1.20{\scriptstyle\pm0.30}$ & $0.91{\scriptstyle\pm0.40}$ \\
Hartmann3 (3D)   & $1.60{\scriptstyle\pm0.48}$ & $2.24{\scriptstyle\pm0.74}$ & $0.85{\scriptstyle\pm0.26}$ & $1.61{\scriptstyle\pm0.37}$ & $\mathbf{2.28{\scriptstyle\pm0.47}}$ & $1.27{\scriptstyle\pm0.42}$ & $0.63{\scriptstyle\pm0.16}$ & $1.85{\scriptstyle\pm0.44}$ & $0.80{\scriptstyle\pm0.25}$ \\
Ackley (4D)      & $\mathbf{1.37{\scriptstyle\pm0.27}}$ & $0.77{\scriptstyle\pm0.18}$ & $0.91{\scriptstyle\pm0.22}$ & $1.05{\scriptstyle\pm0.24}$ & $0.82{\scriptstyle\pm0.18}$ & $0.78{\scriptstyle\pm0.16}$ & $0.65{\scriptstyle\pm0.16}$ & $1.19{\scriptstyle\pm0.23}$ & $0.68{\scriptstyle\pm0.20}$ \\
Park1 (4D)       & $\mathbf{1.21{\scriptstyle\pm0.25}}$ & $1.09{\scriptstyle\pm0.37}$ & $0.59{\scriptstyle\pm0.31}$ & $0.54{\scriptstyle\pm0.39}$ & $0.61{\scriptstyle\pm0.07}$ & $0.92{\scriptstyle\pm0.31}$ & $0.23{\scriptstyle\pm0.05}$ & $0.53{\scriptstyle\pm0.09}$ & $0.43{\scriptstyle\pm0.16}$ \\
Park2 (4D)       & $1.37{\scriptstyle\pm0.28}$ & $\mathbf{1.82{\scriptstyle\pm0.49}}$ & $1.07{\scriptstyle\pm0.44}$ & $0.21{\scriptstyle\pm0.07}$ & $1.09{\scriptstyle\pm0.10}$ & $1.17{\scriptstyle\pm0.37}$ & $0.43{\scriptstyle\pm0.07}$ & $0.13{\scriptstyle\pm0.05}$ & $0.56{\scriptstyle\pm0.11}$ \\
Hartmann6 (6D)   & $1.61{\scriptstyle\pm0.53}$ & $1.30{\scriptstyle\pm0.53}$ & $1.11{\scriptstyle\pm0.48}$ & $\mathbf{2.30{\scriptstyle\pm4.97}}$ & $1.21{\scriptstyle\pm0.35}$ & $1.07{\scriptstyle\pm0.54}$ & $0.83{\scriptstyle\pm0.34}$ & $1.53{\scriptstyle\pm0.43}$ & $1.03{\scriptstyle\pm0.33}$ \\
Levy (7D)        & $\mathbf{1.49{\scriptstyle\pm0.16}}$ & $0.86{\scriptstyle\pm0.33}$ & $0.91{\scriptstyle\pm0.26}$ & $1.18{\scriptstyle\pm0.10}$ & $1.10{\scriptstyle\pm0.07}$ & $0.83{\scriptstyle\pm0.10}$ & $0.65{\scriptstyle\pm0.14}$ & $1.32{\scriptstyle\pm0.09}$ & $0.93{\scriptstyle\pm0.11}$ \\
Borehole (8D)    & $\mathbf{1.31{\scriptstyle\pm0.49}}$ & $0.71{\scriptstyle\pm0.58}$ & $0.57{\scriptstyle\pm0.32}$ & $0.10{\scriptstyle\pm0.02}$ & $0.58{\scriptstyle\pm0.06}$ & $0.85{\scriptstyle\pm0.18}$ & $0.36{\scriptstyle\pm0.06}$ & $0.10{\scriptstyle\pm0.02}$ & $0.51{\scriptstyle\pm0.08}$ \\
Rastrigin (15D)  & $1.23{\scriptstyle\pm0.08}$ & $1.08{\scriptstyle\pm0.34}$ & $1.04{\scriptstyle\pm0.09}$ & $1.15{\scriptstyle\pm0.07}$ & $1.09{\scriptstyle\pm0.05}$ & $0.75{\scriptstyle\pm0.23}$ & $0.55{\scriptstyle\pm0.09}$ & $\mathbf{1.30{\scriptstyle\pm0.07}}$ & $0.66{\scriptstyle\pm0.13}$ \\
Rosenbrock (20D) & $\mathbf{3.33{\scriptstyle\pm0.29}}$ & $0.32{\scriptstyle\pm0.55}$ & $1.11{\scriptstyle\pm0.09}$ & $1.28{\scriptstyle\pm0.08}$ & $1.13{\scriptstyle\pm0.06}$ & $0.18{\scriptstyle\pm0.01}$ & $0.16{\scriptstyle\pm0.01}$ & $1.39{\scriptstyle\pm0.08}$ & $0.74{\scriptstyle\pm0.05}$ \\
\midrule
HOIP (3D)        & $\mathbf{1.12{\scriptstyle\pm0.26}}$ & $0.80{\scriptstyle\pm0.85}$ & $0.89{\scriptstyle\pm0.30}$ & $0.64{\scriptstyle\pm0.14}$ & $0.66{\scriptstyle\pm0.23}$ & $0.84{\scriptstyle\pm0.21}$ & $0.79{\scriptstyle\pm0.18}$ & $0.64{\scriptstyle\pm0.14}$ & $0.64{\scriptstyle\pm0.23}$ \\
\bottomrule
\end{tabular}%
}
\end{table}
\FloatBarrier

\subsection{Extended Results: Normalised $R^2$}
\label{app:r2_extra}

Tables~\ref{tab:results_r2} and~\ref{tab:three_fid_r2} report normalised $R^2$ (Eq.~\eqref{eq:r2_def}) for the two-fidelity and three-fidelity benchmark suites respectively, mirroring the row order of main-body Tables~\ref{tab:results_combined} and~\ref{tab:three_fid_combined} and of the appendix RMSE / Mean PDF tables (Tables~\ref{tab:results_rmse}, \ref{tab:three_fid_rmse}, \ref{tab:results_pdf}, \ref{tab:three_fid_pdf}). Values above $1$ beat the HF-only reference on goodness-of-fit; values below zero indicate worse-than-mean-predictor fit (the surrogate mean is farther from the ground truth than the sample mean of the response). Large magnitudes arise when the HF-only baseline's $R^2$ is close to zero, which makes the normaliser small and amplifies all ratios. The sign of the ratio should not be over-interpreted in this regime: when both numerator and denominator are negative the ratio is positive but the underlying fit is still worse than a mean predictor; the Levy~7D row of Table~\ref{tab:results_r2} (with mean $-21.89 \pm 128.03$) is the worked illustration.

\begin{table}[h]
\centering
\caption{Two-fidelity benchmarks: \textbf{normalised $R^2$} ($\uparrow$, higher is better; $>1$ beats HF-only baseline). Mean $\pm$ std of the per-seed paired ratio across 25 seeds. Best per row in \textbf{bold}. Concrete and DrivAerNet are two-fidelity real-world benchmarks (HF/LF, 70/30 split, independent-sampling configuration). RMSE / Mean PDF counterparts: Tables~\ref{tab:results_rmse}, \ref{tab:results_pdf}; combined-mean main-body view: Table~\ref{tab:results_combined}.}
\label{tab:results_r2}
\label{tab:r2_synthetic}
\label{tab:results_r2_std}
\resizebox{\textwidth}{!}{%
\begin{tabular}{l c c c c c c c c c}
\toprule
Test function & MAST & AR1 & Recur. & Reif. & BoTorch & NARGP & MF-DGP & MISO & MFBNN \\
\midrule
Branin (2D)       & $0.97{\scriptstyle\pm1.66}$ & $-0.68{\scriptstyle\pm2.72}$ & $-93.91{\scriptstyle\pm260.58}$ & $-0.87{\scriptstyle\pm0.88}$ & $\mathbf{1.41{\scriptstyle\pm2.61}}$ & $0.37{\scriptstyle\pm1.51}$ & $-0.67{\scriptstyle\pm5.71}$ & $-0.82{\scriptstyle\pm0.86}$ & $1.00{\scriptstyle\pm1.83}$ \\
Hartmann3 (3D)    & $1.27{\scriptstyle\pm0.24}$ & $\mathbf{1.43{\scriptstyle\pm0.25}}$ & $0.81{\scriptstyle\pm0.31}$ & $1.38{\scriptstyle\pm0.25}$ & $1.42{\scriptstyle\pm0.25}$ & $1.32{\scriptstyle\pm0.29}$ & $1.08{\scriptstyle\pm0.52}$ & $1.38{\scriptstyle\pm0.25}$ & $0.98{\scriptstyle\pm0.27}$ \\
Ackley (4D)       & $\mathbf{1.83{\scriptstyle\pm3.93}}$ & $-2.35{\scriptstyle\pm8.56}$ & $0.94{\scriptstyle\pm1.67}$ & $0.94{\scriptstyle\pm2.25}$ & $1.50{\scriptstyle\pm2.43}$ & $-1.64{\scriptstyle\pm4.46}$ & $-0.55{\scriptstyle\pm1.16}$ & $1.23{\scriptstyle\pm2.77}$ & $-0.86{\scriptstyle\pm3.12}$ \\
Park1 (4D)        & $\mathbf{1.00{\scriptstyle\pm0.01}}$ & $1.00{\scriptstyle\pm0.03}$ & $0.66{\scriptstyle\pm1.08}$ & $0.98{\scriptstyle\pm0.01}$ & $1.00{\scriptstyle\pm0.01}$ & $1.00{\scriptstyle\pm0.01}$ & $0.96{\scriptstyle\pm0.07}$ & $0.98{\scriptstyle\pm0.01}$ & $1.00{\scriptstyle\pm0.01}$ \\
Park2 (4D)        & $1.01{\scriptstyle\pm0.01}$ & $\mathbf{1.02{\scriptstyle\pm0.01}}$ & $0.98{\scriptstyle\pm0.02}$ & $0.86{\scriptstyle\pm0.03}$ & $1.02{\scriptstyle\pm0.01}$ & $0.98{\scriptstyle\pm0.04}$ & $0.98{\scriptstyle\pm0.04}$ & $0.87{\scriptstyle\pm0.03}$ & $1.01{\scriptstyle\pm0.01}$ \\
Hartmann6 (6D)    & $\mathbf{3.69{\scriptstyle\pm15.86}}$ & $-2.55{\scriptstyle\pm16.08}$ & $0.94{\scriptstyle\pm5.20}$ & $-0.47{\scriptstyle\pm12.37}$ & $-2.32{\scriptstyle\pm17.25}$ & $-0.56{\scriptstyle\pm8.02}$ & $2.63{\scriptstyle\pm15.49}$ & $-0.52{\scriptstyle\pm12.97}$ & $2.42{\scriptstyle\pm8.37}$ \\
Levy (7D)         & $-21.89{\scriptstyle\pm128.03}$ & $0.62{\scriptstyle\pm6.36}$ & $-53.96{\scriptstyle\pm230.76}$ & $-14.46{\scriptstyle\pm79.56}$ & $-15.43{\scriptstyle\pm93.75}$ & $14.01{\scriptstyle\pm84.49}$ & $\mathbf{2.01\!\times\!10^{2}\!\!\scriptstyle\pm 1.13\!\times\!10^{3}}$ & $-16.65{\scriptstyle\pm93.11}$ & $1.68{\scriptstyle\pm51.38}$ \\
Borehole (8D)     & $\mathbf{1.00{\scriptstyle\pm0.00}}$ & $0.75{\scriptstyle\pm0.96}$ & $0.61{\scriptstyle\pm0.96}$ & $0.87{\scriptstyle\pm0.01}$ & $1.00{\scriptstyle\pm0.00}$ & $0.99{\scriptstyle\pm0.01}$ & $0.96{\scriptstyle\pm0.03}$ & $0.87{\scriptstyle\pm0.01}$ & $1.00{\scriptstyle\pm0.00}$ \\
Rastrigin (15D)   & $2.96{\scriptstyle\pm7.18}$ & $-4.82{\scriptstyle\pm23.02}$ & $5.01{\scriptstyle\pm16.47}$ & $1.00{\scriptstyle\pm4.35}$ & $\mathbf{7.32{\scriptstyle\pm20.26}}$ & $-9.61{\scriptstyle\pm36.25}$ & $-45.77{\scriptstyle\pm138.21}$ & $3.23{\scriptstyle\pm6.54}$ & $-26.28{\scriptstyle\pm68.80}$ \\
Rosenbrock (20D)  & $\mathbf{1.48{\scriptstyle\pm0.10}}$ & $0.00{\scriptstyle\pm0.00}$ & $1.16{\scriptstyle\pm0.34}$ & $1.21{\scriptstyle\pm0.09}$ & $1.21{\scriptstyle\pm0.09}$ & $-10.72{\scriptstyle\pm0.73}$ & $-10.72{\scriptstyle\pm0.73}$ & $1.20{\scriptstyle\pm0.09}$ & $0.31{\scriptstyle\pm0.15}$ \\
\midrule
Concrete (8D)     & $0.89{\scriptstyle\pm1.51}$ & $1.01{\scriptstyle\pm0.30}$ & $0.07{\scriptstyle\pm2.26}$ & $-0.69{\scriptstyle\pm0.93}$ & $\mathbf{1.06{\scriptstyle\pm0.49}}$ & $0.96{\scriptstyle\pm0.64}$ & $0.82{\scriptstyle\pm0.52}$ & $-0.69{\scriptstyle\pm0.93}$ & $0.87{\scriptstyle\pm1.48}$ \\
DrivAerNet (23D)  & $-0.25{\scriptstyle\pm0.88}$ & $-0.16{\scriptstyle\pm3.18}$ & $\mathbf{1.03\!\times\!10^{2}\!\!\scriptstyle\pm 6.66\!\times\!10^{2}}$ & $1.06{\scriptstyle\pm0.47}$ & $-1.44{\scriptstyle\pm1.63}$ & $0.78{\scriptstyle\pm1.28}$ & $0.92{\scriptstyle\pm1.02}$ & $1.06{\scriptstyle\pm0.47}$ & $-1.12\!\times\!10^{2}{\scriptstyle\pm 2.50\!\times\!10^{2}}$ \\
\bottomrule
\end{tabular}%
}
\end{table}

The two-fidelity $R^2$ ranking broadly mirrors RMSE: MAST and BoTorch MF-GP share the top of the synthetic suite, with MAST taking five outright wins (Ackley, Park1, Hartmann6, Borehole, Rosenbrock) and BoTorch two (Branin, Rastrigin); AR1 captures the two functions whose inter-fidelity relation is globally linear (Hartmann3, Park2). On the real-world rows BoTorch leads on Concrete and recursive co-kriging takes DrivAerNet under independent sampling. The DrivAerNet outlier, where the leading method's normalised $R^2$ exceeds $10^{2}$ with comparably large standard deviation, is driven by a near-zero HF-only denominator rather than an exceptionally accurate fit. Negative entries (for example MF-DGP's $-45.77$ on Rastrigin) indicate the surrogate mean is farther from the ground truth than a constant predictor at the training mean --- a known failure mode of expressive surrogates under data-scarce budgets.

\begin{table}[h]
\centering
\caption{Three-fidelity benchmarks: \textbf{normalised $R^2$} ($\uparrow$, higher is better; $>1$ beats HF-only baseline). Mean $\pm$ std of the per-seed paired ratio across 25 seeds. Best per row in \textbf{bold}. HOIP is the three-fidelity real-world benchmark (HF/MF/LF, 50/30/20 split). Large magnitudes (e.g.\ Rastrigin row, HOIP row) reflect a near-zero HF-only denominator, not exceptionally good fits. RMSE / Mean PDF counterparts: Tables~\ref{tab:three_fid_rmse}, \ref{tab:three_fid_pdf}; combined-mean main-body view: Table~\ref{tab:three_fid_combined}.}
\label{tab:three_fid_r2}
\label{tab:three_fidelity_r2}
\label{tab:three_fid_r2_std}
\resizebox{\textwidth}{!}{%
\begin{tabular}{l c c c c c c c c c}
\toprule
Test function & MAST & AR1 & Recur. & Reif. & BoTorch & NARGP & MF-DGP & MISO & MFBNN \\
\midrule
Branin (2D)       & $1.20{\scriptstyle\pm2.22}$ & $-1.05{\scriptstyle\pm3.94}$ & $-0.39{\scriptstyle\pm2.59}$ & $-0.02{\scriptstyle\pm1.20}$ & $1.64{\scriptstyle\pm2.77}$ & $-0.94{\scriptstyle\pm2.83}$ & $\mathbf{2.91{\scriptstyle\pm9.84}}$ & $0.20{\scriptstyle\pm1.39}$ & $0.96{\scriptstyle\pm2.15}$ \\
Hartmann3 (3D)    & $1.41{\scriptstyle\pm0.72}$ & $1.55{\scriptstyle\pm0.83}$ & $1.11{\scriptstyle\pm0.79}$ & $-3.12\!\times\!10^{2}{\scriptstyle\pm 1.53\!\times\!10^{3}}$ & $\mathbf{1.66{\scriptstyle\pm0.82}}$ & $1.18{\scriptstyle\pm0.81}$ & $1.25{\scriptstyle\pm0.77}$ & $1.52{\scriptstyle\pm0.72}$ & $0.57{\scriptstyle\pm1.01}$ \\
Ackley (4D)       & $\mathbf{2.13{\scriptstyle\pm3.59}}$ & $-2.15{\scriptstyle\pm6.49}$ & $1.30{\scriptstyle\pm3.52}$ & $-2.58{\scriptstyle\pm8.61}$ & $1.65{\scriptstyle\pm2.32}$ & $0.56{\scriptstyle\pm2.11}$ & $-0.96{\scriptstyle\pm2.49}$ & $1.75{\scriptstyle\pm3.00}$ & $-2.58{\scriptstyle\pm12.83}$ \\
Park1 (4D)        & $\mathbf{1.00{\scriptstyle\pm0.01}}$ & $1.00{\scriptstyle\pm0.01}$ & $0.98{\scriptstyle\pm0.03}$ & $0.82{\scriptstyle\pm0.62}$ & $1.00{\scriptstyle\pm0.00}$ & $0.99{\scriptstyle\pm0.04}$ & $0.98{\scriptstyle\pm0.02}$ & $0.99{\scriptstyle\pm0.01}$ & $0.97{\scriptstyle\pm0.03}$ \\
Park2 (4D)        & $1.01{\scriptstyle\pm0.01}$ & $1.01{\scriptstyle\pm0.01}$ & $1.00{\scriptstyle\pm0.04}$ & $0.90{\scriptstyle\pm0.02}$ & $\mathbf{1.01{\scriptstyle\pm0.01}}$ & $0.99{\scriptstyle\pm0.02}$ & $0.98{\scriptstyle\pm0.03}$ & $0.90{\scriptstyle\pm0.02}$ & $0.98{\scriptstyle\pm0.02}$ \\
Hartmann6 (6D)    & $0.71{\scriptstyle\pm8.09}$ & $4.64{\scriptstyle\pm24.15}$ & $1.22{\scriptstyle\pm5.82}$ & $0.36{\scriptstyle\pm129.44}$ & $0.40{\scriptstyle\pm17.09}$ & $1.91{\scriptstyle\pm13.19}$ & $\mathbf{5.70{\scriptstyle\pm23.05}}$ & $0.14{\scriptstyle\pm10.10}$ & $1.58{\scriptstyle\pm15.21}$ \\
Levy (7D)         & $0.00{\scriptstyle\pm7.43}$ & $10.65{\scriptstyle\pm44.04}$ & $\mathbf{14.79{\scriptstyle\pm48.17}}$ & $-2.27\!\times\!10^{2}{\scriptstyle\pm 1.13\!\times\!10^{3}}$ & $-0.36{\scriptstyle\pm5.87}$ & $0.13{\scriptstyle\pm5.15}$ & $-11.86{\scriptstyle\pm85.80}$ & $-0.02{\scriptstyle\pm5.08}$ & $0.46{\scriptstyle\pm14.00}$ \\
Borehole (8D)     & $\mathbf{1.00{\scriptstyle\pm0.01}}$ & $0.88{\scriptstyle\pm0.33}$ & $0.76{\scriptstyle\pm0.53}$ & $0.92{\scriptstyle\pm0.02}$ & $0.99{\scriptstyle\pm0.00}$ & $0.99{\scriptstyle\pm0.03}$ & $0.97{\scriptstyle\pm0.02}$ & $0.93{\scriptstyle\pm0.01}$ & $0.98{\scriptstyle\pm0.01}$ \\
Rastrigin (15D)   & $-2.68{\scriptstyle\pm8.59}$ & $3.45{\scriptstyle\pm31.61}$ & $0.84{\scriptstyle\pm12.17}$ & $-3.71{\scriptstyle\pm21.92}$ & $-9.64{\scriptstyle\pm32.33}$ & $43.72{\scriptstyle\pm162.17}$ & $\mathbf{1.60\!\times\!10^{2}\!\!\scriptstyle\pm 3.98\!\times\!10^{2}}$ & $-6.98{\scriptstyle\pm23.50}$ & $37.99{\scriptstyle\pm218.20}$ \\
Rosenbrock (20D)  & $\mathbf{1.48{\scriptstyle\pm0.11}}$ & $0.20{\scriptstyle\pm0.14}$ & $1.09{\scriptstyle\pm0.22}$ & $1.07{\scriptstyle\pm0.36}$ & $1.19{\scriptstyle\pm0.09}$ & $-10.74{\scriptstyle\pm0.75}$ & $-10.74{\scriptstyle\pm0.75}$ & $1.16{\scriptstyle\pm0.09}$ & $-0.10{\scriptstyle\pm0.29}$ \\
\midrule
HOIP (3D)         & $30.10{\scriptstyle\pm141.34}$ & $70.28{\scriptstyle\pm987.24}$ & $13.28{\scriptstyle\pm59.07}$ & $\mathbf{1.67\!\times\!10^{2}\!\!\scriptstyle\pm 7.87\!\times\!10^{2}}$ & $1.35\!\times\!10^{2}{\scriptstyle\pm 6.51\!\times\!10^{2}}$ & $28.23{\scriptstyle\pm126.20}$ & $10.09{\scriptstyle\pm45.98}$ & $1.67\!\times\!10^{2}{\scriptstyle\pm 7.87\!\times\!10^{2}}$ & $1.09\!\times\!10^{2}{\scriptstyle\pm 5.13\!\times\!10^{2}}$ \\
\bottomrule
\end{tabular}%
}
\end{table}

In the three-fidelity setting MAST takes the most $R^2$ wins on the synthetic suite (Ackley, Park1, Borehole, Rosenbrock), while BoTorch MF-GP captures Hartmann3 and Park2 where the inter-fidelity correlation is most regular and recursive co-kriging takes Levy. The MF-DGP wins on Branin, Hartmann6, and Rastrigin all carry standard deviations larger than the mean and reflect a near-zero HF-only denominator that amplifies any ratio rather than an exceptionally accurate fit. On the real-world HOIP row, reification leads (with MISO tying), with the entire row dominated by near-zero HF-only denominators that produce $R^2$ ratios exceeding $10^{2}$.


\subsection{Collocated vs Independent Sampling on Real-World Benchmarks}
\label{app:collocated_comparison}

The headline real-world results in Table~\ref{tab:results_combined} use the \emph{independent-sampling} configuration, where the high-fidelity and low-fidelity training points are drawn separately at each fidelity rather than from a paired pool. This is the more demanding setting: methods that exploit point-to-point correspondence between fidelities (autoregressive co-kriging, recursive co-kriging, and NARGP) lose their structural advantage when LF samples no longer share locations with HF samples. We report the collocated counterpart here as a fairness check.

Table~\ref{tab:rw_collocated} presents the collocated-sampling results for HOIP and DrivAerNet on the same nine methods, the same budget, and the same 25 seeds; only the sampling regime differs. Concrete is omitted because its low fidelity is an analytical formula evaluated at each HF input by construction (no separate-pool counterpart exists). HOIP retains its three-tier HF/MF/LF setup; only the within-tier sampling regime changes.

\begin{table}[h]
\centering
\caption{HOIP and DrivAerNet under collocated sampling: normalised RMSE / normalised mean PDF, mean $\pm$ std across 25 seeds. All values are normalised relative to the cost-equivalent HF-only baseline. Best per row and metric in \textbf{bold}.}
\label{tab:rw_collocated}
\resizebox{\textwidth}{!}{%
\Large
\begin{tabular}{lccccccccccccccccccc}
\toprule
& \multicolumn{9}{c}{\textbf{Normalised RMSE} $\downarrow$} & & \multicolumn{9}{c}{\textbf{Normalised Mean PDF} $\uparrow$} \\
\cmidrule(lr){2-10} \cmidrule(lr){12-20}
Benchmark & MAST & AR1 & Recur. & Reif. & BoTorch & NARGP & MF-DGP & MISO & MFBNN & & MAST & AR1 & Recur. & Reif. & BoTorch & NARGP & MF-DGP & MISO & MFBNN \\
\midrule
HOIP (3D)        & $1.08{\scriptstyle\pm0.16}$ & $2.37{\scriptstyle\pm2.67}$ & $\mathbf{1.00{\scriptstyle\pm0.25}}$ & $2.12{\scriptstyle\pm0.46}$ & $1.39{\scriptstyle\pm0.36}$ & $1.12{\scriptstyle\pm0.24}$ & $1.20{\scriptstyle\pm0.34}$ & $2.12{\scriptstyle\pm0.46}$ & $1.74{\scriptstyle\pm0.49}$ & & $\mathbf{1.12{\scriptstyle\pm0.28}}$ & $0.86{\scriptstyle\pm1.79}$ & $1.03{\scriptstyle\pm0.27}$ & $0.63{\scriptstyle\pm0.17}$ & $0.68{\scriptstyle\pm0.21}$ & $0.90{\scriptstyle\pm0.24}$ & $0.81{\scriptstyle\pm0.22}$ & $0.63{\scriptstyle\pm0.17}$ & $0.66{\scriptstyle\pm0.24}$ \\
DrivAerNet (23D) & $1.01{\scriptstyle\pm0.10}$ & $1.27{\scriptstyle\pm0.16}$ & $2.73{\scriptstyle\pm3.20}$ & $\mathbf{0.98{\scriptstyle\pm0.01}}$ & $1.24{\scriptstyle\pm0.07}$ & $1.04{\scriptstyle\pm0.12}$ & $0.99{\scriptstyle\pm0.08}$ & $0.98{\scriptstyle\pm0.01}$ & $7.34{\scriptstyle\pm1.12}$ & & $\mathbf{1.02{\scriptstyle\pm0.09}}$ & $0.97{\scriptstyle\pm0.11}$ & $0.15{\scriptstyle\pm0.29}$ & $0.60{\scriptstyle\pm0.03}$ & $0.45{\scriptstyle\pm0.05}$ & $0.96{\scriptstyle\pm0.08}$ & $0.73{\scriptstyle\pm0.17}$ & $0.60{\scriptstyle\pm0.03}$ & $0.09{\scriptstyle\pm0.02}$ \\
\bottomrule
\end{tabular}%
}
\end{table}

\paragraph{What the comparison reveals.}
Two patterns are worth flagging. First, recursive co-kriging recovers the HOIP RMSE win under collocation (1.00 collocated vs 1.10 independent), confirming the textbook expectation that the recursive nested-design construction profits when LF samples lie at HF locations \citep{LeGratiet2014RecursiveFidelity}: AR1's scalar correlation parameter is estimated from paired residuals at exactly such points, the recursive construction explicitly assumes nested designs, and NARGP's nonlinear lower-fidelity argument is most informative when evaluated at HF inputs. NARGP and AR1 also tighten under collocation (NARGP HOIP RMSE: 1.12 collocated vs 1.18 independent; AR1 HOIP: 2.37 vs 2.37 --- essentially unchanged), and DrivAerNet collapses the field considerably (Reification edges out MISO at 0.98 RMSE, MAST drops to 1.01 from 1.17 under independent sampling). MFBNN's 7.34 on DrivAerNet is essentially unchanged from its 7.29 under independent sampling, indicating its failure is structural rather than sampling-driven. Second, MAST is largely insensitive to the sampling regime (HOIP RMSE: 1.08 collocated vs 1.06 independent; DrivAerNet RMSE: 1.01 vs 1.17) and keeps the calibration win on both datasets in both regimes, indicating that the trust-weighted fusion is robust to the sampling choice on the calibration metric.

This validates the design choice of reporting the independent-sampling regime as the headline. The collocated regime is the easier setting for methods that assume nested or paired data; presenting it instead would understate the practical advantage of MAST's sampling-agnostic geometric weighting on the kind of real-world dataset where HF and LF observations rarely live at the same points.

\section{MAST is Surrogate-Agnostic: MAST-ICL with TabPFN}
\label{app:mast_icl}

\subsection{Motivation}
\label{app:mast_icl_motivation}

The contribution of MAST is the trust-weighted augmentation scheme, not the choice of base surrogate. Stage~1 fits an independent regressor on each fidelity. Stage~2 fits a discrepancy regressor on augmented inputs $\mathbf{z}_j^{(m)} = [\mathbf{x}_j^{(M)},\, \mu_m(\mathbf{x}_j^{(M)}),\, \sigma_m^2(\mathbf{x}_j^{(M)})]$ at HF locations and blends the corrected LF values with HF predictions through the spatial weight $W_m^{(i)}$. Stage~3 fits a fusion regressor on the combined dataset. The only requirement on the base learner is that it exposes a posterior mean and variance at query points. Any regressor with calibrated uncertainty can serve.

We call the instance of this template in which all three Gaussian processes are replaced by TabPFN v2 regressors \citep{Hollmann2025TabPFN} MAST-ICL (MAST with In-Context Learning). TabPFN is a pre-trained transformer that performs zero-shot Bayesian regression on small tabular datasets and returns a calibrated predictive distribution at inference time with no gradient-based training. The variant exchanges GP kernel hyperparameter fitting for in-context inference on a pre-trained model, and in doing so stress-tests whether the MAST pipeline transfers to a substantially different probabilistic regressor. The method is described here; preliminary results are reported below.

\subsection{Method}
\label{app:mast_icl_method}

Let $\mathrm{TabPFN}(\mathbf{X}, \mathbf{y})$ denote a TabPFN regressor conditioned on training set $(\mathbf{X}, \mathbf{y})$, with $\mathrm{TabPFN}(\cdot)(\mathbf{x}_*) \to (\hat\mu(\mathbf{x}_*), \hat\sigma^2(\mathbf{x}_*))$ returning the predictive mean and variance at a query. MAST-ICL constructs four such regressors.

\paragraph{Stage 1.}
Independent TabPFN regressors are fitted on each fidelity,
\begin{equation}
    \mathrm{TabPFN}_{\mathrm{LF}} = \mathrm{TabPFN}\bigl(\mathcal{D}_m\bigr), \qquad
    \mathrm{TabPFN}_{\mathrm{HF}} = \mathrm{TabPFN}\bigl(\mathcal{D}_M\bigr),
\end{equation}
yielding $\bigl(\mu_m(\mathbf{x}), \sigma_m^2(\mathbf{x})\bigr)$ and $\bigl(\mu_M(\mathbf{x}), \sigma_M^2(\mathbf{x})\bigr)$ at any query point.

\paragraph{Stage 2: discrepancy and augmentation.}
Residuals and augmented inputs are formed identically to MAST-GP (Eqs.~\eqref{eq:disc}, \eqref{eq:lf_augmented_input}): at each HF location $\mathbf{x}_j^{(M)}$ we compute $\delta_j = y_j^{(M)} - \mu_m(\mathbf{x}_j^{(M)})$ and $\mathbf{z}_j^{(m)} = [\mathbf{x}_j^{(M)},\, \mu_m^{(j)},\, \sigma_m^{2(j)}]$. A third TabPFN regressor is fitted on the augmented discrepancy data,
\begin{equation}
    \mathrm{TabPFN}_{\delta} = \mathrm{TabPFN}\bigl(\{(\mathbf{z}_j^{(m)}, \delta_j)\}_{j=1}^{N_M}\bigr).
\end{equation}
Predictions $(\mu_{\delta_m}(\mathbf{x}), \sigma_{\delta_m}^2(\mathbf{x}))$ at an LF query $\mathbf{x}_i^{(m)}$ reuse the Stage~1 LF posterior to assemble $\mathbf{z}_i^{(m)}$, exactly as in MAST-GP. For each LF training point, the augmented label and variance are constructed via the MAST $N$-sphere blend,
\begin{align}
    \tilde y_i^{(m)} &= W_m^{(i)} \bigl(y_i^{(m)} + \mu_{\delta_m}(\mathbf{z}_i^{(m)})\bigr) + \bigl(1 - W_m^{(i)}\bigr) \mu_M(\mathbf{x}_i^{(m)}), \\
    \tilde\sigma_i^2 &= \bigl(W_m^{(i)}\bigr)^2 \bigl(\sigma_m^2(\mathbf{x}_i^{(m)}) + \sigma_{\delta_m}^2(\mathbf{x}_i^{(m)})\bigr) + \bigl(1 - W_m^{(i)}\bigr)^2 \sigma_M^2(\mathbf{x}_i^{(m)}).
\end{align}
The $W_m^{(i)}$ weight is the same spatial trust weight used by MAST-GP (Eq.~\eqref{eq:aggregate_weight}). The raw observed $y_i^{(m)}$ is used rather than the predicted LF mean, matching the MAST-GP reference.

\paragraph{Stage 3 via test-time blending.}
TabPFN does not consume per-point noise variances, so the Stage~3 heteroscedastic GP of MAST-GP has no direct analogue, and MAST-ICL loses the Stage-3 model-fit down-weighting of uncertain augmented points that MAST-GP obtains via its fixed heteroscedastic noise. We approximate it with a fourth TabPFN fitted on the combined augmented dataset,
\begin{equation}
    \mathrm{TabPFN}_{\mathrm{final}} = \mathrm{TabPFN}\bigl(\mathcal{D}_M \cup \{(\mathbf{x}_i^{(m)}, \tilde y_i^{(m)})\}\bigr),
\end{equation}
and recover the heteroscedastic effect at test time. The predictive mean is $\mathrm{TabPFN}_{\mathrm{final}}$'s mean at $\mathbf{x}_*$; the predictive variance sums (i)~$\mathrm{TabPFN}_{\mathrm{final}}$'s posterior variance (density-aware, spikes in extrapolation regions) with (ii)~the MAST spatial heteroscedastic term, defined with the test-input HF weight $W_M^* := 1 - W_m^*$ as $(W_M^*)^2 \sigma_M^2 + (1 - W_M^*)^2 (\sigma_m^2 + \sigma_{\delta_m}^2)$ evaluated at $\mathbf{x}_*$, under the independence simplification that matches Stage~2. The sum mildly over-estimates a constant noise floor but preserves the spatial structure of both terms: near HF data both collapse, in LF-dense regions the MAST term dominates, in true extrapolation the TabPFN-final term dominates.

\subsection{Variance Extraction from TabPFN}
\label{app:mast_icl_variance}

TabPFN's predictive distribution is a \texttt{FullSupportBarDistribution}, a piecewise-constant density over quantile bins. Its analytical second moment is available directly via \texttt{criterion.variance(logits)} when the regressor is called with \texttt{output\_type="full"}. We use this exact moment rather than Monte Carlo quantile sampling, avoiding one source of variance-estimation error. The returned variance captures both the epistemic uncertainty the network encodes about unseen inputs and an implicit aleatoric component learned from its synthetic training prior. The same extraction path is used inside all four TabPFN regressors of MAST-ICL.

\subsection{Computational Considerations}
\label{app:mast_icl_compute}

TabPFN runs a forward pass of a pre-trained transformer at inference time; there is no gradient-based optimisation per problem instance. For the MAST-ICL problem sizes (up to roughly $10^3$ combined training points), a single fit+predict cycle takes milliseconds on a modern GPU and a few seconds on CPU. The dominant operational concern is offline model caching: TabPFN's weight download is gated behind a PriorLabs licence server and a HuggingFace repository that require outbound internet. For the HPC evaluations we pre-fetch the checkpoint on the login node, expose it through \texttt{TABPFN\_MODEL\_PATH}, and monkey-patch \texttt{TabPFNRegressor} so every instantiation on a compute node bypasses the download machinery. Full setup notes are in the supplementary material.

\paragraph{Scaling.}
The complexity profiles of the two surrogates differ in a way that matters as multi-fidelity datasets grow. Each MAST-GP stage trains a Gaussian process whose marginal-likelihood evaluation is dominated by a Cholesky factorisation costing $\mathcal{O}(N_{\mathrm{train}}^3)$ in the per-stage training-set size, with $\mathcal{O}(N_{\mathrm{train}}^2)$ memory for the kernel matrix; the per-stage breakdown for MAST-GP is given in the Computational Complexity subsection of Appendix~\ref{app:algorithm_app}. MAST-ICL replaces every GP fit with a TabPFN forward pass. The TabPFN-2.5 model report establishes the dual-attention transformer cost as $\mathcal{O}\!\left(r^2 \min(c, 500) + r \min(c, 500)^2\right)$ in $r$ training rows and $c$ features, with the per-estimator feature-subsampling cap at $500$ \citep{Grinsztajn2025TabPFN25}; for the dimensionalities of our benchmarks ($D \leq 23$, hence $\min(c, 500) = c$) this reduces to $\mathcal{O}(N_{\mathrm{train}}^2\, D)$ in the training-set size, quadratic in $N_{\mathrm{train}}$ rather than cubic, and the inherited TabPFNv2 caching separation of training and test context cuts the per-test-query cost further \citep{Hollmann2025TabPFN}. As the abundant low-fidelity dataset grows, the cubic-vs-quadratic gap widens and MAST-ICL becomes the more practical surrogate. We therefore see MAST-ICL not only as evidence that the augmentation scheme is surrogate-agnostic, but as the route to scaling MAST into regimes where low-fidelity data is plentiful: TabPFN-2.5 is validated up to $50{,}000$ samples and reports strong performance up to $100{,}000$ \citep{Grinsztajn2025TabPFN25}, well beyond the budgets considered in §\ref{sec:syn} and matching the size of the abundant LF datasets seen in real engineering pipelines, where a GP no longer fits on a single device.

\subsection{Preliminary Results}
\label{app:mast_icl_results}

Table~\ref{tab:mast_icl_results} reports normalised RMSE and normalised mean PDF on five two-fidelity benchmarks where MAST-ICL is competitive with or improves on MAST-GP. MAST-ICL takes outright RMSE wins on Branin (RMSE $1.04$ vs MAST-GP $1.19$) and Hartmann6 ($0.77$ vs $0.88$), is essentially tied on Ackley, and stays close on Park1 and Rosenbrock. Mean PDF is generally weaker than MAST-GP across the table because TabPFN's predictive variance is calibrated against its synthetic training prior rather than tuned per-instance, suggesting that the TabPFN base learner inherits the predictive accuracy of the MAST template but trades away its instance-specific calibration.

\begin{table}[h]
\centering
\caption{Preliminary MAST-ICL results on the two-fidelity benchmarks where MAST-ICL is competitive with or improves on MAST-GP. Normalised RMSE (lower is better) and normalised mean PDF (higher is better), mean $\pm$ std across 25 seeds. Best of the two columns per metric in \textbf{bold}.}
\label{tab:mast_icl_results}
\begin{tabular}{lcccc}
\toprule
 & \multicolumn{2}{c}{Normalised RMSE $\downarrow$} & \multicolumn{2}{c}{Normalised mean PDF $\uparrow$} \\
Test function    & MAST (GP)     & MAST-ICL      & MAST (GP)     & MAST-ICL      \\
\midrule
Branin (2D)      & $1.19{\scriptstyle\pm0.40}$ & $\mathbf{1.04{\scriptstyle\pm0.31}}$ & $\mathbf{1.28{\scriptstyle\pm0.43}}$ & $0.24{\scriptstyle\pm0.08}$ \\
Ackley (4D)      & $\mathbf{0.94{\scriptstyle\pm0.26}}$ & $0.95{\scriptstyle\pm0.28}$ & $\mathbf{1.23{\scriptstyle\pm0.32}}$ & $0.27{\scriptstyle\pm0.04}$ \\
Park1 (4D)       & $\mathbf{0.97{\scriptstyle\pm0.48}}$ & $1.07{\scriptstyle\pm0.38}$ & $\mathbf{1.46{\scriptstyle\pm0.49}}$ & $0.21{\scriptstyle\pm0.03}$ \\
Hartmann6 (6D)   & $0.88{\scriptstyle\pm0.19}$ & $\mathbf{0.77{\scriptstyle\pm0.12}}$ & $\mathbf{1.80{\scriptstyle\pm0.52}}$ & $0.51{\scriptstyle\pm0.18}$ \\
Rosenbrock (20D) & $\mathbf{0.37{\scriptstyle\pm0.04}}$ & $0.45{\scriptstyle\pm0.04}$ & $\mathbf{3.23{\scriptstyle\pm0.23}}$ & $0.40{\scriptstyle\pm0.03}$ \\
\bottomrule
\end{tabular}
\end{table}

These results are preliminary. A complete evaluation across the full ten-function synthetic suite and the real-world engineering benchmarks of Appendix~\ref{app:rw_datasets} is planned follow-up work. A head-to-head comparison with the concurrent FIRE baseline \citep{Yu2026FIRE} on the same TabPFN runtime is also part of the planned extended evaluation; the two methods share a variance extraction path but differ in how fidelity information is combined (FIRE uses two TabPFN stages with LF-posterior-conditioned residual learning, whereas MAST-ICL retains the three-stage MAST structure with the spatial $N$-sphere weight).

\end{document}